\documentclass[letterpaper]{article} % DO NOT CHANGE THIS
\usepackage{aaai2026}  % DO NOT CHANGE THIS
\usepackage{times}  % DO NOT CHANGE THIS
\usepackage{helvet}  % DO NOT CHANGE THIS
\usepackage{courier}  % DO NOT CHANGE THIS
\usepackage[hyphens]{url}  % DO NOT CHANGE THIS
\usepackage{graphicx} % DO NOT CHANGE THIS
\usepackage{natbib}  % DO NOT CHANGE THIS AND DO NOT ADD ANY OPTIONS TO IT
\usepackage{caption} % DO NOT CHANGE THIS AND DO NOT ADD ANY OPTIONS TO IT
\usepackage{algorithm}
\usepackage{algorithmic}

\usepackage{newfloat}
\usepackage{listings}
\DeclareCaptionStyle{ruled}{labelfont=normalfont,labelsep=colon,strut=off} % DO NOT CHANGE THIS
\floatstyle{ruled}
\newfloat{listing}{tb}{lst}{}
\floatname{listing}{Listing}
\usepackage{amsmath}
\usepackage{amssymb}
\usepackage{mathtools}
\usepackage{amsthm}
\usepackage{makecell}
\usepackage{bm}
\usepackage{multirow}
\usepackage{subcaption}
\usepackage{booktabs}
\title{Shaping Parameter Contribution Patterns for Out-of-Distribution Detection}
\author{
    Haonan Xu\textsuperscript{\rm 1},
    Yang Yang\textsuperscript{\rm 1}\footnote{Corresponding author}
}
\affiliations{
    \textsuperscript{\rm 1}Nanjing University of Science and Technology\\
    \{xhnxhn, yyang\}@njust.edu.cn
}

\usepackage{bibentry}
\begin{document}

\maketitle

\begin{abstract}
Out-of-distribution (OOD) detection is a well-known challenge due to deep models often producing overconfident. In this paper, we reveal a key insight that trained classifiers tend to rely on sparse parameter contribution patterns, meaning that only a few dominant parameters drive predictions. This brittleness can be exploited by OOD inputs that anomalously trigger these parameters, resulting in overconfident predictions. To address this issue, we propose a simple yet effective method called Shaping Parameter Contribution Patterns (SPCP), which enhances OOD detection robustness by encouraging the classifier to learn boundary-oriented dense contribution patterns. Specifically, SPCP operates during training by rectifying excessively high parameter contributions based on a dynamically estimated threshold. This mechanism promotes the classifier to rely on a broader set of parameters for decision-making, thereby reducing the risk of overconfident predictions caused by anomalously triggered parameters, while preserving in-distribution (ID) performance. Extensive experiments under various OOD detection setups verify the effectiveness of SPCP.
\end{abstract}


\section{Introduction}
Deep neural networks have been widely applied in various fields \cite{radford2018improving,DBLP:conf/iclr/DosovitskiyB0WZ21,wanprobabilistic} and have achieved remarkable success \cite{DBLP:conf/iccv/LiuL00W0LG21,achiam2023gpt,DBLP:conf/nips/0074WJ024}. However, when deployed in real-world scenarios, deep models may fail by confidently yet erroneously classifying OOD data as one of the predefined training classes \cite{DBLP:conf/cvpr/NguyenYC15,DBLP:conf/cvpr/BendaleB16,DBLP:journals/corr/abs-2110-11334}. The presence of such unreliable behavior can introduce considerable risks, particularly in safety-critical domains like autonomous driving \cite{DBLP:conf/cvpr/GeigerLU12} and medical diagnosis \cite{litjens2017survey}. Therefore, equipping models with the capability to reliably identify and reject predictions for OOD inputs, a task known as OOD detection, is essential to ensure the trustworthiness of AI systems.

To date, extensive efforts have been dedicated to advancing reliable methods for OOD detection \cite{DBLP:journals/corr/abs-2306-09301}. One line of research focuses on designing suitable OOD scoring functions \cite{DBLP:conf/iclr/HendrycksG17,DBLP:conf/nips/LiuWOL20,DBLP:conf/icml/HendrycksBMZKMS22} for a given well-trained model to estimate OOD uncertainty, or to further improve them through post-hoc network adjustments \cite{DBLP:conf/eccv/ZhangLDZBCW20,DBLP:conf/nips/ZhuCXLZ00ZC22,DBLP:conf/iclr/XuCFY24,DBLP:conf/aaai/XuY25}. A complementary line of work aims to strengthen the model's OOD detection capabilities with training-time regularization. Among these, some methods utilize outlier exposure \cite{DBLP:conf/iclr/HendrycksMD19,DBLP:conf/wacv/ZhangILCL23,DBLP:conf/iclr/WangY0DKLH023} to guide the model in learning more robust decision boundaries. However, such outlier data may not always be readily available in practice. Alternatively, other methods tackle OOD detection by imposing favorable constraints during the training process \cite{DBLP:conf/icml/WeiXCF0L22,DBLP:conf/cvpr/RegmiPDGSB22,DBLP:conf/icml/ZhuLYLX023,DBLP:conf/nips/00030P0C24,DBLP:conf/aaai/GhosalSL24}, which do not require any additional data and offer a promising path toward improved performance, being the primary focus of this paper.

\begin{figure}[t]
    \centering
    \begin{subfigure}{0.498\textwidth}
        \centering
        \hspace{-5mm}
        \begin{subfigure}{0.488\textwidth}
            \centering
            \includegraphics[width=\textwidth]{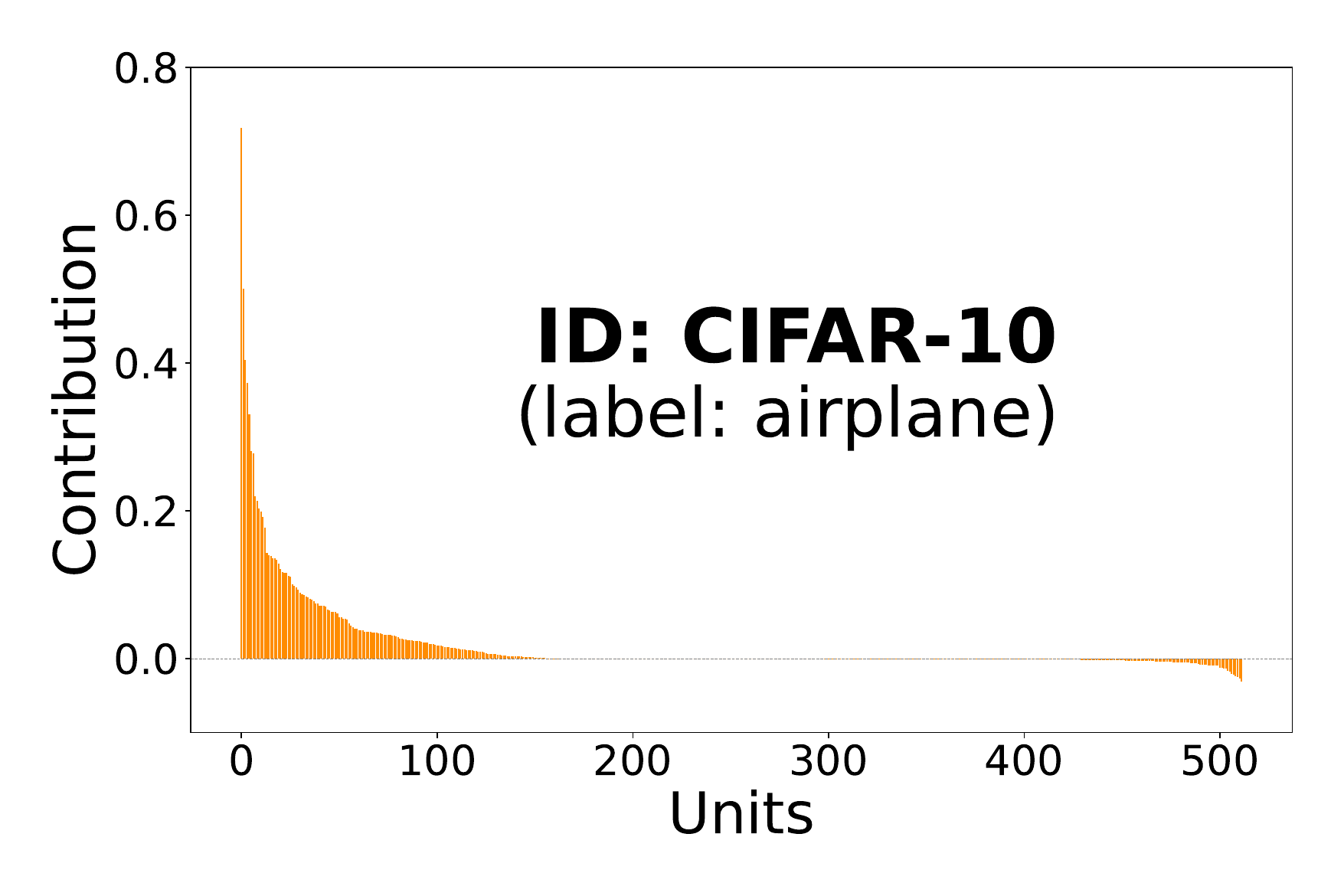}
        \end{subfigure}%
        \captionsetup{skip=1pt}
        \hspace{-3mm}
        \begin{subfigure}{0.488\textwidth}
            \centering
            \includegraphics[width=\textwidth]{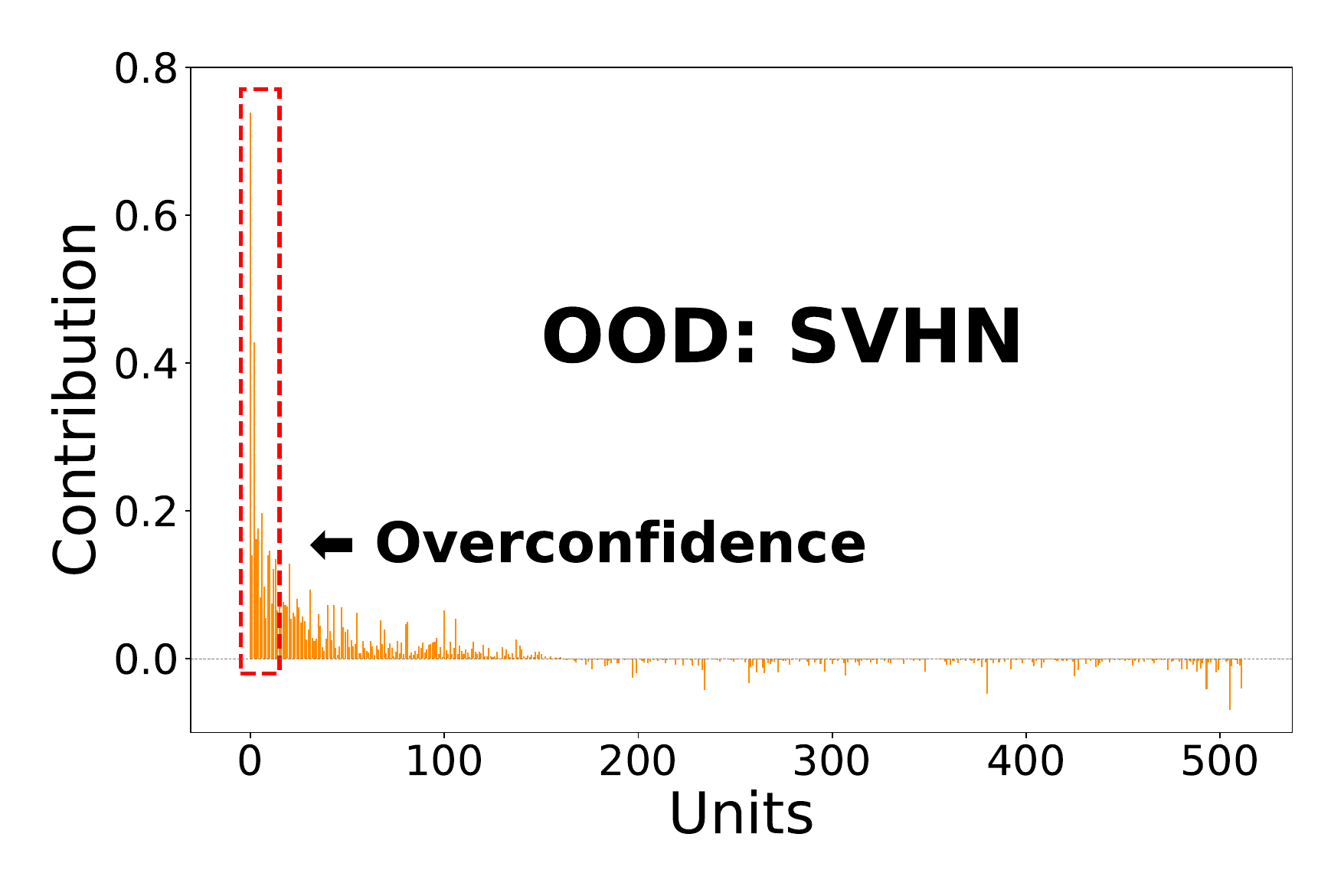}
        \end{subfigure}
    \end{subfigure}
    % \vspace{-4mm}
    \captionsetup{skip=0.5pt}
    % \caption*{}
    \captionsetup{skip=1.0pt}
    \caption{An illustrative example of parameter contribution patterns in a ResNet-18 classifier trained on CIFAR-10 using cross-entropy loss. Units are sorted in the same order. This example focuses on the ‘airplane’ class, showcasing the contributions of each unit in the classifier weights to the model outputs for both ID and OOD samples. The parameter contributions are defined with reference to Eq. \eqref{eq:contribution}. Since model outputs are directly determined by parameter contributions, OOD input that anomalously triggers dominant parameters can be confidently yet incorrectly classified as ID categories and hurt OOD detection.}
    \label{fig:intro}
\end{figure}

In this paper, we present a novel perspective on the causes of model overconfidence by examining how classifier parameters contribute to predictions. As illustrated in Figure \ref{fig:intro}, we reveal the empirical observation that a well-trained model's classifier tends to exhibit sparse contribution patterns in its predictions (see more examples in Appendix A), a finding also supported by \cite{DBLP:conf/iclr/FrankleC19,DBLP:conf/eccv/SunL22a}. The sparsity of parameter contribution patterns indicates that a small subset of parameters dominates the classifier's predictions. This pattern makes OOD detection brittle because some of these dominant parameters can be anomalously triggered by OOD inputs \cite{DBLP:conf/cvpr/NguyenYC15,DBLP:conf/nips/SunGL21}. Such parameters are exploited by OOD inputs, leading to overconfident predictions favoring ID categories (as shown in the right panel of Figure \ref{fig:intro}). Therefore, suppressing the dominance of parameters with excessive predictive power is crucial for robust OOD detection.

The above analysis naturally motivates the proposal of Shaping Parameter Contribution Patterns (SPCP), a simple yet effective method for OOD detection. The core idea behind SPCP is to induce contribution bounds for the classifier's parameters to enhance the robustness of OOD detection. To achieve this, SPCP truncates parameter contributions that exceed a dynamically estimated threshold during the training process. This strategy enforces bounded constraints on the dominant parameter contributions, compelling the classifier to rely on a broader subset of parameters during decision-making. As a result, SPCP reduces the risk of overconfident predictions potentially caused by anomalously triggered parameters, making the OOD scores derived from the calibrated model more reliable and improving the ID-OOD separation. Extensive experiments on the authoritative OpenOOD benchmark \cite{DBLP:journals/corr/abs-2306-09301} verify the effectiveness of SPCP, demonstrating that our SPCP can enhance the model's OOD detection capability across various OOD scenarios, while also preserving ID task performance.

\section{Preliminaries}
\noindent\textbf{Setup.} 
In this paper, we focus on the setting of $K$-way image classification. Formally, let $\mathcal{X}$ denote the input space and let the ID label space be defined as $\mathcal{Y} = \left\{1, 2, ..., K\right\}$. The learner has access to a labeled training set $\mathcal{D}_{\text{train}} = \{(\mathbf{x}_i, \mathbf{y}_i)\}_{i=1}^n$, where samples are drawn \emph{i.i.d.} from a joint distribution $\mathcal{P}_{\mathcal{X}\mathcal{Y}}$. Let $f: \mathcal{X} \to \mathbb{R}^K$ denote the classification model. For typical architectures, $f$ first extracts a $D$-dimensional penultimate feature representation $h(\mathbf{x}) \in \mathbb{R}^D$ from an input $\mathbf{x} \in \mathcal{X}$. The classifier layer, parameterized by a weight matrix $\mathbf{W} \in \mathbb{R}^{D \times K}$ and a bias vector $\mathbf{b} \in \mathbb{R}^K$, then maps $h(\mathbf{x})$ to the output vector $f(\mathbf{x}) \in \mathbb{R}^K$. Formally, the output of the model can be written as:
\begin{equation}
f(\mathbf{x}) = \mathbf{W}^\top h(\mathbf{x}) + \mathbf{b}.
\end{equation}

\noindent\textbf{Out-of-distribution Detection.} The goal of OOD detection is to determine whether a given input $\mathbf{x}$ originates from an irrelevant distribution whose label set does not intersect with $\mathcal{Y}$ \cite{DBLP:journals/corr/abs-2110-11334}. In practice, this task is often formulated as a binary decision problem using level-set estimation:
\begin{align}
g_\tau(\mathbf{x}) &= 
\begin{cases}
\text{ID},  & \text{if } S(\mathbf{x}; f) > \tau; \\
\text{OOD}, & \text{if } S(\mathbf{x}; f) \leq \tau,
\end{cases}
\end{align}
where $\tau$ is a threshold typically chosen such that the majority (\emph{e.g.}, 95\%) of ID data are correctly classified, and $S(\mathbf{x}; f): \mathcal{X} \to \mathbb{R}$ is the scoring function that quantifies the model's predictive uncertainty as a scalar value. A widely adopted option for $S(\mathbf{x}; f)$ is the Energy score \cite{DBLP:conf/nips/LiuWOL20}, which is defined as follows:
\begin{equation}
~S_{\text{Energy}}(\mathbf{x}; f) = \log{\sum\limits_{k\in\mathcal{Y}}{\exp(f_k(\mathbf{x}))}},
\end{equation}
where $f_k(\cdot)$ denotes the $k$-th output logit. The underlying assumption is that OOD inputs are expected to receive lower scores than ID cases, as they lie outside the training label space. However, OOD detection remains a non-trivial task because deep models may behave overconfidently when presented with OOD data \cite{DBLP:conf/cvpr/NguyenYC15}.

\section{Proposed Method}
% This section begins by defining parameter contribution, proceeds to introduce our method for shaping contribution patterns, and concludes with a justification of the key insights.

\subsection{Defining the Parameter Contribution}
For a given input $\mathbf{x}$, the contribution $c_k(\mathbf{x};\boldsymbol{\theta}_{ij})$ of a specific parameter $\boldsymbol{\theta}_{ij}$ to class $k$ is defined as the change in the model's $k$-th output when $\boldsymbol{\theta}_{ij}$ is present versus when it is absent (\emph{i.e.}, setting $\boldsymbol{\theta}_{ij}$ to 0) \cite{DBLP:conf/aaai/XuY25}, formally expressed as:
\begin{equation}
\label{eq:contribution}
c_k(\mathbf{x};\boldsymbol{\theta}_{ij}) = f_k(\mathbf{x}) - f_k(\mathbf{x};\boldsymbol{\theta}_{ij}=0).
\end{equation}
Consistent with prior work \cite{DBLP:conf/eccv/SunL22a, DBLP:conf/nips/ChenFLCTY23, DBLP:conf/aaai/XuY25}, we primarily focus on the contribution of the parameters that are directly responsible for making predictions, \emph{i.e.}, the classifier weight matrix $\mathbf{W}$. This focus is motivated by two main considerations: (1) the adjustment of the deepest classifier layer is most effective for OOD detection, as the high-level semantics captured by the classifier layer are generally the most relevant and impactful for identifying OOD samples \cite{DBLP:conf/nips/ZhuCXLZ00ZC22,DBLP:conf/aaai/XuY25}, and (2) the per-parameter contribution of the classifier layer is computationally efficient, since the contribution of the element $\mathbf{W}_{ij}$ to the $k$-th class can be expressed in a simplified form (see Appendix B for details):
\begin{equation}
c_{k}(\mathbf{x}; \mathbf{W}_{ij}) =
\begin{cases} 
\mathbf{W}_{ij} \cdot h_{i}(\mathbf{x}), & \text{if } k = j, \\
0, & \text{if } k \ne j,
\end{cases}
\end{equation}
where each $\mathbf{W}_{ij}$ is exclusively responsible for the corresponding class $j$. Accordingly, the model's $k$-th output $f_k(\mathbf{x})$ can be rewritten in terms of the classifier's weight parameter contributions as:
\begin{equation} 
f_k(\mathbf{x}) = \sum_{d=1}^{D}{c_{k}(\mathbf{x}; \mathbf{W}_{dk})} + \mathbf{b}_k. 
\end{equation}

\subsection{Shaping Parameter Contribution Patterns}
\noindent\textbf{Motivation.} As illustrated in Figure \ref{fig:intro}, the classifier layer typically relies on sparse contribution patterns when making predictions. That is, the model’s decisions are driven by a small subset of parameters that exert disproportionately high influence. This brittleness can be exploited by OOD inputs that anomalously trigger these parameters \cite{DBLP:conf/cvpr/NguyenYC15,DBLP:conf/nips/SunGL21}, resulting in overconfident predictions. This critical issue motivates us to propose SPCP, which encourages the model to learn a more robust contribution pattern for improved OOD detection.

\noindent\textbf{Training Procedure.} SPCP explicitly imposes an upper bound on the contributions during the training process, applied element-wise to the weight parameters $\mathbf{W}$ of the model's classifier layer:
\begin{equation}
\label{equ: bound}
\overline{c}^\lambda_{k}(\mathbf{x}; \mathbf{W}_{ij}) = \min(c_{k}(\mathbf{x}; \mathbf{W}_{ij}),\lambda),
\end{equation}
where $\lambda$ denotes the threshold. As $\lambda \to \infty$, Eq. \eqref{equ: bound} becomes equivalent to the original unbounded formulation. In effect, this operation truncates the contribution above $\lambda$ to prevent the classifier's parameters from producing disproportionately large contributions. In this context, the model output for the $k$-th class after applying SPCP is given by:
\begin{equation} 
f^{\text{SPCP}}_k(\mathbf{x};\lambda) = \sum_{d=1}^D{\overline{c}^\lambda_{k}(\mathbf{x}; \mathbf{W}_{dk})} + \mathbf{b}_k. 
\end{equation}
Similarly to previous methods \cite{DBLP:conf/nips/SunGL21,DBLP:conf/nips/ChenFLCTY23}, we obtain the threshold $\lambda$ by using a percentile $\rho$. Let $C(\mathbf{x})$ denote the contribution matrix \emph{w.r.t.} the classifier weights $\mathbf{W}$ for a given sample $\mathbf{x}$. The threshold $\lambda$ is set to the values corresponding to the top $\rho$-th percentile of $C(\mathbf{x})$, averaged over the entire training set $\mathcal{D}_{\text{train}}$. To adapt $\lambda$ to the dynamic behavior of training while adhering to mini-batch processing, we apply an Exponential Moving Average (EMA) for estimation at each iteration $t$:
\begin{equation}
\label{equ: p}
\lambda_{t+1} = \beta \cdot \lambda_t + (1 - \beta) \cdot \frac{1}{|\mathcal{B}_t|} \sum_{\mathbf{x}_i \in \mathcal{B}_t} \text{Top}(\rho, C(\mathbf{x}_i)),
\end{equation}
where $\beta \in [0, 1]$ is the smoothing factor controlling the update rate, $\mathcal{B}_t$ denotes the mini-batch of data at the $t$-th iteration, $|\mathcal{B}_t|$ denotes the cardinality of the set $\mathcal{B}_t$, and $\text{Top}(\rho, C(\mathbf{x}))$ refers to the value corresponding to the top $\rho$-th percentile of the contribution matrix $C(\mathbf{x})$ for the given sample $\mathbf{x}$. To stabilize early training, $\lambda_0$ is intentionally initialized with a relatively large value for warm-up. The learning objective, based on the cross-entropy loss $\ell_{\text{CE}}$, is to minimize the following expected risk:
\begin{equation} 
\mathcal{R}(f^{\text{SPCP}})=\mathbb{E}_{(\mathbf{x},\mathbf{y})\in\mathcal{D}_{\text{train}}}\ell_{\text{CE}}(f^{\text{SPCP}}(\mathbf{x};\lambda),\mathbf{y}).
\end{equation}

\noindent\textbf{Inferring Procedure.} SPCP adopts the Energy Score \cite{DBLP:conf/nips/LiuWOL20} as the default OOD scoring function, as it is provably aligned with the input density and generally performs well. Given a test input $\mathbf{x}$, the predictive uncertainty of the model after applying SPCP is quantified as:
\begin{equation} 
S_{\text{SPCP}}(\mathbf{x};f^{\text{SPCP}}) = \log \sum_{k \in \mathcal{Y}} \exp(f_k^{\text{SPCP}}(\mathbf{x};\lambda)).
\end{equation}
To ensure consistency, $\lambda$ is set to the value estimated from Eq. \eqref{equ: p} at the point of training completion. The pseudo code of SPCP is available in Appendix K.

\subsection{Insight Justification}

\begin{figure}[t]
    \centering
    \begin{subfigure}{0.495\textwidth}
        \centering
        \hspace{-3mm}
        \begin{subfigure}{0.495\textwidth}
            \centering
            \includegraphics[width=\textwidth]{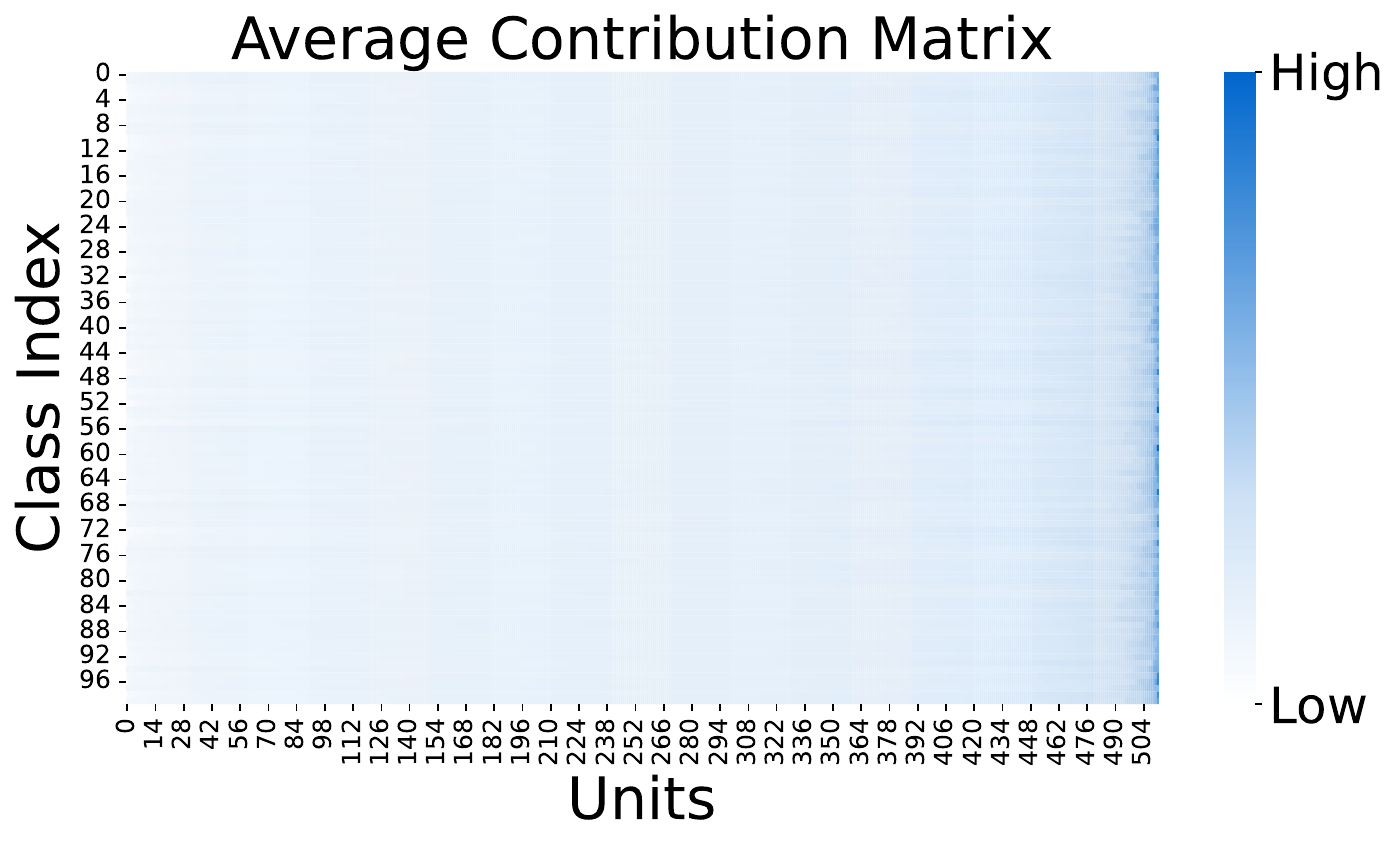}
            \captionsetup{skip=1pt}
            \caption{Vanilla training}
        \end{subfigure}%
        \hspace{-1.5mm}
        \captionsetup{skip=1pt}
        \begin{subfigure}{0.495\textwidth}
            \centering
            \includegraphics[width=\textwidth]{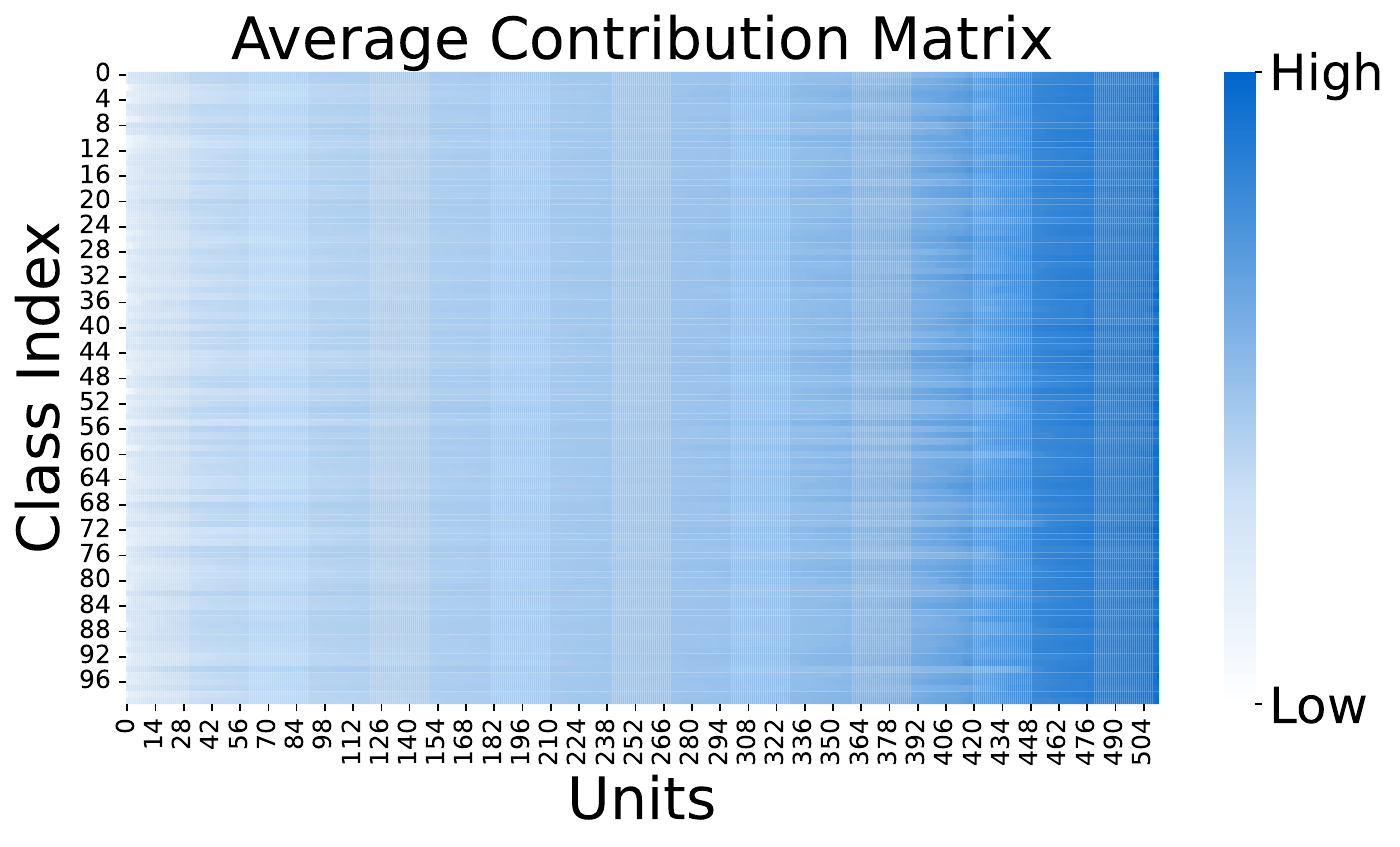}
            \captionsetup{skip=1pt}
            \caption{SPCP training}
        \end{subfigure}
    \end{subfigure}%
    \captionsetup{skip=1.0pt}
    \caption{Comparison of the classifier's parameter contribution patterns before and after applying SPCP, with the average contribution matrix on the ID test set sorted for clarity.}
    \label{fig:remark1}
\end{figure}

\noindent\textbf{Remark 1. SPCP effectively prevents predictions from being dominated by a small subset of classifier's parameters.}
In modern deep learning models, many design choices drive sparse contribution patterns, including over-parameterization \cite{DBLP:conf/eccv/SunL22a}, regularization techniques (\emph{e.g.}, $L_{1}$ regularization \cite{tibshirani1996regression}), and sparsity-inducing activations (\emph{e.g.}, ReLU \cite{DBLP:journals/corr/abs-1803-08375}). Such sparsity causes a small subset of parameters to dominate the prediction disproportionately. In response, our proposed SPCP explicitly enforces an upper bound $\lambda$ on the contributions of the classifier’s parameters. This restriction limits their excessive influence on predictions during training and encourages the adoption of relatively denser contribution patterns. As shown in Figure \ref{fig:remark1}, the model's decisions are based on a broader set of parameters after SPCP training, effectively mitigating the prediction from being dominated by only a few parameters.

\begin{figure}[t]
    \centering
    \begin{subfigure}{0.47\textwidth}
        \centering
        \begin{subfigure}{0.4\textwidth}
            \centering
            \includegraphics[width=\textwidth]{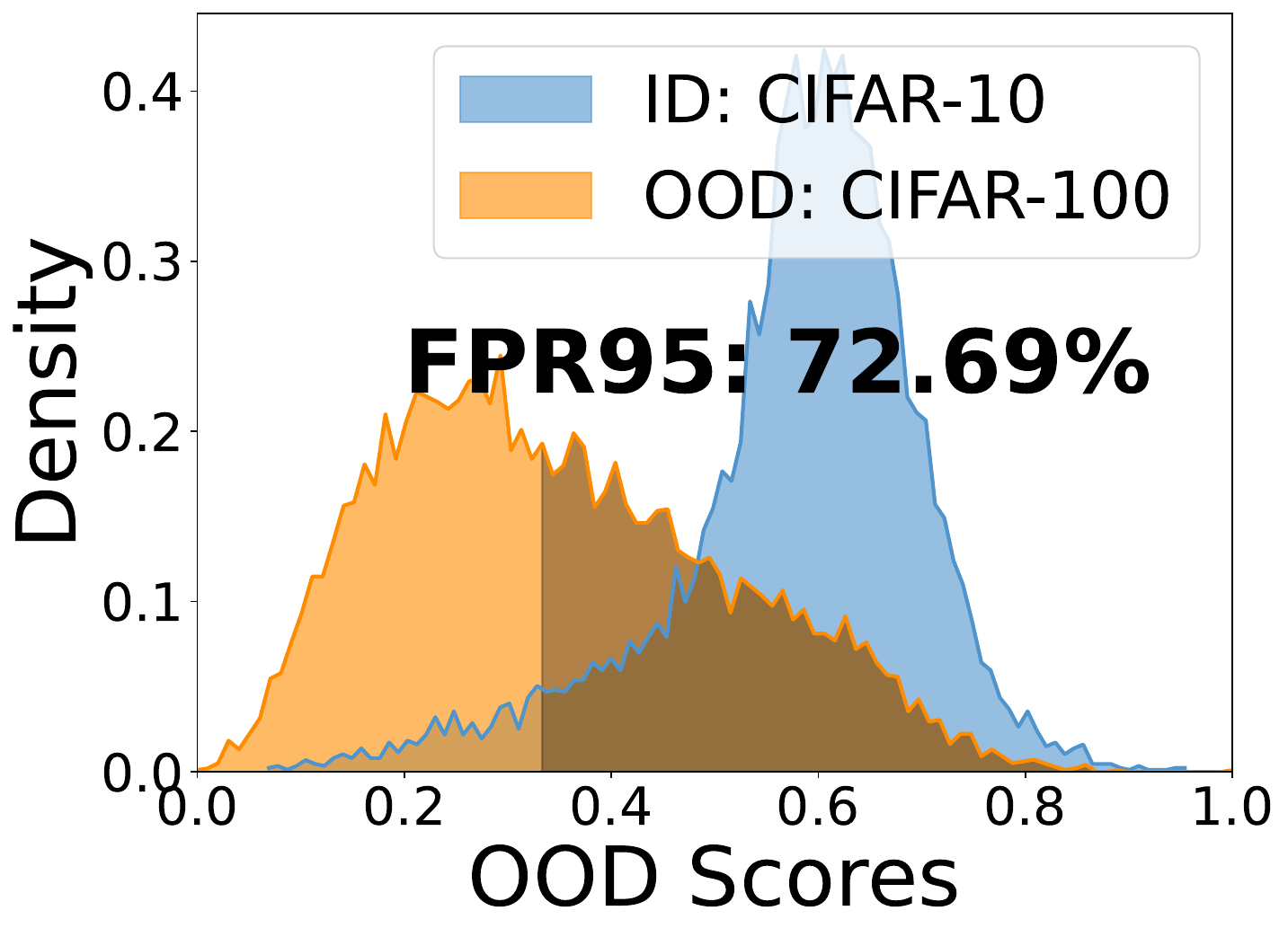}
            \captionsetup{skip=1pt}
            \caption*{~~~~Energy}
        \end{subfigure}%
        \captionsetup{skip=1pt}
        \hspace{8mm}
        \begin{subfigure}{0.4\textwidth}
            \centering
            \includegraphics[width=\textwidth]{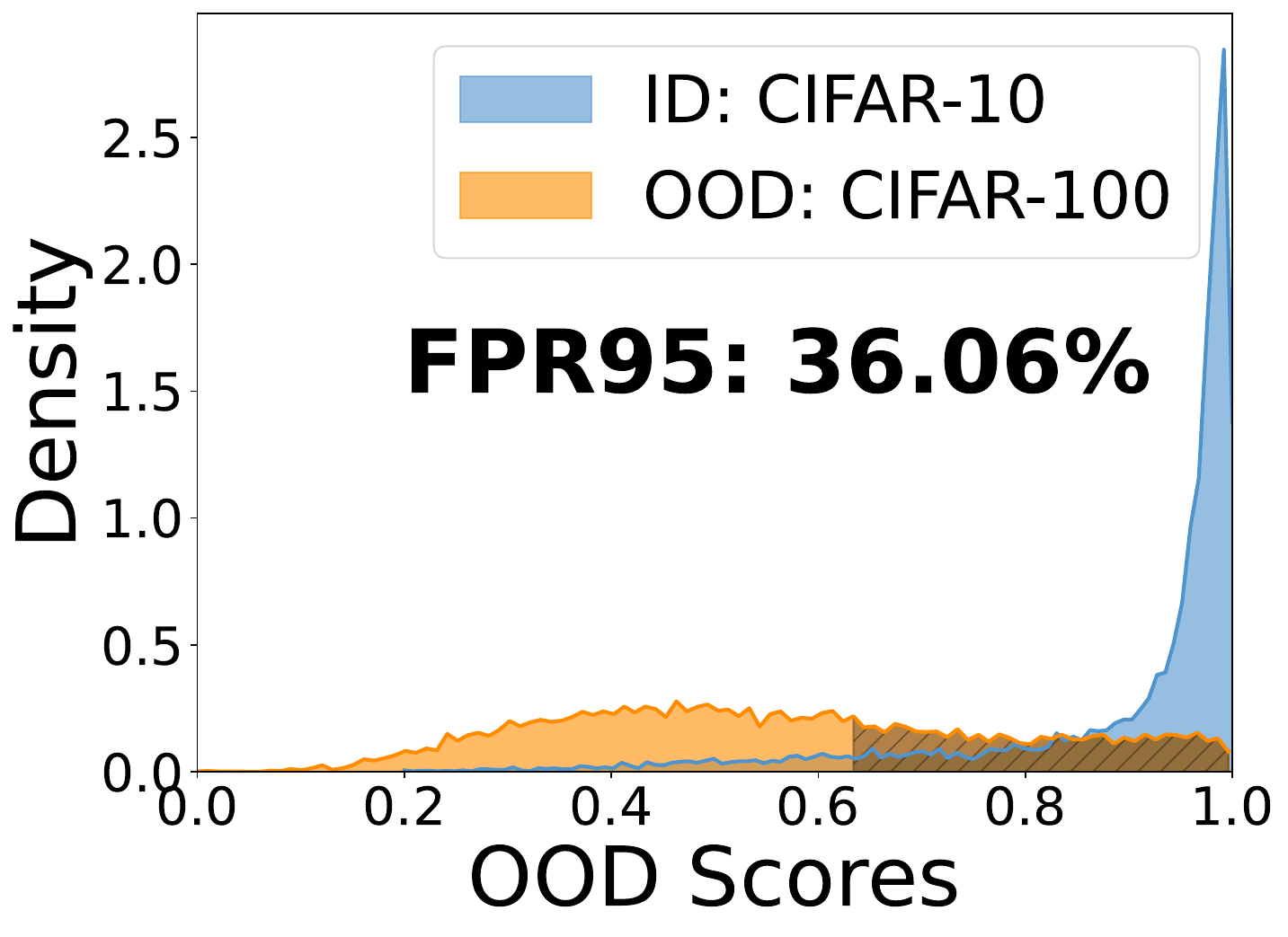}
            \captionsetup{skip=1pt}
            \caption*{~~~~SPCP}
        \end{subfigure}
        \caption{SPCP on near-OOD scenarios}
    \end{subfigure}%
    \hspace{5mm}
    \begin{subfigure}{0.47\textwidth}
        \centering
        \begin{subfigure}{0.4\textwidth}
            \centering
            \includegraphics[width=\textwidth]{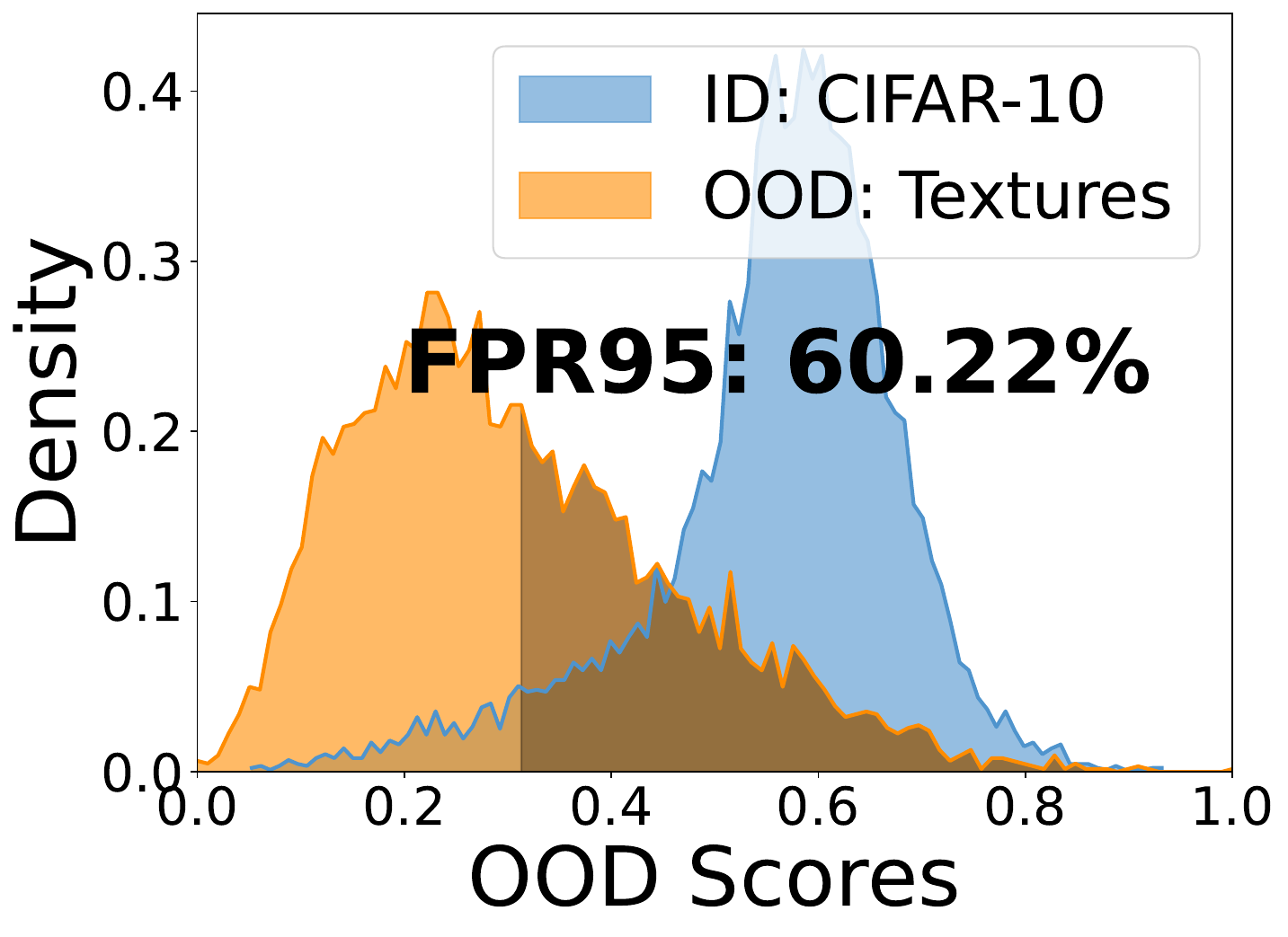}
            \captionsetup{skip=1pt}
            \caption*{~~~~Energy}
        \end{subfigure}%
        \captionsetup{skip=1pt}
        \hspace{8mm}
        \begin{subfigure}{0.4\textwidth}
            \centering
            \includegraphics[width=\textwidth]{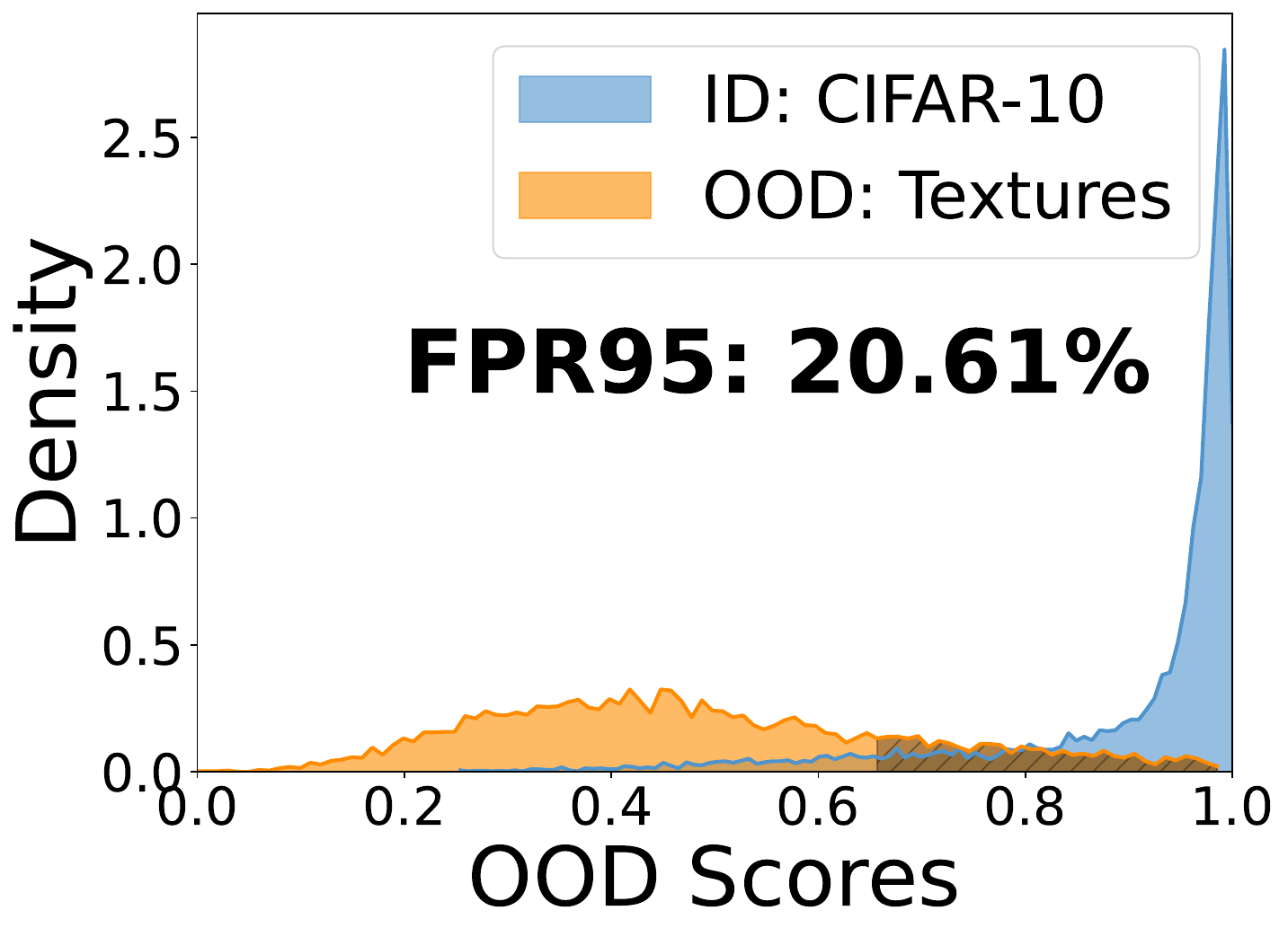}
            \captionsetup{skip=1pt}
            \caption*{~~~~SPCP}
        \end{subfigure}
        \caption{SPCP on far-OOD scenarios}
    \end{subfigure}
    % \vspace{-5mm}
    \caption{Comparison of normalized OOD score distributions before and after applying SPCP.}
    \label{fig:remark2}
\end{figure}

\begin{table*}[t]
    % \vspace{-3mm}
    \begin{center}
    \resizebox{0.865\textwidth}{!}{ 
        \begin{tabular}{@{\hspace{0.10cm}}c@{\hspace{0.15cm}}|@{\hspace{0.15cm}}l@{\hspace{0.15cm}}|c@{\hspace{0.15cm}}c@{\hspace{0.15cm}}|c@{\hspace{0.15cm}}c@{\hspace{0.10cm}}}
        \toprule[1.5pt]
        \multicolumn{2}{c|}{ID datasets} & \multicolumn{2}{c|}{CIFAR-10} & \multicolumn{2}{c}{CIFAR-100} \\
        \cmidrule{1-6}
        \multicolumn{2}{c|}{OOD datasets} & FPR95$\downarrow$ & AUROC$\uparrow$ & FPR95$\downarrow$ & AUROC$\uparrow$ \\ 
        \midrule
        \multicolumn{2}{c|}{} & \multicolumn{4}{c}{Vanilla training / \textbf{SPCP training (Ours)}} \\
        \midrule
        \multirow{4}{*}{Near-OOD} & CIFAR-10  & -- & -- & \textbf{59.21{\scriptsize$\pm$0.75}} / 60.10{\scriptsize$\pm$0.95} & 79.05{\scriptsize$\pm$0.11} / \textbf{79.09{\scriptsize$\pm$0.22}} \\
        & CIFAR-100 & 66.60{\scriptsize$\pm$4.46} / \textbf{35.74{\scriptsize$\pm$0.51}} & 86.36{\scriptsize$\pm$0.58} / \textbf{90.42{\scriptsize$\pm$0.17}} & -- & -- \\
        & TIN & 56.08{\scriptsize$\pm$4.83} / \textbf{27.59{\scriptsize$\pm$0.48}} & 88.80{\scriptsize$\pm$0.36} / \textbf{92.72{\scriptsize$\pm$0.07}} & 52.03{\scriptsize$\pm$0.50} / \textbf{50.33{\scriptsize$\pm$0.85}} & 82.76{\scriptsize$\pm$0.08} / \textbf{83.40{\scriptsize$\pm$0.28}} \\
        & Average & 61.34{\scriptsize$\pm$4.63} / \textbf{31.67{\scriptsize$\pm$0.25}} & 87.58{\scriptsize$\pm$0.46} / \textbf{91.57{\scriptsize$\pm$0.11}} & 55.62{\scriptsize$\pm$0.61} / \textbf{55.21{\scriptsize$\pm$0.85}} & 80.91{\scriptsize$\pm$0.08} / \textbf{81.25{\scriptsize$\pm$0.19}} \\
        \midrule
        \multirow{5}{*}{Far-OOD} & MNIST & 24.99{\scriptsize$\pm$12.93} / \textbf{19.09{\scriptsize$\pm$1.14}} & 94.32{\scriptsize$\pm$2.53} / \textbf{94.76{\scriptsize$\pm$0.69}} & 52.62{\scriptsize$\pm$3.83} / \textbf{49.33{\scriptsize$\pm$3.55}} & 79.18{\scriptsize$\pm$1.37} / \textbf{80.98{\scriptsize$\pm$2.70}} \\
        & SVHN & 35.12{\scriptsize$\pm$6.11} / \textbf{16.52{\scriptsize$\pm$2.74}} & 91.79{\scriptsize$\pm$0.98} / \textbf{95.35{\scriptsize$\pm$1.33}} & 53.62{\scriptsize$\pm$3.14} / \textbf{43.00{\scriptsize$\pm$6.73}} & 82.03{\scriptsize$\pm$1.74} / \textbf{86.02{\scriptsize$\pm$3.60}} \\
        & Textures & 51.82{\scriptsize$\pm$6.11} / \textbf{20.09{\scriptsize$\pm$0.57}} & 89.47{\scriptsize$\pm$0.70} / \textbf{95.00{\scriptsize$\pm$0.03}} & 62.35{\scriptsize$\pm$2.06} / \textbf{54.87{\scriptsize$\pm$1.21}} & 78.35{\scriptsize$\pm$0.83} / \textbf{81.32{\scriptsize$\pm$0.28}} \\
        & Places365 & 54.85{\scriptsize$\pm$6.52} / \textbf{26.47{\scriptsize$\pm$0.54}} & 89.25{\scriptsize$\pm$0.78} / \textbf{93.09{\scriptsize$\pm$0.12}} & 57.75{\scriptsize$\pm$0.86} / \textbf{55.16{\scriptsize$\pm$0.61}} & 79.52{\scriptsize$\pm$0.23} / \textbf{80.37{\scriptsize$\pm$0.38}} \\
        & Average & 41.69{\scriptsize$\pm$5.32} / \textbf{20.54{\scriptsize$\pm$0.67}} & 91.21{\scriptsize$\pm$0.92} / \textbf{94.55{\scriptsize$\pm$0.42}} & 56.59{\scriptsize$\pm$1.38} / \textbf{50.59{\scriptsize$\pm$1.83}} & 79.77{\scriptsize$\pm$0.61} / \textbf{82.17{\scriptsize$\pm$1.32}} \\
        \bottomrule[1.5pt]
        \end{tabular}
    }
    \caption{OOD detection performance on CIFAR benchmarks. All values are percentages, and the best results are in bold. $\uparrow$ indicates that larger values are better, $\downarrow$ indicates that smaller values are better.}
    \label{tab:vanilla_cifar} 
    \end{center}
\end{table*}

\begin{table*}[t]
    % \vspace{-3mm}
    \begin{center}
    \resizebox{0.63\textwidth}{!}{ 
        \begin{tabular}{@{\hspace{0.10cm}}c@{\hspace{0.45cm}}|@{\hspace{0.45cm}}l@{\hspace{0.45cm}}|c@{\hspace{0.65cm}}c@{\hspace{0.10cm}}}
        \toprule[1.5pt]
        \multicolumn{2}{c|}{ID datasets} & \multicolumn{2}{c}{ImageNet-200} \\
        \cmidrule{1-4}
        \multicolumn{2}{c|}{OOD datasets} & FPR95$\downarrow$ & AUROC$\uparrow$ \\ 
        \midrule
        \multicolumn{2}{c|}{} & \multicolumn{2}{c}{Vanilla training / \textbf{SPCP training (Ours)}} \\
        \midrule
        \multirow{3}{*}{Near-OOD} & SSB-hard  & \textbf{69.77{\scriptsize$\pm$0.32}} / 70.23{\scriptsize$\pm$0.21} & \textbf{79.83{\scriptsize$\pm$0.02}} / 79.80{\scriptsize$\pm$0.05} \\
        & NINCO & 50.70{\scriptsize$\pm$0.89} / \textbf{49.32{\scriptsize$\pm$0.40}} & 85.17{\scriptsize$\pm$0.11} / \textbf{85.24{\scriptsize$\pm$0.06}} \\
        & Average & 60.24{\scriptsize$\pm$0.57} / \textbf{59.77{\scriptsize$\pm$0.25}} & 82.50{\scriptsize$\pm$0.05} / \textbf{82.52{\scriptsize$\pm$0.03}} \\
        \midrule
        \multirow{4}{*}{Far-OOD} & iNaturalist & 26.41{\scriptsize$\pm$2.29} / \textbf{23.86{\scriptsize$\pm$0.94}} & 92.55{\scriptsize$\pm$0.50} / \textbf{92.91{\scriptsize$\pm$0.08}} \\
        & Textures & 41.43{\scriptsize$\pm$1.85} / \textbf{33.83{\scriptsize$\pm$0.69}} & 90.79{\scriptsize$\pm$0.16} / \textbf{91.73{\scriptsize$\pm$0.21}} \\
        & OpenImage-O & 36.74{\scriptsize$\pm$1.14} / \textbf{33.59{\scriptsize$\pm$0.34}} & 89.23{\scriptsize$\pm$0.26} / \textbf{89.75{\scriptsize$\pm$0.16}} \\
        & Average & 34.86{\scriptsize$\pm$1.30} / \textbf{30.43{\scriptsize$\pm$0.45}} & 90.86{\scriptsize$\pm$0.21} / \textbf{91.46{\scriptsize$\pm$0.14}} \\
        \bottomrule[1.5pt]
        \end{tabular}
    }
    \caption{OOD detection performance on large-scale ImageNet benchmark.}
    \label{tab:vanilla_imagenet} 
    \end{center}
\end{table*}

\noindent\textbf{Remark 2. SPCP mitigates overconfidence by shaping boundary-oriented dense contribution patterns.}
Due to deep models that may activate neurons even for unfamiliar inputs \cite{DBLP:conf/cvpr/NguyenYC15,DBLP:conf/nips/SunGL21}, the behavior of dominant parameters may pose significant risks under sparse contribution patterns. Specifically, anomalous triggering of dominant parameters can drive the prediction toward incorrect ID categories with high confidence, resulting in overconfident predictions. To address this issue, our proposed SPCP guides the model to learn boundary-oriented dense contribution patterns, thereby enhancing the model's resilience to anomalous parameter contributions induced by OOD inputs. As shown in Figure \ref{fig:remark2}, SPCP reduces the overlap in the right tails of the OOD score distributions between OOD and ID samples, achieving improved ID-OOD separation in both near-OOD and far-OOD scenarios. These results demonstrate that SPCP effectively mitigates the model’s overconfidence on OOD inputs.

\begin{table*}[t]
    % \vspace{-3mm}
    \begin{center}
    \resizebox{\textwidth}{!}{
        \begin{tabular}{@{\hspace{0.10cm}}l|@{\hspace{0.15cm}}c@{\hspace{0.15cm}}c@{\hspace{0.15cm}}c@{\hspace{0.15cm}}c@{\hspace{0.15cm}}c@{\hspace{0.15cm}}|c@{\hspace{0.15cm}}c@{\hspace{0.15cm}}c@{\hspace{0.15cm}}c@{\hspace{0.15cm}}c@{\hspace{0.10cm}}}
        \toprule[1.5pt]
        \multirow{3}{*}{\textbf{Method}} & \multicolumn{5}{@{\hspace{0.15cm}}c@{\hspace{0.15cm}}|}{\textbf{CIFAR-10}} & \multicolumn{5}{@{\hspace{0.15cm}}c@{\hspace{0.15cm}}}{\textbf{CIFAR-100}} \\
        \cmidrule{2-11}
        & \multicolumn{2}{c}{\textbf{Near-OOD}} & \multicolumn{2}{c}{\textbf{Far-OOD}} & \multirow{2}{*}{ID ACC$\uparrow$} & \multicolumn{2}{c}{\textbf{Near-OOD}} & \multicolumn{2}{c}{\textbf{Far-OOD}} & \multirow{2}{*}{ID ACC$\uparrow$} \\
        & FPR95$\downarrow$ & AUROC$\uparrow$ & FPR95$\downarrow$ & AUROC$\uparrow$ & & FPR95$\downarrow$ & AUROC$\uparrow$ & FPR95$\downarrow$ & AUROC$\uparrow$ \\
        \midrule
        & \multicolumn{10}{c}{\textit{Post-hoc methods (vanilla training with cross-entropy)}} \\
        MSP \cite{DBLP:conf/iclr/HendrycksG17} & 48.17{\scriptsize$\pm$3.92} & 88.03{\scriptsize$\pm$0.25} & 31.72{\scriptsize$\pm$1.84} & 90.73{\scriptsize$\pm$0.43} & \underline{95.06{\scriptsize$\pm$0.30}} & \textbf{54.80{\scriptsize$\pm$0.33}} & 80.27{\scriptsize$\pm$0.11} & 58.70{\scriptsize$\pm$1.06} & 77.76{\scriptsize$\pm$0.44} & \underline{77.25{\scriptsize$\pm$0.10}} \\
        Energy \cite{DBLP:conf/nips/LiuWOL20} & 61.34{\scriptsize$\pm$4.63} & 87.58{\scriptsize$\pm$0.46} & 41.69{\scriptsize$\pm$5.32} & 91.21{\scriptsize$\pm$0.92} & \underline{95.06{\scriptsize$\pm$0.30}} & \underline{55.62{\scriptsize$\pm$0.61}} & \underline{80.91{\scriptsize$\pm$0.08}} & 56.59{\scriptsize$\pm$1.38} & 79.77{\scriptsize$\pm$0.61} & \underline{77.25{\scriptsize$\pm$0.10}} \\ 
        % VIM \cite{DBLP:conf/cvpr/Wang0F022} & 44.84{\scriptsize$\pm$2.31} & 88.68{\scriptsize$\pm$0.28} & 25.05{\scriptsize$\pm$0.52} & 93.48{\scriptsize$\pm$0.24} & \underline{95.06{\scriptsize$\pm$0.30}} & 62.63{\scriptsize$\pm$0.27} & 74.98{\scriptsize$\pm$0.13} & \underline{50.74{\scriptsize$\pm$1.00}} & 81.70{\scriptsize$\pm$0.62} & \underline{77.25{\scriptsize$\pm$0.10}} \\
        DICE \cite{DBLP:conf/eccv/SunL22a} & 70.04{\scriptsize$\pm$7.64} & 78.34{\scriptsize$\pm$0.79} & 51.76{\scriptsize$\pm$4.42} & 84.23{\scriptsize$\pm$1.89} & \underline{95.06{\scriptsize$\pm$0.30}} & 57.95{\scriptsize$\pm$0.53} & 79.38{\scriptsize$\pm$0.23} & 56.25{\scriptsize$\pm$0.60} & 80.01{\scriptsize$\pm$0.18} & \underline{77.25{\scriptsize$\pm$0.10}} \\ 
        ReAct \cite{DBLP:conf/nips/SunGL21} & 63.56{\scriptsize$\pm$7.33} & 87.11{\scriptsize$\pm$0.61} & 44.90{\scriptsize$\pm$8.37} & 90.42{\scriptsize$\pm$1.41} & \underline{95.06{\scriptsize$\pm$0.30}} & 56.39{\scriptsize$\pm$0.34} & 80.77{\scriptsize$\pm$0.05} & 54.20{\scriptsize$\pm$1.56} & 80.39{\scriptsize$\pm$0.49} & \underline{77.25{\scriptsize$\pm$0.10}} \\
        ASH \cite{DBLP:conf/iclr/DjurisicBAL23} & 86.78{\scriptsize$\pm$1.82} & 75.27{\scriptsize$\pm$1.04} & 79.03{\scriptsize$\pm$4.22} & 78.49{\scriptsize$\pm$2.58} & \underline{95.06{\scriptsize$\pm$0.30}} & 65.71{\scriptsize$\pm$0.24} & 78.20{\scriptsize$\pm$0.15} & 59.20{\scriptsize$\pm$2.46} & 80.58{\scriptsize$\pm$0.66} & \underline{77.25{\scriptsize$\pm$0.10}} \\
        SCALE \cite{DBLP:conf/iclr/XuCFY24} & 80.45{\scriptsize$\pm$4.02} & 82.55{\scriptsize$\pm$0.36} & 67.53{\scriptsize$\pm$7.50} & 86.39{\scriptsize$\pm$1.86} & \underline{95.06{\scriptsize$\pm$0.30}} & 55.68{\scriptsize$\pm$0.69} & \underline{80.99{\scriptsize$\pm$0.12}} & 54.09{\scriptsize$\pm$1.07} & 81.42{\scriptsize$\pm$0.43} & \underline{77.25{\scriptsize$\pm$0.10}} \\
        \midrule
        & \multicolumn{10}{c}{\textit{Training-time regularization methods}} \\
        % MOS \cite{DBLP:conf/cvpr/HuangL21} & 78.72{\scriptsize$\pm$5.86} & 71.45{\scriptsize$\pm$3.09} & 76.41{\scriptsize$\pm$5.93} & 62.90{\scriptsize$\pm$6.62} & 94.83{\scriptsize$\pm$0.37} & 56.05{\scriptsize$\pm$1.01} & 80.40{\scriptsize$\pm$0.18} & 57.28{\scriptsize$\pm$3.29} & 80.17{\scriptsize$\pm$1.21} & 76.98{\scriptsize$\pm$0.20} \\
        % VOS \cite{DBLP:conf/iclr/DuWCL22} & 57.03{\scriptsize$\pm$1.92} & 87.70{\scriptsize$\pm$0.48} & 40.43{\scriptsize$\pm$4.53} & 90.83{\scriptsize$\pm$0.92} & 94.31{\scriptsize$\pm$0.64} & \underline{55.56{\scriptsize$\pm$0.77}} & \underline{80.93{\scriptsize$\pm$0.29}} & 53.70{\scriptsize$\pm$0.38} & 81.32{\scriptsize$\pm$0.09} & 77.20{\scriptsize$\pm$0.10} \\
        LogitNorm \cite{DBLP:conf/icml/WeiXCF0L22} & \underline{29.34{\scriptsize$\pm$0.81}} & \underline{92.33{\scriptsize$\pm$0.08}} & \underline{13.81{\scriptsize$\pm$0.20}} & \underline{96.74{\scriptsize$\pm$0.06}} & 94.30{\scriptsize$\pm$0.25} & 62.89{\scriptsize$\pm$0.57} & 78.47{\scriptsize$\pm$0.31} & 53.61{\scriptsize$\pm$3.45} & 81.53{\scriptsize$\pm$1.26} & 76.34{\scriptsize$\pm$0.17} \\
        CIDER \cite{DBLP:conf/iclr/MingSD023} & 32.11{\scriptsize$\pm$0.94} & 90.71{\scriptsize$\pm$0.16} & 20.72{\scriptsize$\pm$0.85} & \underline{94.71{\scriptsize$\pm$0.36}} & -- & 72.02{\scriptsize$\pm$0.31} & 73.10{\scriptsize$\pm$0.39} & 54.22{\scriptsize$\pm$1.24} & 80.49{\scriptsize$\pm$0.68} & -- \\
        UMAP \cite{DBLP:conf/icml/ZhuLYLX023} & 33.12{\scriptsize$\pm$0.06} & 91.00{\scriptsize$\pm$0.07} & 21.70{\scriptsize$\pm$1.57} & 94.20{\scriptsize$\pm$0.36} & \underline{95.06{\scriptsize$\pm$0.30}} & 59.71{\scriptsize$\pm$0.65} & 79.49{\scriptsize$\pm$0.23} & \underline{52.11{\scriptsize$\pm$2.36}} & 81.62{\scriptsize$\pm$1.37} & \underline{77.25{\scriptsize$\pm$0.10}} \\
        SNN \cite{DBLP:conf/aaai/GhosalSL24} & 37.21{\scriptsize$\pm$0.70} & 90.25{\scriptsize$\pm$0.09} & 26.05{\scriptsize$\pm$2.34} & 92.49{\scriptsize$\pm$0.78} & \textbf{95.11{\scriptsize$\pm$0.13}} & 60.32{\scriptsize$\pm$1.44} & 80.33{\scriptsize$\pm$0.22} & 53.52{\scriptsize$\pm$1.77} & \underline{82.17{\scriptsize$\pm$0.69}} & \underline{77.56{\scriptsize$\pm$0.27}} \\
        T2FNorm \cite{DBLP:conf/cvpr/RegmiPDGSB22} & \textbf{26.47{\scriptsize$\pm$0.35}} & \textbf{92.79{\scriptsize$\pm$0.13}} & \textbf{12.75{\scriptsize$\pm$0.73}} & \textbf{96.98{\scriptsize$\pm$0.23}} & 94.69{\scriptsize$\pm$0.07} & 58.47{\scriptsize$\pm$1.35} & 79.84{\scriptsize$\pm$0.40} & \underline{51.25{\scriptsize$\pm$2.52}} & \textbf{82.73{\scriptsize$\pm$1.01}} & 76.43{\scriptsize$\pm$0.13} \\
        \midrule
        \textbf{SPCP} & \underline{31.67{\scriptsize$\pm$0.25}} & \underline{91.57{\scriptsize$\pm$0.11}} & \underline{20.54{\scriptsize$\pm$0.67}} & 94.55{\scriptsize$\pm$0.42} & \underline{94.91{\scriptsize$\pm$0.27}} & \underline{55.21{\scriptsize$\pm$0.85}} & \textbf{81.25{\scriptsize$\pm$0.19}} & \textbf{50.59{\scriptsize$\pm$1.83}} & \underline{82.17{\scriptsize$\pm$1.32}} & \textbf{77.70{\scriptsize$\pm$0.20}} \\
        \bottomrule[1.5pt]
        \end{tabular}
    }
    \caption{Comparison on CIFAR benchmarks. All values are percentages, and OOD detection results are averaged over multiple OOD datasets. Detailed results for each OOD dataset are provided in Appendix D. The best results are in \textbf{bold}, with the second and third best results \underline{underlined}.}
    \label{tab:cifar}
    \end{center}
\end{table*}

\begin{table*}[t]
    % \vspace{-1mm}
    % \vspace{-3mm}
    \small
    \begin{center}
    \resizebox{0.75\textwidth}{!}{
        \begin{tabular}{@{\hspace{0.10cm}}l|@{\hspace{0.45cm}}c@{\hspace{0.4cm}}c@{\hspace{0.65cm}}c@{\hspace{0.4cm}}c@{\hspace{0.65cm}}c@{\hspace{0.10cm}}}
        \toprule[1.5pt]
        \multirow{2}{*}{\textbf{Method}} & \multicolumn{2}{@{\hspace{0.50cm}}c@{\hspace{0.50cm}}}{\textbf{Near-OOD}} & \multicolumn{2}{@{\hspace{0.50cm}}c@{\hspace{0.50cm}}}{\textbf{Far-OOD}} & \multirow{2}{*}{ID ACC$\uparrow$} \\
        & FPR95$\downarrow$ & AUROC$\uparrow$ & FPR95$\downarrow$ & AUROC$\uparrow$ \\
        \midrule
        & \multicolumn{5}{c}{\textit{Post-hoc methods (vanilla training with cross-entropy)}} \\
        MSP \cite{DBLP:conf/iclr/HendrycksG17} & \textbf{54.82{\scriptsize$\pm$0.35}} & \underline{83.34{\scriptsize$\pm$0.06}} & 35.43{\scriptsize$\pm$0.38} & 90.13{\scriptsize$\pm$0.09} & 86.37{\scriptsize$\pm$0.08} \\
        Energy \cite{DBLP:conf/nips/LiuWOL20} & 60.24{\scriptsize$\pm$0.57} & 82.50{\scriptsize$\pm$0.05} & 34.86{\scriptsize$\pm$1.30} & 90.86{\scriptsize$\pm$0.21} & 86.37{\scriptsize$\pm$0.08} \\
        % VIM \cite{DBLP:conf/cvpr/Wang0F022} & 59.19{\scriptsize$\pm$0.71} & 78.68{\scriptsize$\pm$0.24} & 27.20{\scriptsize$\pm$0.30} & 91.26{\scriptsize$\pm$0.19} & 86.37{\scriptsize$\pm$0.08} \\
        DICE \cite{DBLP:conf/eccv/SunL22a} & 61.88{\scriptsize$\pm$0.67} & 81.78{\scriptsize$\pm$0.14} & 36.51{\scriptsize$\pm$1.18} & 90.80{\scriptsize$\pm$0.31} & 86.37{\scriptsize$\pm$0.08} \\
        ReAct \cite{DBLP:conf/nips/SunGL21} & 62.49{\scriptsize$\pm$2.19} & 81.87{\scriptsize$\pm$0.98} & 28.50{\scriptsize$\pm$0.95} & 92.31{\scriptsize$\pm$0.56} & 86.37{\scriptsize$\pm$0.08} \\
        ASH \cite{DBLP:conf/iclr/DjurisicBAL23} & 64.89{\scriptsize$\pm$0.90} & 82.38{\scriptsize$\pm$0.19} & 27.29{\scriptsize$\pm$1.12} & \underline{93.90{\scriptsize$\pm$0.27}} & 86.37{\scriptsize$\pm$0.08} \\
        SCALE \cite{DBLP:conf/iclr/XuCFY24} & 57.29{\scriptsize$\pm$0.90} & \textbf{84.84{\scriptsize$\pm$0.28}} & 26.46{\scriptsize$\pm$0.81} & \underline{93.98{\scriptsize$\pm$0.25}} & 86.37{\scriptsize$\pm$0.08} \\
        \midrule
        & \multicolumn{5}{c}{\textit{Training-time regularization methods}} \\
        % MOS \cite{DBLP:conf/cvpr/HuangL21} & 71.60{\scriptsize$\pm$0.48} & 69.84{\scriptsize$\pm$0.46} & 51.56{\scriptsize$\pm$0.42} & 80.46{\scriptsize$\pm$0.92} & 85.60{\scriptsize$\pm$0.20} \\
        % VOS \cite{DBLP:conf/iclr/DuWCL22} & 59.89{\scriptsize$\pm$0.47} & 82.51{\scriptsize$\pm$0.11} & 34.01{\scriptsize$\pm$0.97} & 91.00{\scriptsize$\pm$0.28} & 86.23{\scriptsize$\pm$0.19} \\
        LogitNorm \cite{DBLP:conf/icml/WeiXCF0L22} & 56.46{\scriptsize$\pm$0.37} & 82.66{\scriptsize$\pm$0.15} & \underline{26.11{\scriptsize$\pm$0.52}} & 93.04{\scriptsize$\pm$0.21} & 86.04{\scriptsize$\pm$0.15} \\
        CIDER \cite{DBLP:conf/iclr/MingSD023} & 60.10{\scriptsize$\pm$0.73} & 80.58{\scriptsize$\pm$1.75} & 30.17{\scriptsize$\pm$2.75} & 90.66{\scriptsize$\pm$1.68} & -- \\
        UMAP \cite{DBLP:conf/icml/ZhuLYLX023} & 60.81{\scriptsize$\pm$0.84} & 81.08{\scriptsize$\pm$0.39} & 32.47{\scriptsize$\pm$0.67} & 91.62{\scriptsize$\pm$0.29} & 86.37{\scriptsize$\pm$0.08} \\
        SNN \cite{DBLP:conf/aaai/GhosalSL24} & 59.85{\scriptsize$\pm$0.46} & 81.33{\scriptsize$\pm$0.19} & 28.04{\scriptsize$\pm$0.64} & 92.28{\scriptsize$\pm$0.21} & \underline{86.56{\scriptsize$\pm$0.03}} \\
        T2FNorm \cite{DBLP:conf/cvpr/RegmiPDGSB22} & \underline{55.01{\scriptsize$\pm$0.36}} & 83.00{\scriptsize$\pm$0.07} & \underline{25.37{\scriptsize$\pm$0.55}} & 93.55{\scriptsize$\pm$0.17} & \textbf{86.87{\scriptsize$\pm$0.19}} \\
        \midrule
        \textbf{SPCP} & 59.77{\scriptsize$\pm$0.25} & 82.52{\scriptsize$\pm$0.03} & 30.43{\scriptsize$\pm$0.45} & 91.46{\scriptsize$\pm$0.14} & \underline{86.59{\scriptsize$\pm$0.10}} \\
        \textbf{LogitNorm+SPCP} & \underline{55.33{\scriptsize$\pm$0.45}} & \underline{83.20{\scriptsize$\pm$0.07}} & \textbf{21.95{\scriptsize$\pm$0.73}} & \textbf{94.11{\scriptsize$\pm$0.28}} & 86.37{\scriptsize$\pm$0.09} \\
        \bottomrule[1.5pt]
        \end{tabular}
    }
    \caption{Comparison on large-scale ImageNet benchmark.}
    \label{tab:imagenet}
    \end{center}
\end{table*}

\section{Experiments}

% In this section, we first describe our experimental setup, then present the main results on the authoritative benchmarks, followed by ablation studies and further analysis.

\subsection{Experimental Setup}
\label{sec:setup}
Our evaluation is based on the standard practice of the OpenOOD v1.5 benchmark \cite{DBLP:journals/corr/abs-2306-09301}, which includes both small-scale CIFAR benchmarks and large-scale ImageNet benchmark, spanning near-OOD scenarios with semantic shifts and far-OOD scenarios with further obvious covariance shifts \footnote{Code is available at \url{https://github.com/njustkmg/AAAI2026-SPCP}}.

\noindent\textbf{Datasets.} 
The setup for the small-scale experiment uses CIFAR-10/100 \cite{Krizhevsky_2009} as the ID dataset. Evaluations cover near-OOD group, including CIFAR-100/10 and TinyImageNet (TIN) \cite{le2015tiny}, as well as far-OOD group, including MNIST \cite{DBLP:journals/spm/Deng12}, SVHN \cite{Netzer_Wang_Coates_Bissacco_Wu_Ng_2011}, Textures \cite{DBLP:conf/cvpr/CimpoiMKMV14}, and Places365 \cite{DBLP:journals/pami/ZhouLKO018}.

For the large-scale experimental setup, a 200-class subset of ImageNet-1K \cite{DBLP:conf/cvpr/DengDSLL009}, referred to as ImageNet-200, is adopted as the ID dataset. The near-OOD group includes SSB-hard \cite{DBLP:conf/iclr/Vaze0VZ22} and NINCO \cite{DBLP:conf/icml/BitterwolfM023}, while the far-OOD group contains iNaturalist \cite{DBLP:conf/cvpr/HornASCSSAPB18}, Textures \cite{DBLP:conf/cvpr/CimpoiMKMV14}, and OpenImage-O \cite{DBLP:conf/cvpr/Wang0F022}.

\noindent\textbf{Evaluation Metrics.} 
We report the following three widely adopted metrics for comparison: (1) FPR95, the false positive rate of OOD data at a 95\% true positive rate of ID data; (2) AUROC, the area under the receiver operating characteristic curve; and (3) ID ACC, the classification accuracy on the ID test set. The evaluation adopts the same number of independent runs and random seed settings as in OpenOOD \cite{DBLP:journals/corr/abs-2306-09301}, with results reported as the mean and standard deviation.

\noindent\textbf{Baselines.} 
We compare SPCP with a broad range of competitive baselines: (1) post-hoc methods, including OOD scoring methods: MSP \cite{DBLP:conf/iclr/HendrycksG17}, Energy \cite{DBLP:conf/nips/LiuWOL20}; and network adjustment methods: DICE \cite{DBLP:conf/eccv/SunL22a}, ReAct \cite{DBLP:conf/nips/SunGL21}, ASH \cite{DBLP:conf/iclr/DjurisicBAL23}, SCALE \cite{DBLP:conf/iclr/XuCFY24}; and (2) training-time regularization methods: LogitNorm \cite{DBLP:conf/icml/WeiXCF0L22}, CIDER \cite{DBLP:conf/iclr/MingSD023}, UMAP \cite{DBLP:conf/icml/ZhuLYLX023}, SNN \cite{DBLP:conf/aaai/GhosalSL24} and T2FNorm \cite{DBLP:conf/cvpr/RegmiPDGSB22}. Baseline results are sourced from the authoritative OpenOOD leaderboard \cite{DBLP:journals/corr/abs-2306-09301}.

\noindent\textbf{Implementation Details.} 
Our implementation strictly follows the OpenOOD benchmark \cite{DBLP:journals/corr/abs-2306-09301}. Specifically, ResNet-18 \cite{DBLP:conf/cvpr/HeZRS16} is employed as the backbone architecture. The models are trained for 100 epochs using stochastic gradient descent (SGD) with a learning rate of 0.1, following a cosine annealing decay schedule \cite{DBLP:conf/iclr/LoshchilovH17}, a momentum of 0.9, and a weight decay of $5 \times 10^{-4}$. The batch size is set to 128 for the CIFAR experiment and 256 for the ImageNet experiment. More details are available in Appendix C.
% Details on hyperparameter selection and additional implementation specifics are provided in the Appendix C.

% Hyperparameters are selected via a grid search using the ID and OOD validation sets from OpenOOD, with the configuration that delivers the best AUROC chosen for the final evaluation. More implementation details are provided in Appendix.

\begin{table*}[t]
    % \vspace{-2mm}
    % \vspace{-1mm}
    \centering
    \resizebox{0.58\textwidth}{!}{
        \begin{tabular}{@{\hspace{0.1cm}}c@{\hspace{0.25cm}}c|@{\hspace{0.25cm}}c@{\hspace{0.25cm}}c@{\hspace{0.4cm}}c@{\hspace{0.25cm}}c@{\hspace{0.4cm}}c@{\hspace{0.1cm}}}
        \toprule[1.5pt]
        \multicolumn{2}{@{\hspace{0.25cm}}c|@{\hspace{0.25cm}}}{\textbf{Stage}} & \multicolumn{2}{@{\hspace{0.25cm}}c@{\hspace{0.25cm}}}{\textbf{Near-OOD}} & \multicolumn{2}{@{\hspace{0.25cm}}c@{\hspace{0.25cm}}}{\textbf{Far-OOD}} & \multirow{2}{*}{ID ACC$\uparrow$} \\
        Training & Inferring & FPR95$\downarrow$ & AUROC$\uparrow$ & FPR95$\downarrow$ & AUROC$\uparrow$ & \\ 
        \midrule
        $\times$ & $\times$ & 55.62{\scriptsize$\pm$0.61} & 80.91{\scriptsize$\pm$0.08} & 56.59{\scriptsize$\pm$1.38} & 79.77{\scriptsize$\pm$0.61} & 77.25{\scriptsize$\pm$0.10} \\ 
        $\checkmark$ & $\times$ & 55.49{\scriptsize$\pm$0.77} & 81.09{\scriptsize$\pm$0.21} & 51.96{\scriptsize$\pm$2.25} & 81.71{\scriptsize$\pm$1.45} & \textbf{77.70{\scriptsize$\pm$0.24}} \\ 
        $\times$ & $\checkmark$ & 60.28{\scriptsize$\pm$0.46} & 79.50{\scriptsize$\pm$0.08} & 53.66{\scriptsize$\pm$3.13} & 80.59{\scriptsize$\pm$1.10} & 76.94{\scriptsize$\pm$0.12} \\ 
        $\checkmark$ & $\checkmark$ & \textbf{55.21{\scriptsize$\pm$0.85}} & \textbf{81.25{\scriptsize$\pm$0.19}} & \textbf{50.59{\scriptsize$\pm$1.83}} & \textbf{82.17{\scriptsize$\pm$1.32}} & \textbf{77.70{\scriptsize$\pm$0.20}} \\
        \bottomrule[1.5pt]
        \end{tabular}
    }
    \caption{Ablation study of contribution truncation in different stages on the CIFAR-100 benchmark.}
    \label{tab:ablation}
\end{table*}

\begin{figure*}[t]
    % \vspace{-3mm}
    \centering
    \begin{subfigure}[t]{0.235\textwidth}
        \includegraphics[width=\linewidth]{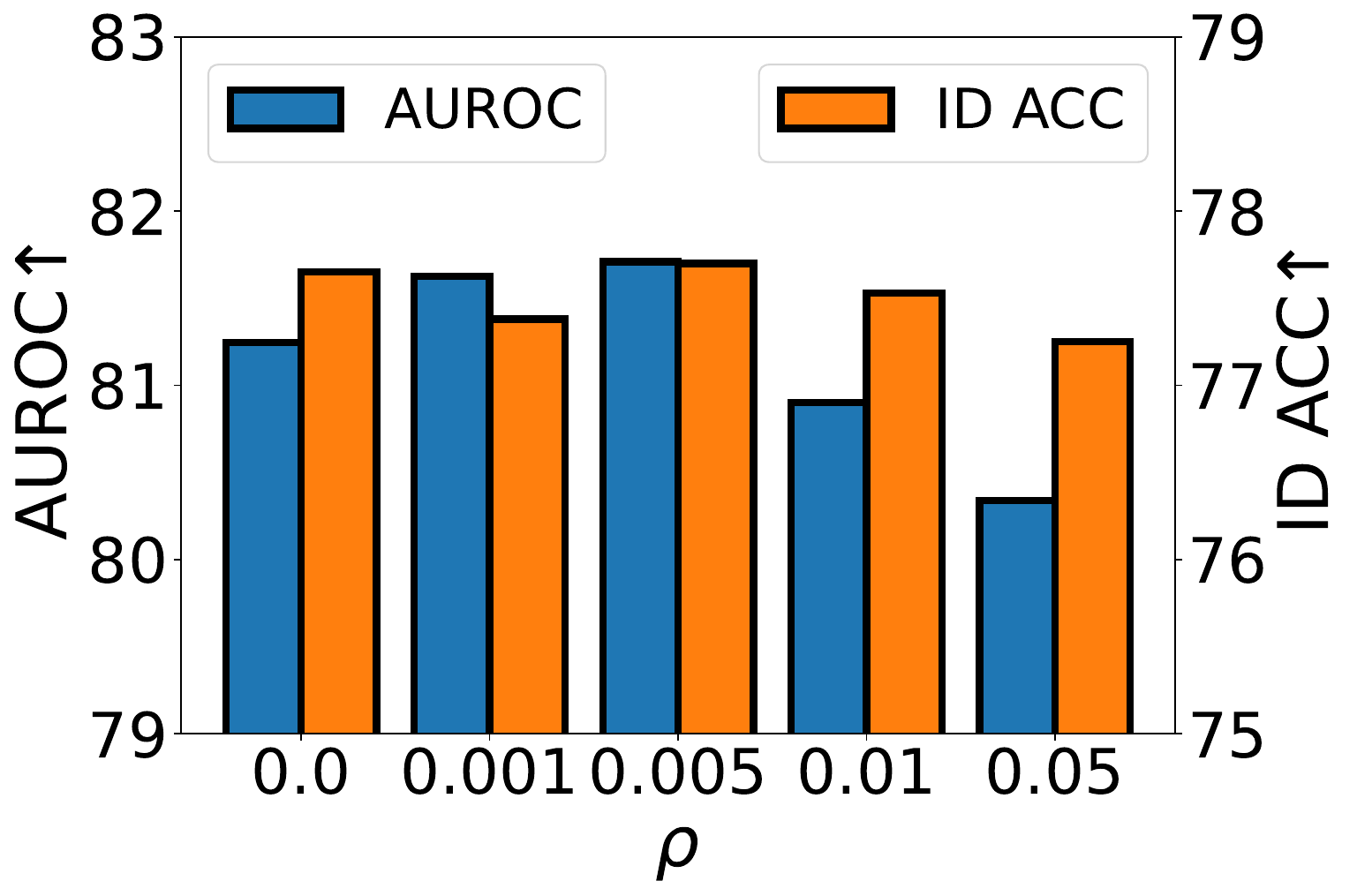}
        \caption{Diff. Percentile for $\rho$}
        \label{fig:parameter_a}
    \end{subfigure}
    \hspace{3mm}
    \begin{subfigure}[t]{0.235\textwidth}
        \includegraphics[width=\linewidth]{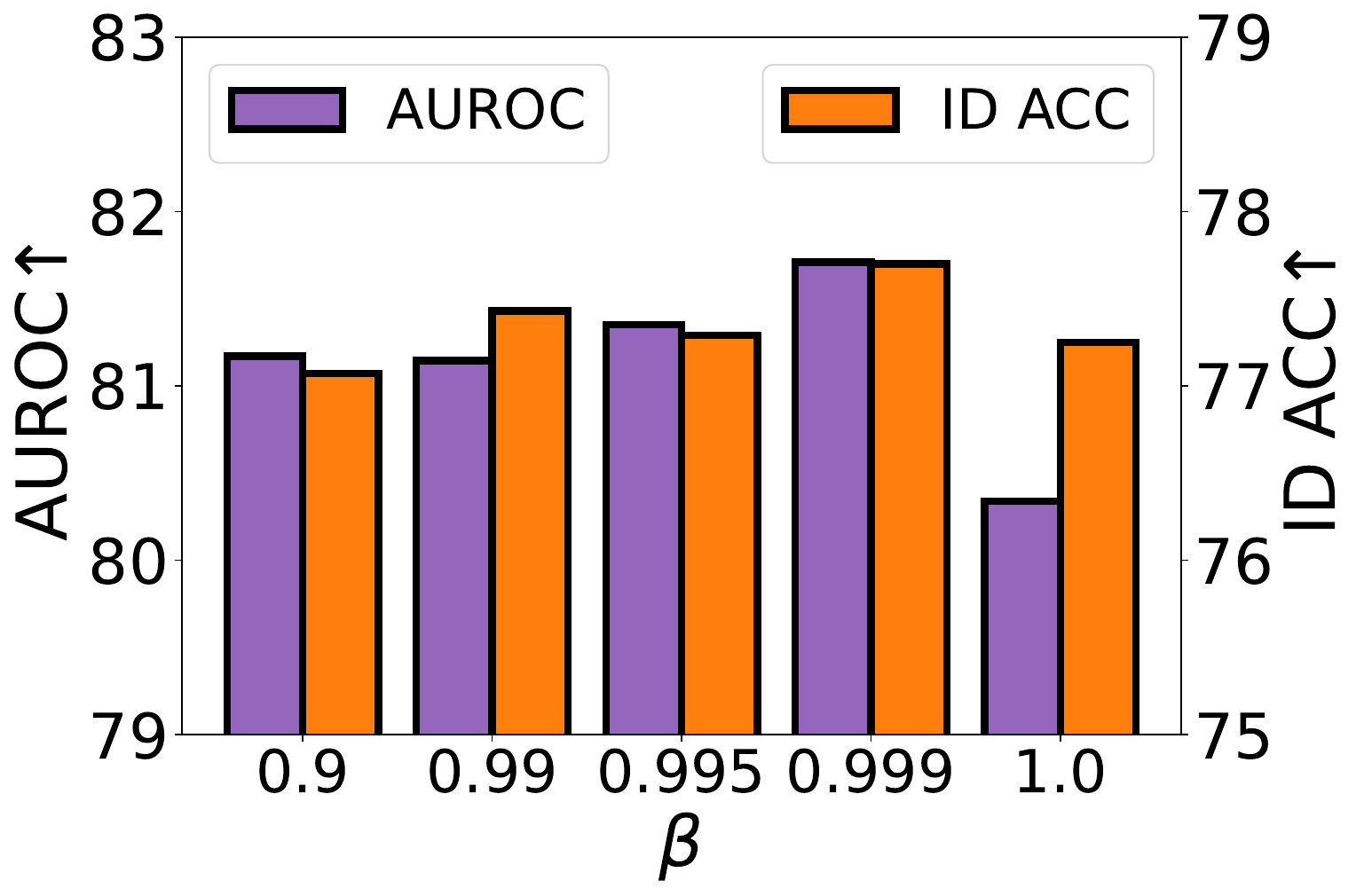}
        \caption{Diff. EMA Factor $\beta$}
        \label{fig:parameter_b}
    \end{subfigure}
    \hspace{3mm}
    \begin{subfigure}[t]{0.235\textwidth}
        \includegraphics[width=\linewidth]{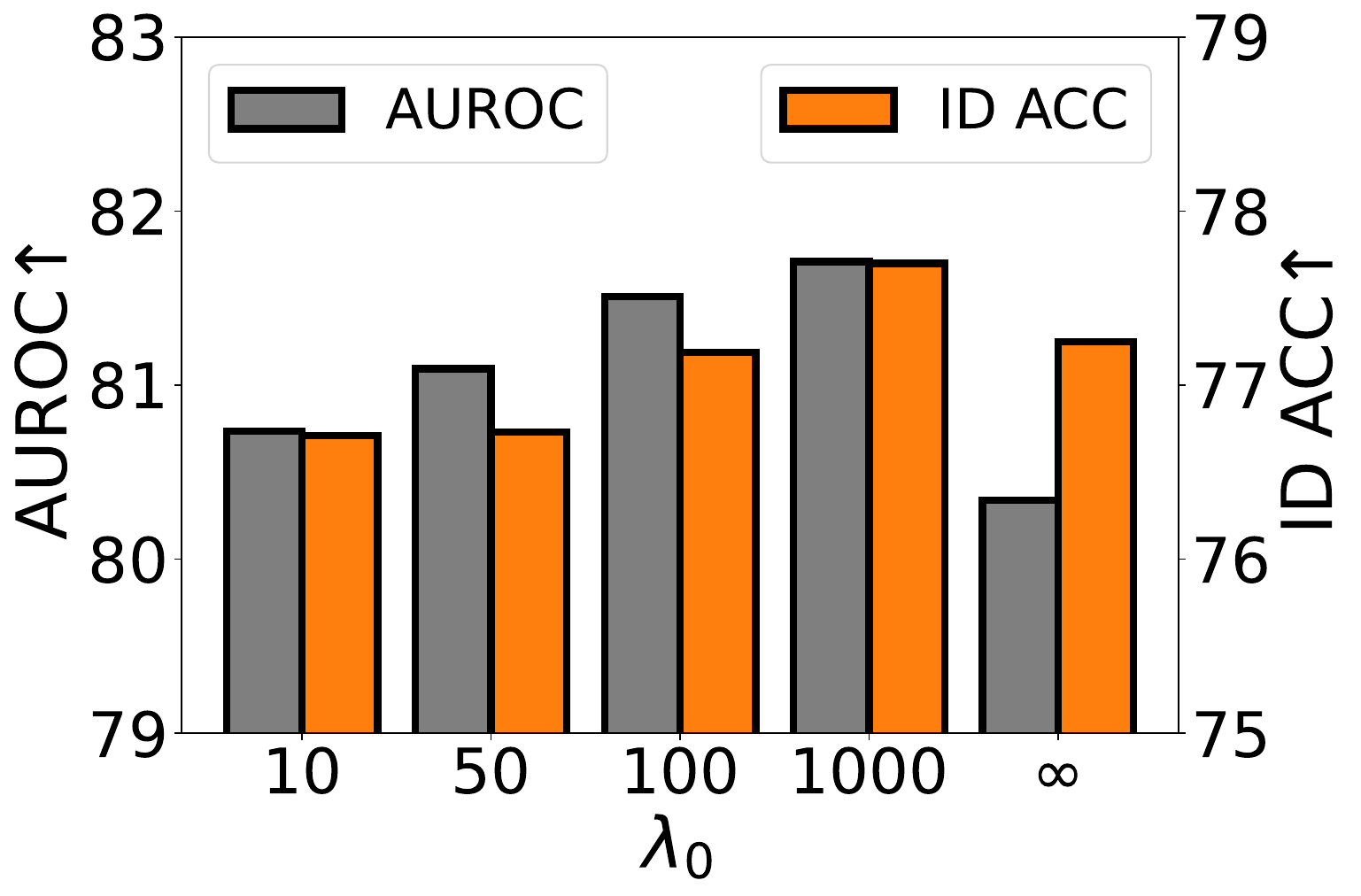}
        \caption{Diff. Initial Threshold $\lambda_0$}
        \label{fig:parameter_c}
    \end{subfigure}
    % \vspace{-3mm}
    \caption{Ablation studies on the hyperparameters: (a) effect of varying the percentile $\rho$ used for threshold estimation; (b) impact of the EMA smoothing factor $\beta$; (c) influence of initial threshold $\lambda_0$. The OOD results are averaged over both near- and far-OOD groups on the CIFAR-100 benchmark.}
    \label{fig:parameter}
\end{figure*}

\begin{table*}[t]
    % \vspace{-1mm}
    % \vspace{-4mm}
    \small
    \begin{center}
    \resizebox{0.64\textwidth}{!}{
        \begin{tabular}{@{\hspace{0.10cm}}c|@{\hspace{0.2cm}}l|@{\hspace{0.2cm}}c@{\hspace{0.2cm}}c@{\hspace{0.4cm}}c@{\hspace{0.2cm}}c@{\hspace{0.4cm}}c@{\hspace{0.10cm}}}
        \toprule[1.5pt]
        \multirow{2}{*}{\textbf{Model}} & \multirow{2}{*}{\textbf{Method}} & \multicolumn{2}{@{\hspace{0.3cm}}c@{\hspace{0.3cm}}}{\textbf{Near-OOD}} & \multicolumn{2}{@{\hspace{0.3cm}}c@{\hspace{0.3cm}}}{\textbf{Far-OOD}} & \multirow{2}{*}{ID ACC$\uparrow$} \\
        & & FPR95$\downarrow$ & AUROC$\uparrow$ & FPR95$\downarrow$ & AUROC$\uparrow$ \\
        \midrule
        \multirow{2}{*}{ResNet-18} & Vanilla & 55.62{\scriptsize$\pm$0.61} & 80.91{\scriptsize$\pm$0.08} & 56.59{\scriptsize$\pm$1.38} & 79.77{\scriptsize$\pm$0.61} & 77.25{\scriptsize$\pm$0.10} \\
        & \textbf{SPCP} & \textbf{55.21{\scriptsize$\pm$0.85}} & \textbf{81.25{\scriptsize$\pm$0.19}} & \textbf{50.59{\scriptsize$\pm$1.83}} & \textbf{82.17{\scriptsize$\pm$1.32}} & \textbf{77.70{\scriptsize$\pm$0.20}} \\
        \midrule
        \multirow{2}{*}{WideResNet-28-10} & Vanilla & 54.48{\scriptsize$\pm$1.30} & 82.31{\scriptsize$\pm$0.41} & 55.34{\scriptsize$\pm$1.67} & 80.87{\scriptsize$\pm$0.80} & 80.53{\scriptsize$\pm$0.13} \\
        & \textbf{SPCP} & \textbf{53.48{\scriptsize$\pm$0.52}} & \textbf{82.63{\scriptsize$\pm$0.24}} & \textbf{52.84{\scriptsize$\pm$0.93}} & \textbf{81.69{\scriptsize$\pm$0.31}} & \textbf{80.56{\scriptsize$\pm$0.22}} \\
        \midrule
        \multirow{2}{*}{DenseNet-101} & Vanilla & 61.57{\scriptsize$\pm$1.36} & 79.27{\scriptsize$\pm$0.36} & 65.49{\scriptsize$\pm$2.45} & 76.16{\scriptsize$\pm$0.81} & 76.54{\scriptsize$\pm$0.42} \\
        & \textbf{SPCP} & \textbf{58.93{\scriptsize$\pm$1.14}} & \textbf{80.28{\scriptsize$\pm$0.48}} & \textbf{56.75{\scriptsize$\pm$2.62}} & \textbf{78.68{\scriptsize$\pm$1.62}} & \textbf{76.86{\scriptsize$\pm$0.29}} \\
        \bottomrule[1.5pt]
        \end{tabular}
    }
    \caption{Generalization to different backbones on the CIFAR-100 benchmark.}
    \label{tab:backbone}
    \end{center}
\end{table*}

\begin{table*}[t]
    % \vspace{-2.5mm}
    % \vspace{-3mm}
    \small
    \begin{center}
    \resizebox{0.715\textwidth}{!}{
        \begin{tabular}{@{\hspace{0.15cm}}l|@{\hspace{0.2cm}}c@{\hspace{0.2cm}}c@{\hspace{0.4cm}}c@{\hspace{0.2cm}}c@{\hspace{0.4cm}}c@{\hspace{0.15cm}}}
        \toprule[1.5pt]
        \multirow{2}{*}{\textbf{Method}} & \multicolumn{2}{@{\hspace{0.2cm}}c@{\hspace{0.2cm}}}{\textbf{Near-OOD}} & \multicolumn{2}{@{\hspace{0.2cm}}c@{\hspace{0.2cm}}}{\textbf{Far-OOD}} & \multirow{2}{*}{ID ACC$\uparrow$} \\
        & FPR95$\downarrow$ & AUROC$\uparrow$ & FPR95$\downarrow$ & AUROC$\uparrow$ \\
        \midrule
        MSP \cite{DBLP:conf/iclr/HendrycksG17} & \textbf{54.80{\scriptsize$\pm$0.33}} & 80.27{\scriptsize$\pm$0.11} & 58.70{\scriptsize$\pm$1.06} & 77.76{\scriptsize$\pm$0.44} & 77.25{\scriptsize$\pm$0.10} \\
        \textbf{MSP+SPCP} & 55.04{\scriptsize$\pm$0.68} & \textbf{80.41{\scriptsize$\pm$0.28}} & \textbf{54.31{\scriptsize$\pm$0.95}} & \textbf{80.22{\scriptsize$\pm$0.93}} & \textbf{77.70{\scriptsize$\pm$0.20}} \\
        \midrule
        Energy \cite{DBLP:conf/nips/LiuWOL20} & 55.62{\scriptsize$\pm$0.61} & 80.91{\scriptsize$\pm$0.08} & 56.59{\scriptsize$\pm$1.38} & 79.77{\scriptsize$\pm$0.61} & 77.25{\scriptsize$\pm$0.10} \\ 
        \textbf{Energy+SPCP} & \textbf{55.21{\scriptsize$\pm$0.85}} & \textbf{81.25{\scriptsize$\pm$0.19}} & \textbf{50.59{\scriptsize$\pm$1.83}} & \textbf{82.17{\scriptsize$\pm$1.32}} & \textbf{77.70{\scriptsize$\pm$0.20}} \\
        \midrule
        GradNorm \cite{DBLP:conf/nips/HuangGL21} & 85.58{\scriptsize$\pm$0.46} & 70.13{\scriptsize$\pm$0.47} & 83.68{\scriptsize$\pm$1.92} & 69.14{\scriptsize$\pm$1.05} & 77.25{\scriptsize$\pm$0.10} \\ 
        \textbf{GradNorm+SPCP} & \textbf{80.33{\scriptsize$\pm$1.28}} & \textbf{73.71{\scriptsize$\pm$0.71}} & \textbf{79.47{\scriptsize$\pm$2.60}} & \textbf{73.70{\scriptsize$\pm$1.69}} & \textbf{77.70{\scriptsize$\pm$0.20}} \\
        \midrule
        KNN \cite{DBLP:conf/icml/SunM0L22} & 61.22{\scriptsize$\pm$0.14} & 80.18{\scriptsize$\pm$0.15} & 53.65{\scriptsize$\pm$0.28} & 82.40{\scriptsize$\pm$0.17} & 77.25{\scriptsize$\pm$0.10} \\
        \textbf{KNN+SPCP} & \textbf{59.13{\scriptsize$\pm$0.46}} & \textbf{80.55{\scriptsize$\pm$0.07}} & \textbf{51.97{\scriptsize$\pm$0.97}} & \textbf{82.52{\scriptsize$\pm$0.79}} & \textbf{77.70{\scriptsize$\pm$0.20}} \\
        \midrule
        ASH \cite{DBLP:conf/iclr/DjurisicBAL23} & 65.71{\scriptsize$\pm$0.24} & 78.20{\scriptsize$\pm$0.15} & 59.20{\scriptsize$\pm$2.46} & 80.58{\scriptsize$\pm$0.66} & 77.25{\scriptsize$\pm$0.10} \\
        \textbf{ASH+SPCP} & \textbf{61.15{\scriptsize$\pm$0.18}} & \textbf{79.75{\scriptsize$\pm$0.13}} & \textbf{56.06{\scriptsize$\pm$1.33}} & \textbf{82.07{\scriptsize$\pm$0.29}} & \textbf{77.70{\scriptsize$\pm$0.20}} \\
        \midrule
        LogitNorm \cite{DBLP:conf/icml/WeiXCF0L22} & 62.89{\scriptsize$\pm$0.57} & 78.47{\scriptsize$\pm$0.31} & 53.61{\scriptsize$\pm$3.45} & 81.53{\scriptsize$\pm$1.26} & \textbf{76.34{\scriptsize$\pm$0.17}} \\
        \textbf{LogitNorm+SPCP} & \textbf{60.70{\scriptsize$\pm$0.77}} & \textbf{79.20{\scriptsize$\pm$0.31}} & \textbf{47.81{\scriptsize$\pm$1.16}} & \textbf{83.99{\scriptsize$\pm$0.66}} & 75.94{\scriptsize$\pm$0.20} \\
        \bottomrule[1.5pt]
        \end{tabular}
    }
    \caption{Compatibility with other OOD detection methods on the CIFAR-100 benchmark.}
    \label{tab:other}
    \end{center}
\end{table*}

\subsection{Main Results}
In this section, we report the performance of SPCP on the CIFAR benchmarks as well as on the more realistic and challenging ImageNet benchmark. Specifically, Tables \ref{tab:vanilla_cifar} and \ref{tab:vanilla_imagenet} provide a fine-grained comparison of our SPCP with vanilla training, while Tables \ref{tab:cifar} and \ref{tab:imagenet} show comparisons with other competitive methods. The results reveal that: (1) SPCP boosts OOD detection in nearly all cases while preserving ID performance. On the CIFAR-10 benchmark, SPCP outperforms the baseline by a large margin, reducing the average FPR95 by 29.67\% in near-OOD settings and 21.25\% in far-OOD settings. (2) In comparison to post-hoc methods, SPCP offers notable improvements by establishing more robust contribution patterns for OOD detection. Moreover, SPCP is complementary to various existing post-hoc methods to push their performance further, as will be discussed in Section \ref{sec: further}. (3) Compared to a suite of training-time regularization methods, SPCP delivers competitive OOD detection performance and achieves promising results. It is noteworthy that no single method emerges as the definitive winner on the authoritative OpenOOD benchmark \cite{DBLP:journals/corr/abs-2306-09301}. Our SPCP achieves top or near-top performance in most OOD settings, indicating the effectiveness of our proposed shaping contribution pattern strategy.

\subsection{Ablation Study}
\noindent\textbf{Effects of Truncating Contributions at Different Stages.} Table \ref{tab:ablation} presents an ablation study on the impact of truncating contributions during training and inference. The results indicate that: (1) Imposing contribution truncation during training is more effective than doing so only during inference, highlighting the critical role of shaping contribution patterns in the training process. (2) Applying contribution truncation exclusively during inference helps mitigate overconfidence and leads to modest improvements in far-OOD scenarios. However, the absence of corresponding regularization during training can hurt performance in near-OOD settings and cause a slight degradation in ID task performance. (3) SPCP achieves contribution truncation in both the training and inference processes, which preserves pattern consistency and further drives performance improvements.

\noindent\textbf{Effect of the Percentile $\rho$.}
In Figure \ref{fig:parameter_a}, we investigate the impact of varying the percentile $\rho$, which is used to determine the parameter contribution threshold $\lambda$ in Eq. \eqref{equ: p}. Compared to the baseline (\emph{i.e.}, $\rho=0$), properly constraining the upper bound of parameter contributions with a suitable $\rho$ improves OOD detection while maintaining generalization on ID tasks. However, an inappropriately small $\rho$ may cause the model to rely excessively on numerous and overlapping parameter decisions, leading to inter-class conflicts and degraded performance. 
% Empirical guidance on the selection of parameter $\rho$ is provided in Appendix C.

\noindent\textbf{Effect of the EMA Factor $\beta$.} 
Figure \ref{fig:parameter_b} presents an analysis of the effect of the EMA smoothing factor as defined in Eq. \eqref{equ: p}. As shown, performance degrades when the dynamic updating strategy is disabled (\emph{i.e.}, $\beta = 1.0$), highlighting the critical role of dynamically adjusting the threshold during training. However, a smaller $\beta$ may lead to suboptimal performance due to unstable threshold estimation from rapid updates. Therefore, striking a moderate balance is more beneficial for performance improvement. 
% We fix $\beta = 0.999$ in all main experiments.

\noindent\textbf{Effect of the Initial Threshold $\lambda_0$.}
Figure \ref{fig:parameter_c} explores the impact of different initial threshold values set for $\lambda_0$ in Eq. \eqref{equ: p}. The results indicate that a smaller initial value of $\lambda_0$ leads to excessive intervention before sufficient learning has occurred, thereby disrupting the early learning process. In contrast, a moderately large $\lambda_0$ serves as a form of warm-up and leads to performance improvements. Consequently, selecting a suitably large $\lambda_0$ is essential for facilitating effective early-stage learning and improving final performance. 
% We fix $\lambda_0 = 1000$ in all main experiments.

\subsection{Further Analysis}
\label{sec: further}
\noindent\textbf{Generalizing to Different Backbones.} In Table \ref{tab:backbone}, we evaluate the generalization of SPCP to different backbones, including the widely used and lightweight ResNet-18 \cite{DBLP:conf/cvpr/HeZRS16}, the high-capacity WideResNet-28-10 \cite{DBLP:conf/bmvc/ZagoruykoK16}, and the densely connected DenseNet-101 \cite{DBLP:conf/cvpr/HuangLMW17}. As shown in the table, our SPCP consistently improves FPR95 and AUROC across both near- and far-OOD scenarios on various backbones, while also preserving ID generalization. These results indicate that SPCP generalizes well across different backbones and effectively enhances OOD detection robustness.

\noindent\textbf{Compatibility with Other OOD Detection Methods.} To validate the compatibility of our SPCP, we integrate representative OOD detection methods spanning diverse information sources: probability space (MSP \cite{DBLP:conf/iclr/HendrycksG17}), logits space (Energy \cite{DBLP:conf/nips/LiuWOL20}), gradient space (GradNorm \cite{DBLP:conf/nips/HuangGL21}), feature space (KNN \cite{DBLP:conf/icml/SunM0L22}), penultimate activation manipulations (ASH \cite{DBLP:conf/iclr/DjurisicBAL23}), and training-time regularization (LogitNorm \cite{DBLP:conf/icml/WeiXCF0L22}). The results in Table \ref{tab:other} show that SPCP consistently improves OOD detection performance across a range of methods, indicating that the contribution patterns shaped by SPCP are broadly applicable and compatible with various algorithmic paradigms.

% \noindent\textbf{Discussion on Relevant Methods.} We now clarify the differences between our SPCP and two representatively related post-hoc methods: DICE \cite{DBLP:conf/eccv/SunL22a} and ReAct \cite{DBLP:conf/nips/SunGL21}. (1) DICE operates by pruning classifier's weight parameters that contribute the least to the prediction, aiming to mitigate the negative impact of redundant parameters in over-parameterized models for OOD detection. Conversely, our SPCP targets the most salient parameters by constraining their upper-bound contributions to promote a more robust contribution pattern. Experimental results demonstrate that shaping a boundary-oriented dense contribution pattern is more effective than the sparsification strategy employed by DICE. (2) ReAct operates by rectifying penultimate activations using a pre-computed upper bound to suppress overconfidence. In contrast, our SPCP goes beyond this by focusing on a more granular level — examining the contribution of individual parameters directly responsible for prediction, which is more impactful for OOD detection. Moreover, ReAct only addresses anomalous activations at test time, while SPCP explicitly guides the model during the training process to develop a boundary-oriented dense contribution pattern, providing substantial gains over simple post-hoc rectification.

\section{Related Work}
\noindent\textbf{Post-hoc Methods} aim to provide suitable measures to indicate the likelihood that a given sample is OOD. Early methods for detecting OOD samples work by scoring the network outputs \cite{DBLP:conf/iclr/LiangLS18,DBLP:conf/nips/LiuWOL20,DBLP:conf/nips/HuangGL21,DBLP:conf/icml/HendrycksBMZKMS22}. For example, MSP \cite{DBLP:conf/iclr/HendrycksG17} uses the maximum SoftMax score as an indicator. Based on the OOD score, various post-hoc network adjustment methods \cite{DBLP:conf/nips/XuLLT23,DBLP:conf/cvpr/AhnPK23,DBLP:conf/aaai/XuY25} are proposed to further enhance OOD score reliability. For example, ReAct \cite{DBLP:conf/nips/SunGL21} and SCALE \cite{DBLP:conf/iclr/XuCFY24} improve ID-OOD separability by rectifying and scaling penultimate-layer activations, respectively. In this work, we reveal that shaping the parameter contribution patterns can help reduce model overconfidence and boost the performance of most existing post-hoc methods.

\noindent\textbf{Training-Time Regularization Methods} aim to provide better discriminative representations for OOD detection by calibrating the model. One category of methods leverages outlier exposure \cite{DBLP:conf/iclr/HendrycksMD19,DBLP:conf/wacv/ZhangILCL23,DBLP:conf/iclr/WangY0DKLH023,DBLP:conf/iclr/JiangCCWW24} to help the model learn more robust decision boundaries. As a representative method, Hendrycks et al. \cite{DBLP:conf/iclr/HendrycksMD19} propose enforcing a uniform predictive distribution for outlier data. Although effective, such methods often hurt ID task performance, and access to outlier data may not always be available. Another category tackles OOD detection by imposing favorable constraints during the training process \cite{DBLP:journals/corr/abs-1802-04865,DBLP:conf/cvpr/HuangL21,DBLP:conf/iclr/DuWCL22,yangstrengthen}. For instance, SNN \cite{DBLP:conf/aaai/GhosalSL24} mitigates the curse-of-dimensionality issue by learning the most relevant subspace. LogitNorm \cite{DBLP:conf/icml/WeiXCF0L22} and T2FNorm \cite{DBLP:conf/cvpr/RegmiPDGSB22} offer a simple fix to cross-entropy loss by decoupling the impact of logits and feature norms, respectively. However, these methods overlook the classifier’s tendency to develop a sparse contribution pattern, which can easily push the model to overconfidence when dominant parameters are spuriously triggered. Our methods alleviate this issue by shaping a more robust contribution pattern for OOD detection while preserving ID performance.

\section{Conclusion}
\label{sec:conclusion}
This paper investigates the underlying factors contributing to the brittleness of OOD detection from the perspective of parameter contribution patterns. We identify that sparse contribution patterns can increase the risk of overconfident predictions and hurt OOD detection. To mitigate this issue, we propose SPCP, a method that constrains the upper bound of parameter contributions during training, thereby promoting the development of dense and bounded contribution patterns. As a result, SPCP enhances the model's resilience to anomalous parameter contributions induced by OOD inputs. Extensive experiments across various OOD scenarios confirm the effectiveness and broad applicability of SPCP.

\section*{Acknowledgements}
This work was supported in part by the National Key R\&D Program of China (2022YFF0712100), NSFC (62276131), the Natural Science Foundation of Jiangsu Province of China under Grant (BK20240081), and the Research on the Teaching Reform of Artificial Intelligence General Education Courses in Jiangsu Undergraduate Universities.

% \noindent\textbf{Limitation and societal impact.} The main limitation of this work lies in the reliance on grid search over a validation set to empirically determine the contribution threshold settings. In future work, we intend to explore adaptive mechanisms for threshold setting to reduce computational resource overhead. This work aims to enhance the reliability of modern deep learning models by reducing false recognition caused by OOD samples, thereby benefiting widespread applications in social life, such as AI for medical and driverless systems. We do not anticipate any negative societal impacts from this work and intend to continue refining and extending our method.

\bibliography{aaai2026}

@inproceedings{DBLP:conf/iclr/FrankleC19,
  author       = {Jonathan Frankle and
                  Michael Carbin},
  title        = {The Lottery Ticket Hypothesis: Finding Sparse, Trainable Neural Networks},
  booktitle    = {ICLR},
  year         = {2019},
}

@inproceedings{DBLP:conf/iclr/MingSD023,
  author       = {Yifei Ming and
                  Yiyou Sun and
                  Ousmane Dia and
                  Yixuan Li},
  title        = {How to Exploit Hyperspherical Embeddings for Out-of-Distribution Detection?},
  booktitle    = {ICLR},
  year         = {2023},
}

@inproceedings{DBLP:conf/iclr/XuCFY24,
  author       = {Kai Xu and
                  Rongyu Chen and
                  Gianni Franchi and
                  Angela Yao},
  title        = {Scaling for Training Time and Post-hoc Out-of-distribution Detection
                  Enhancement},
  booktitle    = {ICLR},
  year         = {2024},
}

@inproceedings{DBLP:conf/iclr/LoshchilovH17,
  author       = {Ilya Loshchilov and
                  Frank Hutter},
  title        = {{SGDR:} Stochastic Gradient Descent with Warm Restarts},
  booktitle    = {ICLR},
  year         = {2017},
}

@inproceedings{DBLP:conf/iclr/Vaze0VZ22,
  author       = {Sagar Vaze and
                  Kai Han and
                  Andrea Vedaldi and
                  Andrew Zisserman},
  title        = {Open-Set Recognition: {A} Good Closed-Set Classifier is All You Need},
  booktitle    = {ICLR},
  year         = {2022},
}

@inproceedings{DBLP:conf/iclr/WangY0DKLH023,
  author       = {Qizhou Wang and
                  Junjie Ye and
                  Feng Liu and
                  Quanyu Dai and
                  Marcus Kalander and
                  Tongliang Liu and
                  Jianye Hao and
                  Bo Han},
  title        = {Out-of-distribution Detection with Implicit Outlier Transformation},
  booktitle    = {ICLR},
  year         = {2023},
}

@inproceedings{DBLP:conf/iclr/DuWCL22,
  author       = {Xuefeng Du and
                  Zhaoning Wang and
                  Mu Cai and
                  Yixuan Li},
  title        = {{VOS:} Learning What You Don't Know by Virtual Outlier Synthesis},
  booktitle    = {ICLR},
  year         = {2022},
}

@inproceedings{DBLP:conf/iclr/JiangCCWW24,
  author       = {Wenyu Jiang and
                  Hao Cheng and
                  Mingcai Chen and
                  Chongjun Wang and
                  Hongxin Wei},
  title        = {{DOS:} Diverse Outlier Sampling for Out-of-Distribution Detection},
  booktitle    = {ICLR},
  year         = {2024},
}

@inproceedings{DBLP:conf/iclr/DjurisicBAL23,
  author       = {Andrija Djurisic and
                  Nebojsa Bozanic and
                  Arjun Ashok and
                  Rosanne Liu},
  title        = {Extremely Simple Activation Shaping for Out-of-Distribution Detection},
  booktitle    = {ICLR},
  year         = {2023},
}

@inproceedings{DBLP:conf/iclr/HendrycksMD19,
  author       = {Dan Hendrycks and
                  Mantas Mazeika and
                  Thomas G. Dietterich},
  title        = {Deep Anomaly Detection with Outlier Exposure},
  booktitle    = {ICLR},
  year         = {2019},
}

@inproceedings{DBLP:conf/iclr/LiangLS18,
  author       = {Shiyu Liang and
                  Yixuan Li and
                  R. Srikant},
  title        = {Enhancing The Reliability of Out-of-distribution Image Detection in
                  Neural Networks},
  booktitle    = {ICLR},
  year         = {2018},
}

@inproceedings{DBLP:conf/iclr/HendrycksG17,
  author       = {Dan Hendrycks and
                  Kevin Gimpel},
  title        = {A Baseline for Detecting Misclassified and Out-of-Distribution Examples
                  in Neural Networks},
  booktitle    = {ICLR},
  year         = {2017},
}

@inproceedings{DBLP:conf/iclr/DosovitskiyB0WZ21,
  author       = {Alexey Dosovitskiy and
                  Lucas Beyer and
                  Alexander Kolesnikov and
                  Dirk Weissenborn and
                  Xiaohua Zhai and
                  Thomas Unterthiner and
                  Mostafa Dehghani and
                  Matthias Minderer and
                  Georg Heigold and
                  Sylvain Gelly and
                  Jakob Uszkoreit and
                  Neil Houlsby},
  title        = {An Image is Worth 16x16 Words: Transformers for Image Recognition
                  at Scale},
  booktitle    = {ICLR},
  year         = {2021},
}

@inproceedings{DBLP:conf/icml/SunM0L22,
  author       = {Yiyou Sun and
                  Yifei Ming and
                  Xiaojin Zhu and
                  Yixuan Li},
  title        = {Out-of-Distribution Detection with Deep Nearest Neighbors},
  booktitle    = {ICML},
  pages        = {20827--20840},
  year         = {2022},
}

@inproceedings{DBLP:conf/icml/BitterwolfM023,
  author       = {Julian Bitterwolf and
                  Maximilian M{\"{u}}ller and
                  Matthias Hein},
  title        = {In or Out? Fixing ImageNet Out-of-Distribution Detection Evaluation},
  booktitle    = {ICML},
  pages        = {2471--2506},
  year         = {2023},
}

@inproceedings{DBLP:conf/icml/ZhuLYLX023,
  author       = {Jianing Zhu and
                  Hengzhuang Li and
                  Jiangchao Yao and
                  Tongliang Liu and
                  Jianliang Xu and
                  Bo Han},
  title        = {Unleashing Mask: Explore the Intrinsic Out-of-Distribution Detection
                  Capability},
  booktitle    = {ICML},
  pages        = {43068--43104},
  year         = {2023},
}

@inproceedings{DBLP:conf/icml/HendrycksBMZKMS22,
  author       = {Dan Hendrycks and
                  Steven Basart and
                  Mantas Mazeika and
                  Andy Zou and
                  Joseph Kwon and
                  Mohammadreza Mostajabi and
                  Jacob Steinhardt and
                  Dawn Song},
  pages        = {8759--8773},
  title        = {Scaling Out-of-Distribution Detection for Real-World Settings},
  booktitle    = {ICML},
  year         = {2022},
}

@inproceedings{DBLP:conf/icml/WeiXCF0L22,
  author       = {Hongxin Wei and
                  Renchunzi Xie and
                  Hao Cheng and
                  Lei Feng and
                  Bo An and
                  Yixuan Li},
  title        = {Mitigating Neural Network Overconfidence with Logit Normalization},
  pages        = {23631--23644},
  booktitle    = {ICML},
  year         = {2022},
}

@inproceedings{DBLP:conf/nips/0074WJ024,
  author       = {Yang Yang and
                  Fengqiang Wan and
                  Qing{-}Yuan Jiang and
                  Yi Xu},
  title        = {Facilitating Multimodal Classification via Dynamically Learning Modality
                  Gap},
  booktitle    = {NeurIPS},
  year         = {2024},
}

@inproceedings{DBLP:conf/nips/00030P0C24,
  author       = {Yonggang Zhang and
                  Jie Lu and
                  Bo Peng and
                  Zhen Fang and
                  Yiu{-}ming Cheung},
  title        = {Learning to Shape In-distribution Feature Space for Out-of-distribution
                  Detection},
  booktitle    = {NeurIPS},
  year         = {2024},
}

@inproceedings{DBLP:conf/nips/LiuWOL20,
  author       = {Weitang Liu and
                  Xiaoyun Wang and
                  John D. Owens and
                  Yixuan Li},
  title        = {Energy-based Out-of-distribution Detection},
  booktitle    = {NeurIPS},
  year         = {2020},
}

@inproceedings{DBLP:conf/nips/SunGL21,
  author       = {Yiyou Sun and
                  Chuan Guo and
                  Yixuan Li},
  title        = {ReAct: Out-of-distribution Detection With Rectified Activations},
  pages        = {144--157},
  booktitle    = {NeurIPS},
  year         = {2021},
}

@inproceedings{DBLP:conf/nips/ChenFLCTY23,
  author       = {Chao Chen and
                  Zhihang Fu and
                  Kai Liu and
                  Ze Chen and
                  Mingyuan Tao and
                  Jieping Ye},
  title        = {Optimal Parameter and Neuron Pruning for Out-of-Distribution Detection},
  booktitle    = {NeurIPS},
  year         = {2023},
}

@inproceedings{DBLP:conf/nips/HuangGL21,
  author       = {Rui Huang and
                  Andrew Geng and
                  Yixuan Li},
  title        = {On the Importance of Gradients for Detecting Distributional Shifts in the Wild},
  pages        = {677--689},
  booktitle    = {NeurIPS},
  year         = {2021},
}

@inproceedings{DBLP:conf/nips/ZhuCXLZ00ZC22,
  author       = {Yao Zhu and
                  Yuefeng Chen and
                  Chuanlong Xie and
                  Xiaodan Li and
                  Rong Zhang and
                  Hui Xue and
                  Xiang Tian and
                  Bolun Zheng and
                  Yaowu Chen},
  title        = {Boosting Out-of-distribution Detection with Typical Features},
  booktitle    = {NeurIPS},
  year         = {2022},
}

@inproceedings{DBLP:conf/nips/XuLLT23,
  author       = {Mingyu Xu and
                  Zheng Lian and
                  Bin Liu and
                  Jianhua Tao},
  title        = {{VRA:} Variational Rectified Activation for Out-of-distribution Detection},
  booktitle    = {NeurIPS},
  year         = {2023},
}

@inproceedings{DBLP:conf/cvpr/RegmiPDGSB22,
  author       = {Sudarshan Regmi and
                  Bibek Panthi and
                  Sakar Dotel and
                  Prashnna K. Gyawali and
                  Danail Stoyanov and
                  Binod Bhattarai},
  title        = {T2FNorm: Train-time Feature Normalization for {OOD} Detection in Image
                  Classification},
  booktitle    = {CVPRW},
  pages        = {153--162},
  year         = {2024},
}

@inproceedings{DBLP:conf/cvpr/Wang0F022,
  author       = {Haoqi Wang and
                  Zhizhong Li and
                  Litong Feng and
                  Wayne Zhang},
  title        = {ViM: Out-Of-Distribution with Virtual-logit Matching},
  booktitle    = {CVPR},
  pages        = {4911--4920},
  year         = {2022},
}

@inproceedings{DBLP:conf/cvpr/HuangL21,
  author       = {Rui Huang and
                  Yixuan Li},
  title        = {{MOS:} Towards Scaling Out-of-Distribution Detection for Large Semantic
                  Space},
  booktitle    = {CVPR},
  pages        = {8710--8719},
  year         = {2021},
}

@inproceedings{DBLP:conf/cvpr/NguyenYC15,
  author       = {Anh Mai Nguyen and
                  Jason Yosinski and
                  Jeff Clune},
  title        = {Deep neural networks are easily fooled: High confidence predictions
                  for unrecognizable images},
  booktitle    = {CVPR},
  pages        = {427--436},
  year         = {2015},
}

@inproceedings{DBLP:conf/cvpr/DengDSLL009,
  author       = {Jia Deng and
                  Wei Dong and
                  Richard Socher and
                  Li{-}Jia Li and
                  Kai Li and
                  Li Fei{-}Fei},
  title        = {ImageNet: {A} large-scale hierarchical image database},
  booktitle    = {CVPR},
  pages        = {248--255},
  year         = {2009},
}

@inproceedings{DBLP:conf/cvpr/AhnPK23,
  author       = {Yong Hyun Ahn and
                  Gyeong{-}Moon Park and
                  Seong Tae Kim},
  title        = {LINe: Out-of-Distribution Detection by Leveraging Important Neurons},
  pages        = {19852--19862},
  booktitle    = {CVPR},
  year         = {2023},
}

@inproceedings{DBLP:conf/cvpr/HornASCSSAPB18,
  author       = {Grant Van Horn and
                  Oisin Mac Aodha and
                  Yang Song and
                  Yin Cui and
                  Chen Sun and
                  Alexander Shepard and
                  Hartwig Adam and
                  Pietro Perona and
                  Serge J. Belongie},
  title        = {The INaturalist Species Classification and Detection Dataset},
  pages        = {8769--8778},
  booktitle    = {CVPR},
  year         = {2018},
}

@inproceedings{DBLP:conf/cvpr/CimpoiMKMV14,
  author       = {Mircea Cimpoi and
                  Subhransu Maji and
                  Iasonas Kokkinos and
                  Sammy Mohamed and
                  Andrea Vedaldi},
  title        = {Describing Textures in the Wild},
  pages        = {3606--3613},
  booktitle    = {CVPR},
  year         = {2014},
}

@inproceedings{DBLP:conf/cvpr/HeZRS16,
  author       = {Kaiming He and
                  Xiangyu Zhang and
                  Shaoqing Ren and
                  Jian Sun},
  title        = {Deep Residual Learning for Image Recognition},
  pages        = {770--778},
  booktitle    = {CVPR},
  year         = {2016},
}

@inproceedings{DBLP:conf/cvpr/GeigerLU12,
  author       = {Andreas Geiger and
                  Philip Lenz and
                  Raquel Urtasun},
  title        = {Are we ready for autonomous driving? The {KITTI} vision benchmark
                  suite},
  pages        = {3354--3361},
  booktitle    = {CVPR},
  year         = {2012},
}

@inproceedings{DBLP:conf/cvpr/HuangLMW17,
  author       = {Gao Huang and
                  Zhuang Liu and
                  Laurens van der Maaten and
                  Kilian Q. Weinberger},
  title        = {Densely Connected Convolutional Networks},
  pages        = {2261--2269},
  booktitle    = {CVPR},
  year         = {2017},
}

@inproceedings{DBLP:conf/cvpr/BendaleB16,
  author       = {Abhijit Bendale and
                  Terrance E. Boult},
  title        = {Towards Open Set Deep Networks},
  pages        = {1563--1572},
  booktitle    = {CVPR},
  year         = {2016},
}

@inproceedings{DBLP:conf/eccv/ZhangLDZBCW20,
  author       = {Youcai Zhang and
                  Zhonghao Lan and
                  Yuchen Dai and
                  Fangao Zeng and
                  Yan Bai and
                  Jie Chang and
                  Yichen Wei},
  title        = {Prime-Aware Adaptive Distillation},
  booktitle    = {ECCV},
  volume       = {12364},
  pages        = {658--674},
  year         = {2020},
}

@inproceedings{DBLP:conf/eccv/SunL22a,
  author       = {Yiyou Sun and
                  Yixuan Li},
  title        = {{DICE:} Leveraging Sparsification for Out-of-Distribution Detection},
  volume       = {13684},
  pages        = {691--708},
  booktitle    = {ECCV},
  year         = {2022},
}

@inproceedings{DBLP:conf/iccv/LiuL00W0LG21,
  author       = {Ze Liu and
                  Yutong Lin and
                  Yue Cao and
                  Han Hu and
                  Yixuan Wei and
                  Zheng Zhang and
                  Stephen Lin and
                  Baining Guo},
  title        = {Swin Transformer: Hierarchical Vision Transformer using Shifted Windows},
  pages        = {9992--10002},
  booktitle    = {ICCV},
  year         = {2021},
}

@inproceedings{DBLP:conf/aaai/XuY25,
  author       = {Haonan Xu and
                  Yang Yang},
  title        = {{ITP:} Instance-Aware Test Pruning for Out-of-Distribution Detection},
  booktitle    = {AAAI},
  pages        = {21743--21751},
  year         = {2025},
}

@inproceedings{DBLP:conf/aaai/GhosalSL24,
  author       = {Soumya Suvra Ghosal and
                  Yiyou Sun and
                  Yixuan Li},
  title        = {How to Overcome Curse-of-Dimensionality for Out-of-Distribution Detection?},
  booktitle    = {AAAI},
  pages        = {19849--19857},
  year         = {2024},
}

@article{DBLP:journals/pami/ZhouLKO018,
  author       = {Bolei Zhou and
                  {\`{A}}gata Lapedriza and
                  Aditya Khosla and
                  Aude Oliva and
                  Antonio Torralba},
  title        = {Places: {A} 10 Million Image Database for Scene Recognition},
  journal      = {IEEE Transactions on Pattern Analysis and Machine Intelligence},
  year         = {2018},
}

@inproceedings{DBLP:conf/wacv/ZhangILCL23,
  author       = {Jingyang Zhang and
                  Nathan Inkawhich and
                  Randolph Linderman and
                  Yiran Chen and
                  Hai Li},
  title        = {Mixture Outlier Exposure: Towards Out-of-Distribution Detection in
                  Fine-grained Environments},
  booktitle    = {WACV},
  pages        = {5520--5529},
  year         = {2023},
}

@article{DBLP:journals/corr/abs-1802-04865,
  author       = {Terrance DeVries and
                  Graham W. Taylor},
  title        = {Learning Confidence for Out-of-Distribution Detection in Neural Networks},
  journal      = {CoRR},
  volume       = {abs/1802.04865},
  year         = {2018},
}

@article{Netzer_Wang_Coates_Bissacco_Wu_Ng_2011,  
    title      = {Reading Digits in Natural Images with Unsupervised Feature Learning}, 
    author     = {Netzer, Yuval and Wang, Tao and Coates, Adam and Bissacco, Alessandro and Wu, Bo and Ng, AndrewY.}, 
    journal    = {},
    year       = {2011}, 
}

@article{Krizhevsky_2009,  
    title      = {Learning Multiple Layers of Features from Tiny Images}, 
    author     = {Krizhevsky, Alex}, 
    journal    = {},
    year       = {2009}, 
}

@article{radford2018improving,
    title      = {Improving language understanding by generative pre-training},
    author     = {Radford, Alec and Narasimhan, Karthik and Salimans, Tim and Sutskever, Ilya and others},
    journal    = {},
    year       = {2018},
}

@article{achiam2023gpt,
    title      = {Gpt-4 technical report},
    author     = {Achiam, Josh and Adler, Steven and Agarwal, Sandhini and Ahmad, Lama and Akkaya, Ilge and Aleman, Florencia Leoni and Almeida, Diogo and Altenschmidt, Janko and Altman, Sam and Anadkat, Shyamal and others},
    volume     = {abs/2303.08774},
    journal    = {CoRR},
    year       = {2023}
}

@article{litjens2017survey,
    title      = {A survey on deep learning in medical image analysis},
    author     = {Litjens, Geert and Kooi, Thijs and Bejnordi, Babak Ehteshami and Setio, Arnaud Arindra Adiyoso and Ciompi, Francesco and Ghafoorian, Mohsen and Van Der Laak, Jeroen Awm and Van Ginneken, Bram and S{\'a}nchez, Clara I},
    journal    = {Medical image analysis},
    volume     = {42},
    pages      = {60--88},
    year       = {2017},
}

@article{DBLP:journals/corr/abs-2110-11334,
  author       = {Jingkang Yang and
                  Kaiyang Zhou and
                  Yixuan Li and
                  Ziwei Liu},
  title        = {Generalized Out-of-Distribution Detection: {A} Survey},
  journal      = {CoRR},
  volume       = {abs/2110.11334},
  year         = {2021},
}

@article{DBLP:journals/corr/abs-2306-09301,
  author       = {Jingyang Zhang and
                  Jingkang Yang and
                  Pengyun Wang and
                  Haoqi Wang and
                  Yueqian Lin and
                  Haoran Zhang and
                  Yiyou Sun and
                  Xuefeng Du and
                  Kaiyang Zhou and
                  Wayne Zhang and
                  Yixuan Li and
                  Ziwei Liu and
                  Yiran Chen and
                  Hai Li},
  title        = {OpenOOD v1.5: Enhanced Benchmark for Out-of-Distribution Detection},
  journal      = {CoRR},
  volume       = {abs/2306.09301},
  year         = {2023},
}

@inproceedings{DBLP:conf/bmvc/ZagoruykoK16,
  author       = {Sergey Zagoruyko and
                  Nikos Komodakis},
  title        = {Wide Residual Networks},
  booktitle    = {BMVC},
  year         = {2016},
}

@article{DBLP:journals/spm/Deng12,
  author       = {Li Deng},
  title        = {The {MNIST} Database of Handwritten Digit Images for Machine Learning
                  Research},
  journal      = {IEEE Signal Processing Magazine},
  volume       = {29},
  number       = {6},
  pages        = {141--142},
  year         = {2012},
}

@article{le2015tiny,
  title={Tiny imagenet visual recognition challenge},
  author={Le, Yann and Yang, Xuan},
  journal={CS 231N},
  volume={7},
  number={7},
  pages={3},
  year={2015}
}

@article{DBLP:journals/corr/abs-1803-08375,
  author       = {Abien Fred Agarap},
  title        = {Deep Learning using Rectified Linear Units (ReLU)},
  journal      = {CoRR},
  volume       = {abs/1803.08375},
  year         = {2018},
}

@article{tibshirani1996regression,
  title={Regression shrinkage and selection via the lasso},
  author={Tibshirani, Robert},
  journal={Journal of the Royal Statistical Society Series B: Statistical Methodology},
  volume={58},
  number={1},
  pages={267--288},
  year={1996},
}

@inproceedings{DBLP:conf/iccv/ArandjelovicZ17,
  author       = {Relja Arandjelovic and
                  Andrew Zisserman},
  title        = {Look, Listen and Learn},
  booktitle    = {ICCV},
  pages        = {609--617},
  year         = {2017},
}

@inproceedings{yangstrengthen,
  title={Strengthen Out-of-Distribution Detection Capability with Progressive Self-Knowledge Distillation},
  author={Yang, Yang and Xu, Haonan},
  booktitle={ICML},
  year={2025}
}

@inproceedings{wanprobabilistic,
  title={Probabilistic Group Mask Guided Discrete Optimization for Incremental Learning},
  author={Wan, Fengqiang and Yang, Yang},
  booktitle={ICML},
  year={2025}
}
\clearpage

\appendix
\section{Sparse Parameter Contribution Patterns}
\label{apd:vis}
\begin{figure*}[h]
    \centering
    \begin{subfigure}{\textwidth}
        \centering
        \begin{subfigure}{0.32\textwidth}
            \centering
            \includegraphics[width=\textwidth]{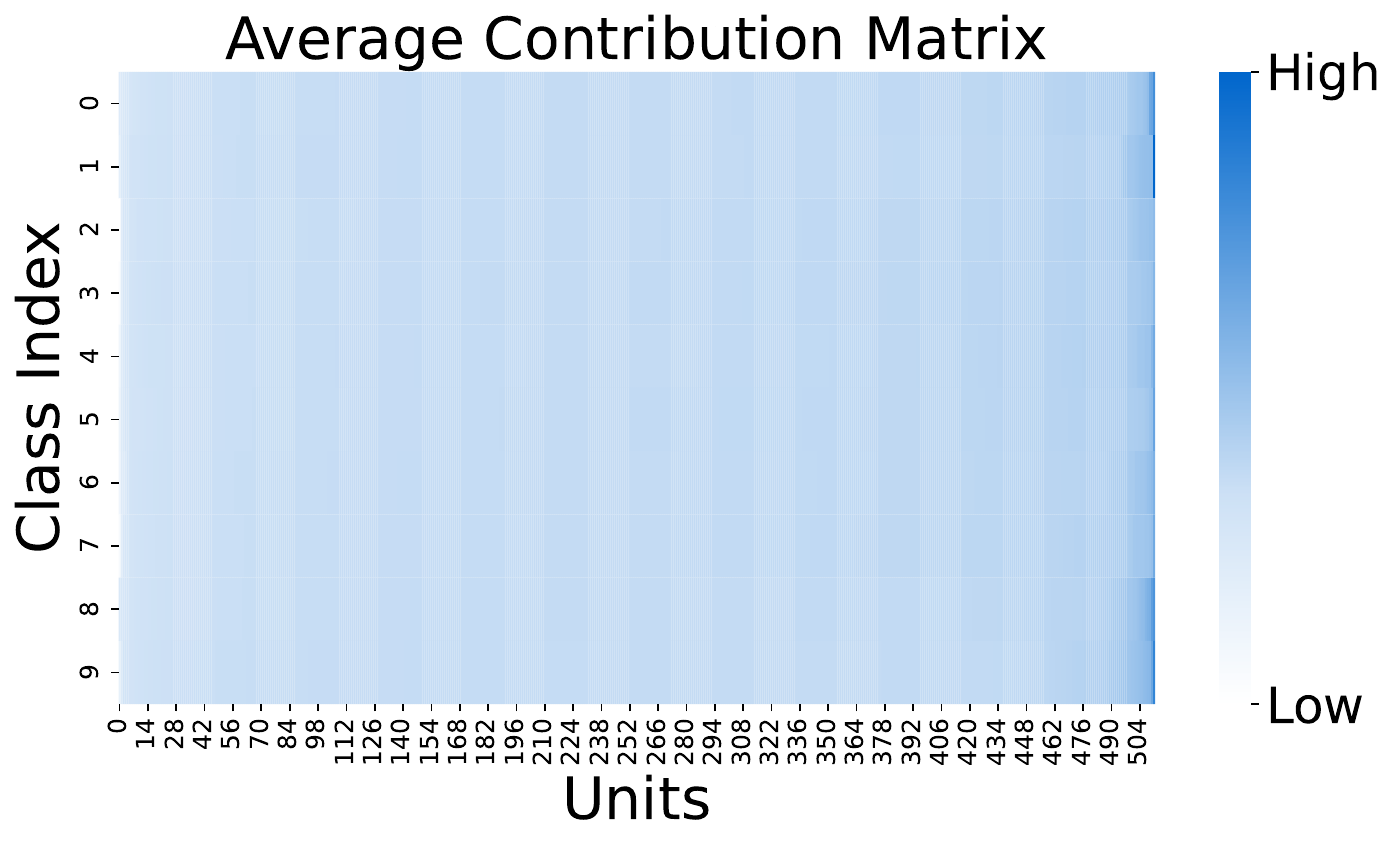}
            \captionsetup{skip=2.5pt}
            \caption*{RN18 on CIFAR-10}
        \end{subfigure}%
        \hfill
        \captionsetup{skip=2.5pt}
        \begin{subfigure}{0.32\textwidth}
            \centering
            \includegraphics[width=\textwidth]{plt/contribution_heatmap_c100_rn18.pdf}
            \captionsetup{skip=2.5pt}
            \caption*{RN18 on CIFAR-100}
        \end{subfigure}
        \hfill
        \captionsetup{skip=2.5pt}
        \begin{subfigure}{0.32\textwidth}
            \centering
            \includegraphics[width=\textwidth]{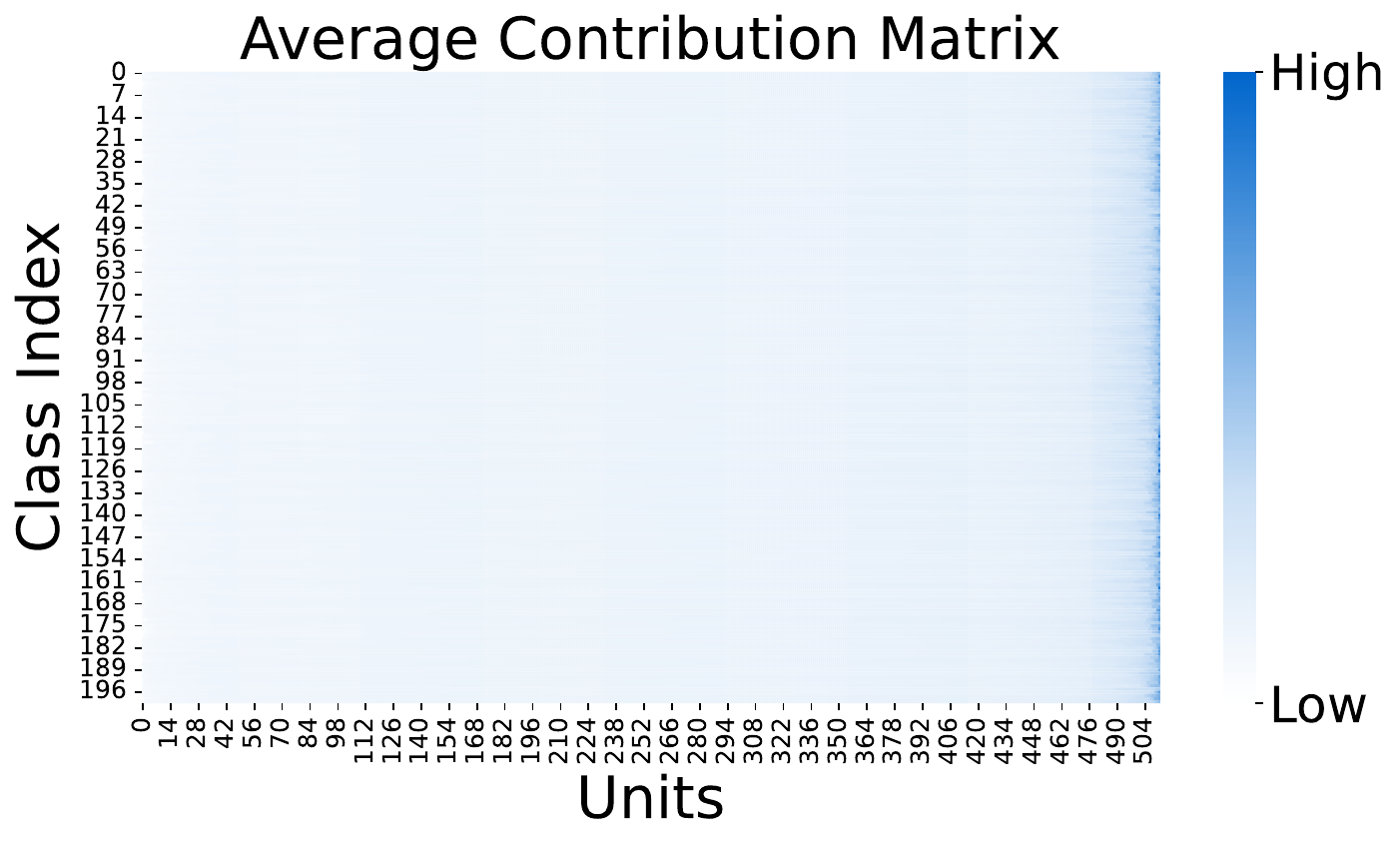}
            \captionsetup{skip=2.5pt}
            \caption*{RN18 on ImageNet-200}
        \end{subfigure}
    \end{subfigure}%
    \\
    \begin{subfigure}{\textwidth}
        \centering
        \begin{subfigure}{0.32\textwidth}
            \centering
            \includegraphics[width=\textwidth]{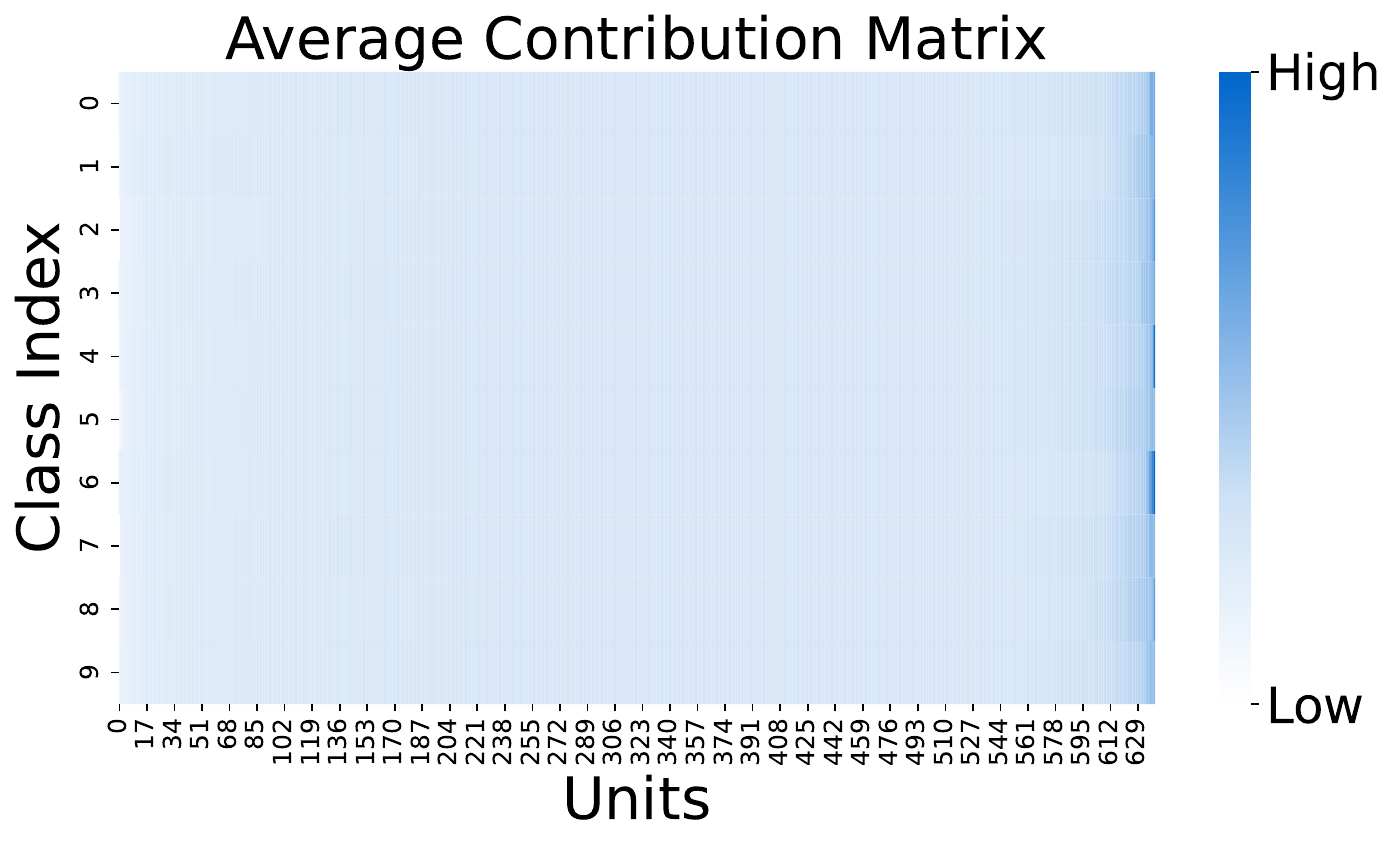}
            \captionsetup{skip=2.5pt}
            \caption*{WRN on CIFAR-10}
        \end{subfigure}%
        \hfill
        \captionsetup{skip=2.5pt}
        \begin{subfigure}{0.32\textwidth}
            \centering
            \includegraphics[width=\textwidth]{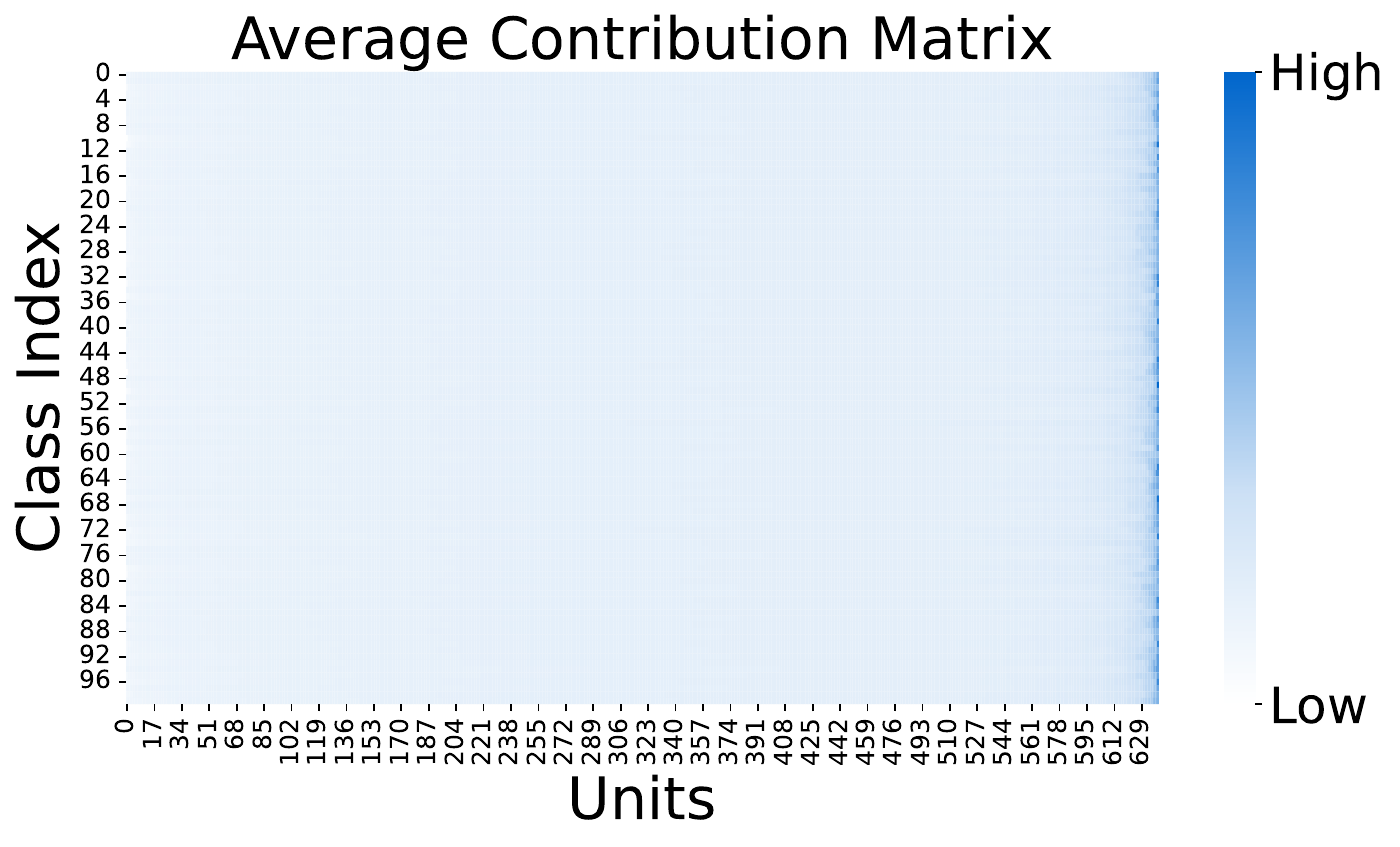}
            \captionsetup{skip=2.5pt}
            \caption*{WRN on CIFAR-100}
        \end{subfigure}
        \hfill
        \captionsetup{skip=2.5pt}
        \begin{subfigure}{0.32\textwidth}
            \centering
            \includegraphics[width=\textwidth]{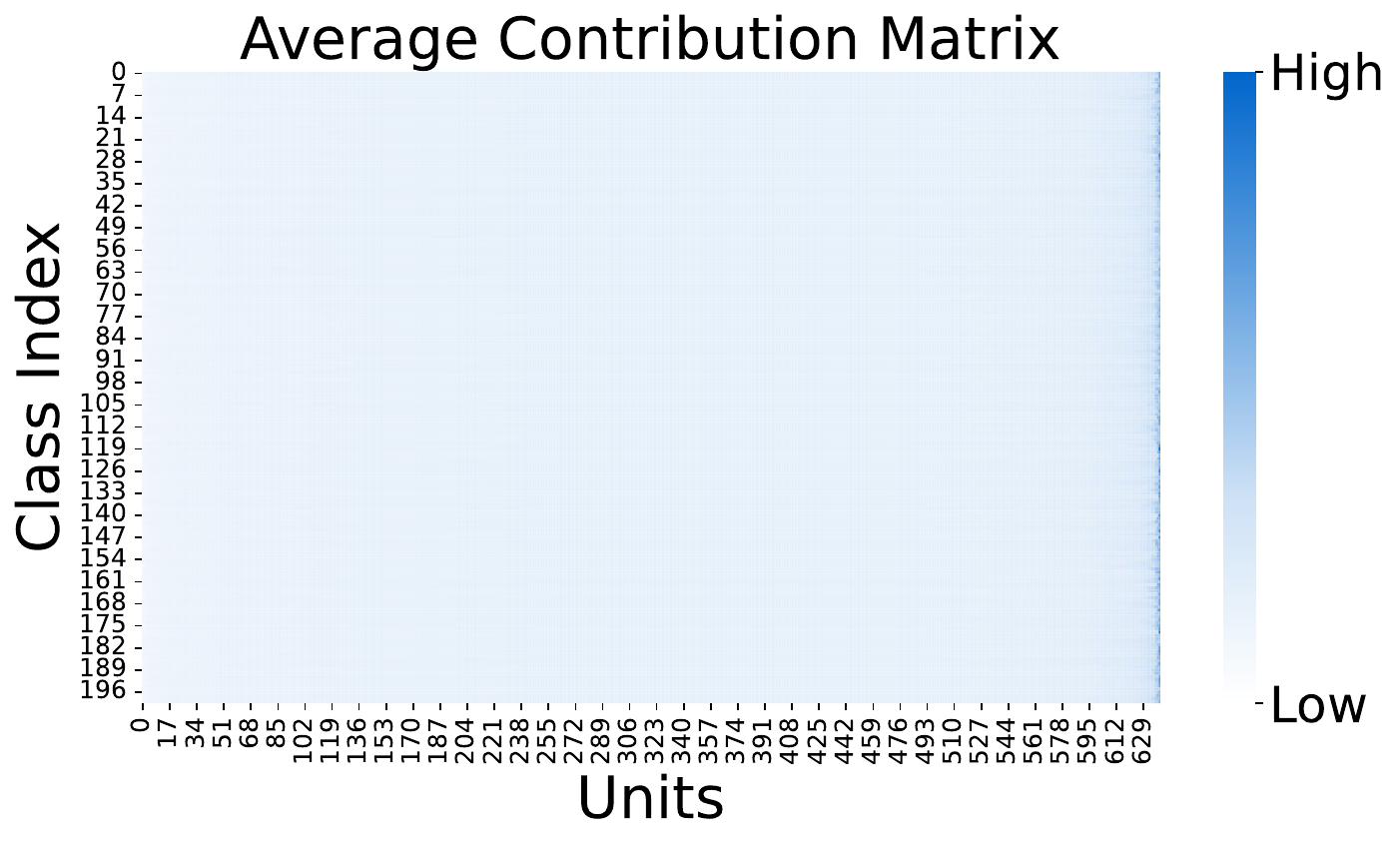}
            \captionsetup{skip=2.5pt}
            \caption*{WRN on ImageNet-200}
        \end{subfigure}
    \end{subfigure}%
    \\
    \begin{subfigure}{\textwidth}
        \centering
        \begin{subfigure}{0.32\textwidth}
            \centering
            \includegraphics[width=\textwidth]{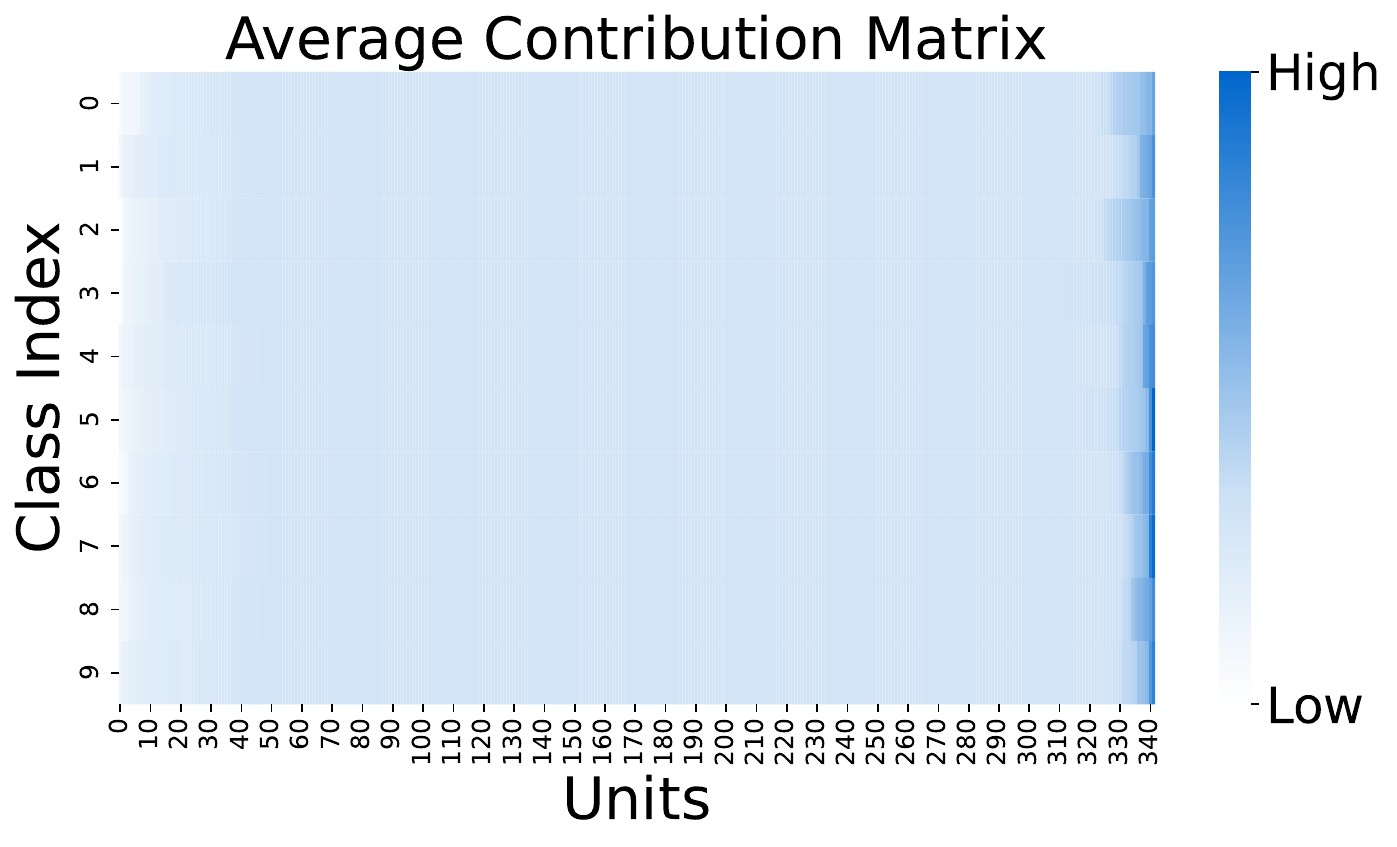}
            \captionsetup{skip=2.5pt}
            \caption*{DN101 on CIFAR-10}
        \end{subfigure}%
        \hfill
        \captionsetup{skip=2.5pt}
        \begin{subfigure}{0.32\textwidth}
            \centering
            \includegraphics[width=\textwidth]{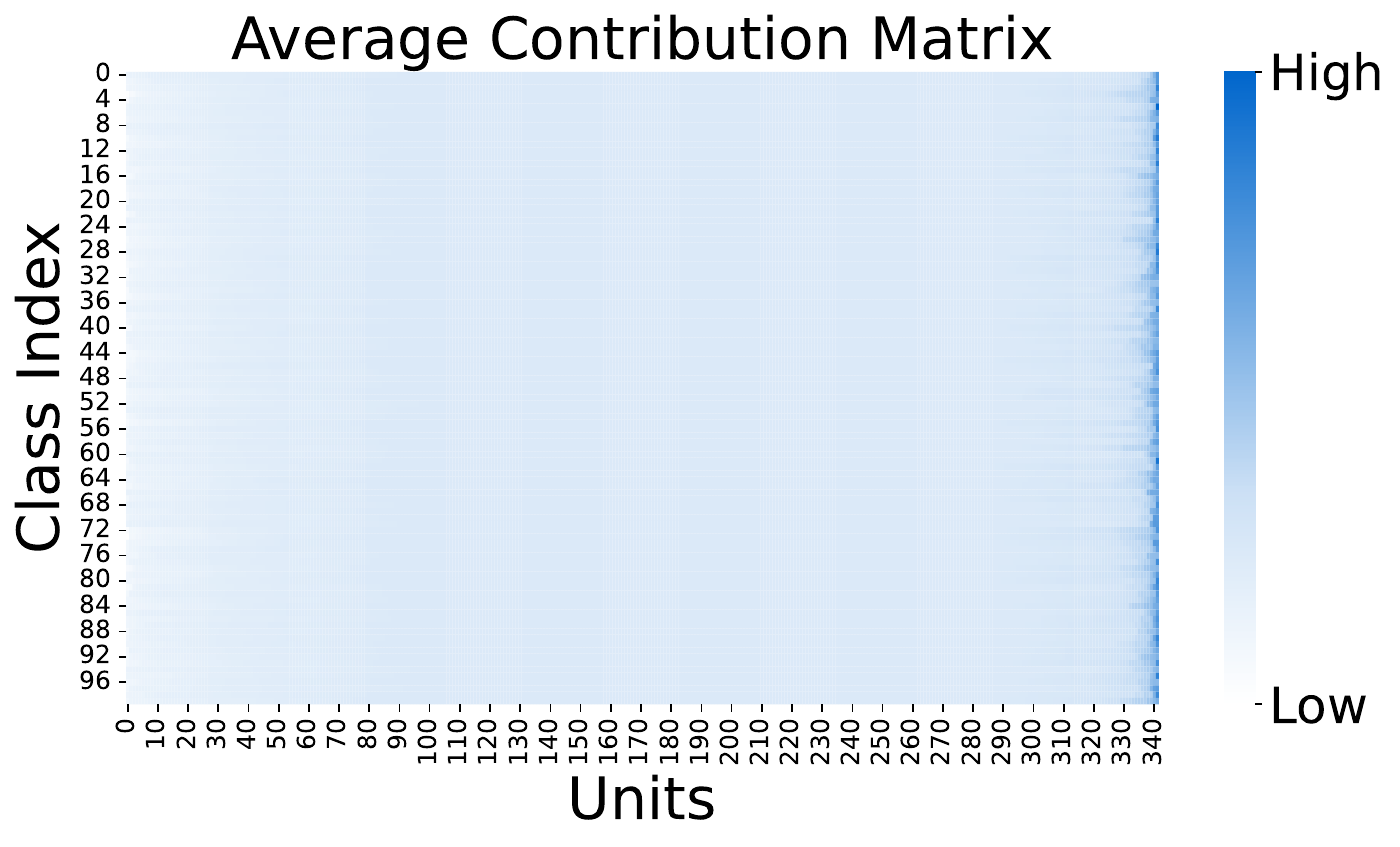}
            \captionsetup{skip=2.5pt}
            \caption*{DN101 on CIFAR-100}
        \end{subfigure}
        \hfill
        \captionsetup{skip=2.5pt}
        \begin{subfigure}{0.32\textwidth}
            \centering
            \includegraphics[width=\textwidth]{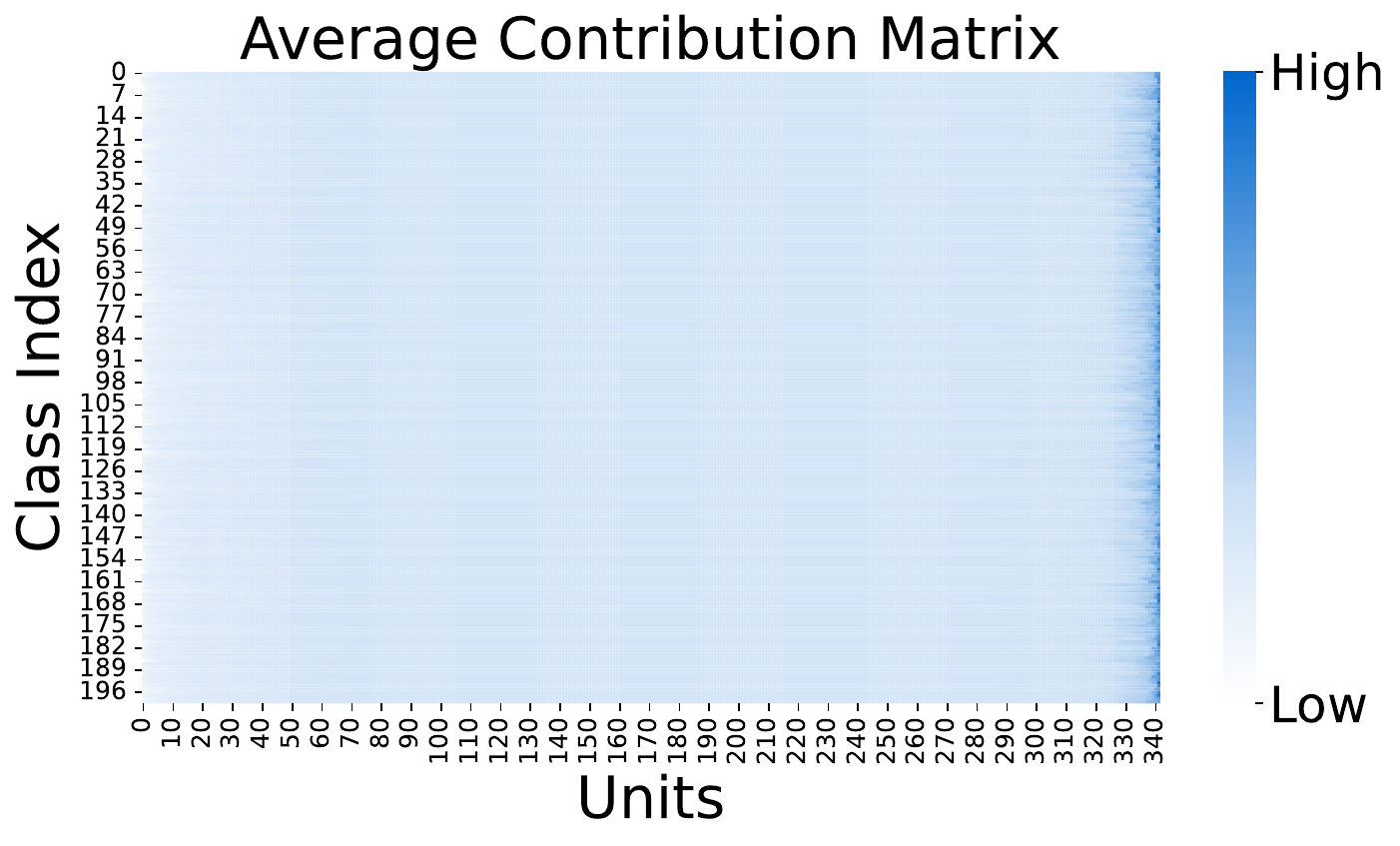}
            \captionsetup{skip=2.5pt}
            \caption*{DN101 on ImageNet-200}
        \end{subfigure}
    \end{subfigure}%
    % \captionsetup{skip=1.0pt}
    \vspace{-1mm}
    \caption{Visualization of the classifier’s parameter contribution patterns, with the average contribution matrix on the ID test set sorted for improved clarity.}
    \label{fig:expamles}
\end{figure*}

To further support our observations on the sparsity of classifier parameter contributions, we provide additional examples across various datasets and model architectures. Specifically, as shown in Figure \ref{fig:expamles}, we present results from models trained on CIFAR-10 \cite{Krizhevsky_2009}, CIFAR-100 \cite{Krizhevsky_2009}, and ImageNet-200 \cite{DBLP:conf/cvpr/DengDSLL009}, using ResNet-18 (RN18) \cite{DBLP:conf/cvpr/HeZRS16}, WideResNet-28-10 (WRN) \cite{DBLP:conf/bmvc/ZagoruykoK16}, and DenseNet-101 (DN101) \cite{DBLP:conf/cvpr/HuangLMW17} as backbones. The results demonstrate that, across diverse settings, predictive behavior in modern deep models is commonly dominated by a small subset of classifier parameters.

\section{Details of Parameter Contribution}
\label{apd:simple}
To derive the simplified contribution form for classifier weights, we consider a linear classifier setup in typical classification model. Let $h(\mathbf{x}) \in \mathbb{R}^D$ be the feature representation of input $\mathbf{x}$ produced by the penultimate layer of the network. The classifier layer then computes logits as:
\begin{equation}
f(\mathbf{x}) = \mathbf{W}^\top h(\mathbf{x}) + \mathbf{b},
\end{equation}
where $\mathbf{W} \in \mathbb{R}^{D \times K}$ is the weight matrix and $\mathbf{b} \in \mathbb{R}^K$ is the bias vector. Accordingly, the $k$-th output logit is given by:
\begin{equation}
f_k(\mathbf{x}) = \sum_{d=1}^{D} \mathbf{W}_{dk} h_d(\mathbf{x}) + \mathbf{b}_k.
\end{equation}
To compute the contribution of an individual parameter $\mathbf{W}_{ij}$ to class $k$, we compare the output $f_k(\mathbf{x})$ with and without $\mathbf{W}_{ij}$:
\begin{align}
c_k(\mathbf{x}; \mathbf{W}_{ij}) 
&= f_k(\mathbf{x}) - f_k(\mathbf{x}; \mathbf{W}_{ij} = 0) \nonumber \\
&= \left( \sum_{d=1}^{D} \mathbf{W}_{dk} h_d(\mathbf{x}) + \mathbf{b}_k \right) \nonumber \\
&\quad - \left( \sum_{d=1}^{D} \mathbf{W}'_{dk} h_d(\mathbf{x}) + \mathbf{b}_k \right),
\end{align}
where $\mathbf{W}'$ is identical to $\mathbf{W}$ except for the element $\mathbf{W}'_{ij} = 0$. This simplifies to:
\begin{equation}
\label{eq:simp}
c_k(\mathbf{x}; \mathbf{W}_{ij}) =
\begin{cases}
\mathbf{W}_{ij} h_i(\mathbf{x}), & \text{if } j = k, \\
0, & \text{if } j \ne k.
\end{cases}
\end{equation}
The principle behind this is that \(\mathbf{W}_{ij}\) directly affects only the logit corresponding to class \(j\), due to the structure of the single matrix-vector product. For input \(\mathbf{x}\), we define the classifier parameter contribution matrix \(\mathbf{C}(\mathbf{x}) \in \mathbb{R}^{D \times K}\) \emph{w.r.t.} the entire weight matrix \(\mathbf{W}\). According to Eq. \eqref{eq:simp}, the contribution matrix can be efficiently computed by broadcasting the feature vector \(h(\mathbf{x})\) to match the shape of \(\mathbf{W}\) and performing the Hadamard product:
\begin{equation}
\mathbf{C}(\mathbf{x}; \textbf{W}) = \mathbf{W} \circ \big(h(\mathbf{x}) \mathbf{1}_K^\top \big),
\end{equation}
where \(\circ\) denotes the Hadamard product and \(\mathbf{1}_K \in \mathbb{R}^{K}\) is a vector of ones. The additional computational overhead introduced by computing the classifier parameter contributions is analyzed in Appendix \ref{apd:overhead}.

\section{Implementation details}
\label{apd:implement}
\noindent\textbf{Hardware and Software.} All experiments in this work are conducted using Python 3.8.19 and PyTorch 2.0.1, running on NVIDIA GeForce RTX 4090 GPUs.

\noindent\textbf{Hyperparameter Selection.}
Following \cite{DBLP:journals/corr/abs-2306-09301}, SPCP performs hyperparameter tuning using the ID and OOD validation sets provided by the OpenOOD benchmark. In cases where a realistic OOD validation set is unavailable, Gaussian noise can effectively substitute as the OOD validation data \cite{DBLP:conf/nips/ZhuCXLZ00ZC22,DBLP:conf/aaai/XuY25}. To ensure hyperparameter consistency across datasets with varying numbers of classes $K$, we introduce a normalized control parameter $\rho_{\text{norm}}$ to replace the original contribution percentile $\rho \in [0, 100]$ (defined in Eq.~(9) of the main paper). Specifically, $\rho$ is redefined as $\frac{\rho_{\text{norm}} \cdot 100}{K}$. $\rho_{\text{norm}}$ decouples the effect of $K$ from $\rho$, enabling scalable and interpretable hyperparameter tuning. We conduct a grid search over $\rho_{\text{norm}}$ values in the set $\{0.1, 0.2, 0.3, 0.4, 0.5, 1.0, 2.0, 3.0, 5.0\}$. Other hyperparameters are fixed based on prior empirical recommendations: $\beta = 0.999$ and $\lambda_0 = 1000$. Following OpenOOD \cite{DBLP:journals/corr/abs-2306-09301}, we select the $\rho_{\text{norm}}$ value that yields the highest AUROC on the validation set for final evaluation on the test set. The selected $\rho_{\text{norm}}$ is then consistently applied across all experiments using the same model. For the CIFAR-10 benchmark, the optimal hyperparameter is $\rho_{\text{norm}} = 3.0$, whereas for CIFAR-100, $\rho_{\text{norm}} = 0.5$ yields the best performance. In the case of ImageNet-200, the optimal setting is $\rho_{\text{norm}} = 0.4$.

\noindent\textbf{Empirical Guidelines for Choosing $\rho_{\text{norm}}$.}
Based on the optimal parameter values identified in the grid search, we observe an empirical trend: datasets with fewer classes tend to benefit from relatively larger values of $\rho_{\text{norm}}$. In such cases, competition among model parameters is less intense, and imposing a stricter upper bound on parameter contributions can enhance the model’s resilience to OOD-triggered parameter, thereby mitigating overconfidence. Conversely, for datasets with a larger number of classes, smaller values of $\rho_{\text{norm}}$ are generally more effective. As class density increases, the risk of inter-class interference becomes more pronounced. In this setting, overly restrictive limits on parameter contributions can lead to conflicts in parameter attribution, reduce class separability, and ultimately degrade OOD detection performance. Overall, this observation provides a useful heuristic for grid search: datasets with fewer classes are likely to favor larger $\rho_{\text{norm}}$ values, whereas those with more classes often require smaller ones.

\noindent\textbf{Sampling Strategy.} To compute the percentile parameter contributions within a batch as defined in Eq.~(9) of the main paper, we adopt different strategies depending on the dataset scale. For small-scale experiments on CIFAR, the estimation is performed using all samples within each batch. As for large-scale experiments on ImageNet, where both the batch size and the number of classifier parameters are substantially larger, we uniformly sample 16 instances from each batch of 256 samples to perform the estimation, striking a balance between computational efficiency and estimation quality.

\section{Fine-grained Experimental Results}
\label{apd:results}
In this section, we provide detailed experimental results on the authoritative OpenOOD benchmark \cite{DBLP:journals/corr/abs-2306-09301} across a diverse range of OOD scenarios. For the CIFAR experiments, evaluations are conducted on six OOD datasets, with the near-OOD group including CIFAR-10/100 \cite{Krizhevsky_2009} and TinyImageNet (TIN) \cite{le2015tiny}, while the far-OOD group comprises MNIST \cite{DBLP:journals/spm/Deng12}, SVHN \cite{Netzer_Wang_Coates_Bissacco_Wu_Ng_2011}, Textures \cite{DBLP:conf/cvpr/CimpoiMKMV14}, and Places365 \cite{DBLP:journals/pami/ZhouLKO018}. For the ImageNet-200 benchmark, evaluations are conducted on five OOD datasets, with the near-OOD group consisting of SSB-Hard \cite{DBLP:conf/iclr/Vaze0VZ22} and NINCO \cite{DBLP:conf/icml/BitterwolfM023}, and the far-OOD group including iNaturalist \cite{DBLP:conf/cvpr/HornASCSSAPB18}, Textures \cite{DBLP:conf/cvpr/CimpoiMKMV14}, and OpenImage-O \cite{DBLP:conf/cvpr/Wang0F022}. The CIFAR-10 results are presented in Tables \ref{apd:r1} and \ref{apd:r2}, CIFAR-100 results in Tables \ref{apd:r3} and \ref{apd:r4}, and ImageNet-200 results in Tables \ref{apd:r5} and \ref{apd:r6}. Our findings include: (1) Our SPCP outperforms post-hoc methods in most OOD scenarios, demonstrating the effectiveness of regularizing the contribution of parameters during the training process. For example, on the CIFAR-10 benchmark, SPCP reduces the FPR95 by 46.05\% on the near-OOD CIFAR-100 and by 29.60\% on the far-OOD MNIST compared to the SCALE baseline. (2) Compared to training-time regularization methods, our SPCP demonstrates competitive performance, achieving optimal or near-optimal results in the vast majority of OOD scenarios. Moreover, when integrated as a plug-in with LogitNorm, SPCP can further enhance the performance of OOD detection, leading to a 4.16\% reduction in FPR95 in the far-OOD setting of the ImageNet benchmark and setting a new state-of-the-art. (3) Regarding performance on ID tasks, SPCP shows a slight decrease only on the CIFAR-10 benchmark, while achieving additional gains on all other benchmarks. Compared to other training-time regularization methods, SPCP effectively strengthens the model's OOD detection capabilities with minimal trade-off in ID task performance.

\section{Discussion on Relevant Method (ReAct \cite{DBLP:conf/nips/SunGL21})}
We clarify the differences between our SPCP and ReAct. ReAct operates by rectifying penultimate activations using a pre-computed upper bound to suppress overconfidence. In contrast, our SPCP goes further by focusing on a more granular level—examining the contributions of individual parameters directly responsible for predictions, which is more impactful for OOD detection. Moreover, while ReAct only addresses anomalous activations at test time, SPCP explicitly guides the model during training to develop a boundary-oriented, dense contribution pattern, yielding substantial gains over simple post-hoc rectification.

\section{Relation to $L_2$ Weight Regularization}
We clarify the difference between our SPCP and $L_2$ regularization in their effects on the classifier’s weight parameters. $L_2$ regularization mitigates overfitting and improves generalization by penalizing the magnitude of the weights. However, this approach focuses solely on the weight values themselves without considering their actual contribution to the model’s output. As a result, it cannot truly achieve a denser pattern of parameter contributions to the model. In contrast, our proposed SPCP method explicitly constrains the contribution patterns of the parameters to the output, by jointly considering both the weight values and the associated activation signals. Although the SPCP regularization is applied only to the classifier layer, the activations are generated through the input and the forward propagation across the entire network. As a result, SPCP indirectly influences the optimization of the whole model. This mechanism encourages the formation of robust and denser parameter contribution patterns during training, thereby enhancing the model’s robustness under OOD input. The results in Table 2 demonstrate that SPCP consistently outperforms $L_2$ regularization in terms of OOD detection performance, corroborating the importance of regularizing the contribution of classifier parameters rather than merely penalizing their magnitudes.

\begin{table}[h]
    \begin{center}
    \resizebox{\linewidth}{!}{
        \begin{tabular}{@{\hspace{0.15cm}}l|@{\hspace{0.2cm}}c@{\hspace{0.2cm}}c@{\hspace{0.4cm}}c@{\hspace{0.2cm}}c@{\hspace{0.4cm}}c@{\hspace{0.15cm}}}
        \toprule[1.5pt]
        \multirow{2}{*}{\textbf{Method}} & \multicolumn{2}{@{\hspace{0.2cm}}c@{\hspace{0.2cm}}}{\textbf{Near-OOD}} & \multicolumn{2}{@{\hspace{0.2cm}}c@{\hspace{0.2cm}}}{\textbf{Far-OOD}} & \multirow{2}{*}{ID ACC$\uparrow$} \\
        & FPR95$\downarrow$ & AUROC$\uparrow$ & FPR95$\downarrow$ & AUROC$\uparrow$ \\
        \midrule
        Vanilla & 55.62{\scriptsize$\pm$0.61} & 80.91{\scriptsize$\pm$0.08} & 56.59{\scriptsize$\pm$1.38} & 79.77{\scriptsize$\pm$0.61} & 77.25{\scriptsize$\pm$0.10} \\ 
        \midrule
        +$L_2$ (0.1) & 63.48{\scriptsize$\pm$0.85} & 79.13{\scriptsize$\pm$0.37} & 65.03{\scriptsize$\pm$2.27} & 78.19{\scriptsize$\pm$0.59} & 76.90{\scriptsize$\pm$0.55} \\
        \midrule
        +$L_2$ (0.01) & 56.66{\scriptsize$\pm$1.05} & 80.75{\scriptsize$\pm$0.15} & 57.20{\scriptsize$\pm$1.28} & 79.34{\scriptsize$\pm$1.17} & \textbf{78.31{\scriptsize$\pm$0.21}} \\
        \midrule
        +$L_2$ (0.001) & 56.54{\scriptsize$\pm$1.35} & 80.65{\scriptsize$\pm$0.32} & 53.93{\scriptsize$\pm$1.93} & 80.70{\scriptsize$\pm$1.18} & 77.60{\scriptsize$\pm$0.31} \\
        \midrule
        +$L_2$ (0.0001) & 55.87{\scriptsize$\pm$1.32} & 80.93{\scriptsize$\pm$0.38} & 54.00{\scriptsize$\pm$1.84} & 80.85{\scriptsize$\pm$0.97} & 77.40{\scriptsize$\pm$0.28} \\
        \midrule
        +$L_2$ (0.00001) & 55.84{\scriptsize$\pm$0.30} & 80.71{\scriptsize$\pm$0.11} & 54.93{\scriptsize$\pm$1.59} & 80.37{\scriptsize$\pm$1.06} & 77.32{\scriptsize$\pm$0.18} \\
        \midrule
        \textbf{+SPCP} & \textbf{55.21{\scriptsize$\pm$0.85}} & \textbf{81.25{\scriptsize$\pm$0.19}} & \textbf{50.59{\scriptsize$\pm$1.83}} & \textbf{82.17{\scriptsize$\pm$1.32}} & 77.70{\scriptsize$\pm$0.20} \\
        \bottomrule[1.5pt]
        \end{tabular}
    }
    \caption{Comparison of $L_2$ weight regularization and SPCP on the CIFAR-100 benchmark.}
    \label{tab:l2}
    \end{center}
\end{table}

\section{Analysis of Computational Overhead}
\label{apd:overhead}
Table \ref{tab:overhead} presents the additional computational overhead introduced by SPCP on both the small-scale CIFAR-10 and the large-scale ImageNet-200 datasets. The results demonstrate that the extra overhead during training and inference is negligible compared to the original computational cost. This efficiency stems from SPCP requiring additional computations only for the parameters in the final classification layer, which constitute a small fraction of the total model parameters. Moreover, as discussed in Appendix \ref{apd:simple}, the contribution of the classifier parameters is efficiently parallelized via a single Hadamard product. Consequently, the increase in computational burden is insignificant.

\begin{table}[t]
    \small
    \centering
    \begin{tabular}{@{\hspace{0.2cm}}c|@{\hspace{0.40cm}}l|@{\hspace{0.40cm}}c@{\hspace{0.43cm}}c@{\hspace{0.2cm}}}
    \toprule
    \multirow{2}{*}{\textbf{Dataset}} & \multirow{2}{*}{\textbf{Methods}} & \multicolumn{2}{c}{\textbf{Latency (s)}} \\
    & & \textbf{Training} & \textbf{Inferring} \\ 
    \midrule
    \multirow{3}{*}{CIFAR-10} & Vanilla & 7.82 & 2.35 \\
    & +LogitNorm & 7.99 & 2.37 \\ 
    & \textbf{+SPCP} & 8.19 & 2.48 \\ 
    \midrule
    \multirow{3}{*}{ImageNet-200} & Vanilla & 515.71 & 10.94 \\ 
     & +LogitNorm & 518.43 & 10.97 \\ 
    & \textbf{+SPCP} & 521.84 & 11.72 \\ 
    \bottomrule
    \end{tabular}
    \caption{Analysis of the training and inference overhead introduced by SPCP. This overhead is measured by latency ($\downarrow$), defined as the time required to process the entire training set during training and the entire test set during inference in a single run. Lower latency indicates reduced computational cost. All experiments are conducted under consistent software and hardware settings as detailed in Appendix \ref{apd:implement}, with results averaged over multiple independent runs.}
    \label{tab:overhead}
\end{table}

\section{Comparison with ITP}
SPCP and ITP differ fundamentally. SPCP builds a solid model basis for OOD detection by addressing sparse contribution patterns, while ITP performs post-hoc model pruning. The experimental results in Table \ref{tab:itp} further validate the superiority of SPCP.

\begin{table}[h]
    \small
    \centering
    \begin{tabular}{@{\hspace{0.1cm}}l|@{\hspace{0.15cm}}c@{\hspace{0.15cm}}c@{\hspace{0.15cm}}c@{\hspace{0.15cm}}c@{\hspace{0.1cm}}}
    \toprule
    & \multicolumn{2}{c}{\textbf{CIFAR-10}} & \multicolumn{2}{c}{\textbf{ImageNet-200}} \\ 
    & Near-OOD & Far-OOD & Near-OOD & Far-OOD \\ 
    \midrule
    ITP & 52.72 / 87.31 & 35.70 / 90.30 & 60.09 / \textbf{82.68} & 32.17 / 91.28 \\
    \textbf{SPCP} & \textbf{31.67 / 91.57} & \textbf{20.54 / 94.55} & \textbf{59.77} / 82.52 & \textbf{30.43 / 91.46} \\ 
    \bottomrule
    \end{tabular}
    \caption{Comparison of OOD Detection Performance between SPCP and ITP.}
    \label{tab:itp}
\end{table}

\section{Comparison with fixed $\lambda$}
This section presents an ablation study on the adaptive updating strategy of $\lambda$, and the results are shown in Table \ref{tab:adaptive}. For comparison, we also adopt a fixed-$\lambda$ strategy, where the fixed value is set to the final value obtained by our adaptive approach. The experimental results demonstrate the effectiveness of our adaptive $\lambda$-updating strategy.

\begin{table}[h]
    \small
    \centering
    \begin{tabular}{@{\hspace{0.2cm}}l|@{\hspace{0.40cm}}c@{\hspace{0.43cm}}c@{\hspace{0.2cm}}}
    \toprule
    & Near-OOD & Far-OOD \\ 
    \midrule
    Fixed & 35.19 / 90.86 & 26.69 / 92.53 \\
    \textbf{Ours} & \textbf{31.67 / 91.57} & \textbf{20.54 / 94.55} \\ 
    \bottomrule
    \end{tabular}
    \caption{Ablation study on the adaptive $\lambda$ conducted on the CIFAR-10 benchmark. Results (FPR95$\downarrow$ / AUROC $\uparrow$) are averaged over multiple runs.}
    \label{tab:adaptive}
\end{table}

\section{Generalization to non-image tasks}
Following the experimental setup of \cite{yangstrengthen}, we extend SPCP to the audio domain (Kinetics-Sound \cite{DBLP:conf/iccv/ArandjelovicZ17}). The audio OOD detection results presented in Table \ref{tab:audio} indicate that SPCP generalizes well beyond image data.

\begin{table}[h]
    \small
    \centering
    \begin{tabular}{@{\hspace{0.2cm}}l|@{\hspace{0.40cm}}c@{\hspace{0.2cm}}}
    \toprule
    & Kinetics-Sound \\ 
    \midrule
    Vanilla & 76.95 / 69.67 \\
    \textbf{Ours} & \textbf{69.20 / 73.48} \\ 
    \bottomrule
    \end{tabular}
    \caption{Extend SPCP to audio OOD detection tasks. Results (FPR95$\downarrow$ / AUROC $\uparrow$) are averaged over multiple runs.}
    \label{tab:audio}
\end{table}

\section{Pseudo Code of SPCP}
The training procedure of SPCP is presented in detail in Algorithm \ref{alg:algorithm}.

\begin{algorithm}[h]
\caption{Training Procedure of SPCP}
\label{alg:algorithm}
\textbf{Input}: Training dataset $\mathcal{D}_{\text{train}}$, percentile $\rho$, smoothing factor $\beta$, initial threshold $\lambda_0$ \\
\textbf{Output}: Trained model $f_\theta$
\begin{algorithmic}[1]
\STATE Initialize the threshold $\lambda \leftarrow \lambda_0$
\WHILE{$f_\theta$ has not converged}
    \STATE Sample a mini-batch $\mathcal{B} \subset \mathcal{D}_{\text{train}}$
    \STATE Compute the contribution matrix $C(\mathbf{x})$ for each $\mathbf{x} \in \mathcal{B}_t$
    \STATE Update the threshold $\lambda$ using the Exponential Moving Average (EMA):
    \[
    \lambda = \beta \cdot \lambda + (1 - \beta) \cdot \frac{1}{|\mathcal{B}|} 
    \sum_{\mathbf{x}_i \in \mathcal{B}} \text{Top}(\rho, C(\mathbf{x}_i))
    \]
    \STATE Truncate the contributions for each $\mathbf{x}_i \in \mathcal{B}$ as $\overline{C}^\lambda(\mathbf{x}_i) = \min(C(\mathbf{x}_i), \lambda)$
    \STATE Update $f_\theta$ by minimizing $\mathcal{R}(f^{\text{SPCP}})$ in Equation~(10) over $\mathcal{B}$
\ENDWHILE
\STATE \textbf{Return} $f_\theta$
\end{algorithmic}
\end{algorithm}

\clearpage
% \newpage
% \vspace*{\fill}
\begin{table*}[t]
    \caption{Fine-grained results (AUROC$\uparrow$) on the CIFAR-10 benchmark.}
    \begin{center}
    \label{apd:r1}
    \resizebox{0.985\linewidth}{!}{
        \begin{tabular}{@{\hspace{0.0cm}}l|@{\hspace{0.15cm}}c@{\hspace{0.15cm}}c@{\hspace{0.15cm}}c@{\hspace{0.15cm}}|c@{\hspace{0.15cm}}c@{\hspace{0.15cm}}c@{\hspace{0.15cm}}c@{\hspace{0.15cm}}c@{\hspace{0.15cm}}|c@{\hspace{0.15cm}}c@{\hspace{0.0cm}}}
        \toprule[1.5pt]
         \multirow{2}{*}{\textbf{Method}} & \multicolumn{3}{c|}{\textbf{Near-OOD}} & \multicolumn{5}{c|}{\textbf{Far-OOD}} & \multirow{2}{*}{\textbf{ID ACC}} \\
        \cline{2-9}
        & \multicolumn{1}{c}{\textbf{CIFAR-100}} & \multicolumn{1}{c}{\textbf{TIN}} & \multicolumn{1}{c|}{\textbf{Average}} & \multicolumn{1}{c}{\textbf{MNIST}} & \multicolumn{1}{c}{\textbf{SVHN}} & \multicolumn{1}{c}{\textbf{Textures}} & \multicolumn{1}{c}{\textbf{Places365}} & \multicolumn{1}{c|}{\textbf{Average}} & \\
        \midrule
        \multicolumn{10}{c}{\textit{Post-hoc methods (vanilla training with cross-entropy)}} \\
        MSP & 87.19{\scriptsize$\pm$0.33} & 88.87{\scriptsize$\pm$0.19} & 88.03{\scriptsize$\pm$0.25} & 92.63{\scriptsize$\pm$1.57} & 91.46{\scriptsize$\pm$0.40} & 89.89{\scriptsize$\pm$0.71} & 88.92{\scriptsize$\pm$0.47} & 90.73{\scriptsize$\pm$0.43} & \underline{95.06{\scriptsize$\pm$0.30}} \\
        Energy & 86.36{\scriptsize$\pm$0.58} & 88.80{\scriptsize$\pm$0.36} & 87.58{\scriptsize$\pm$0.46} & 94.32{\scriptsize$\pm$2.53} & 91.79{\scriptsize$\pm$0.98} & 89.47{\scriptsize$\pm$0.70} & 89.25{\scriptsize$\pm$0.78} & 91.21{\scriptsize$\pm$0.92} & \underline{95.06{\scriptsize$\pm$0.30}} \\
        % VIM & 87.75{\scriptsize$\pm$0.28} & 89.62{\scriptsize$\pm$0.33} & 88.68{\scriptsize$\pm$0.28} & 94.76{\scriptsize$\pm$0.38} & 94.50{\scriptsize$\pm$0.48} & \underline{95.15{\scriptsize$\pm$0.34}} & 89.49{\scriptsize$\pm$0.39} & 93.48{\scriptsize$\pm$0.24} & \underline{95.06{\scriptsize$\pm$0.30}} \\
        DICE & 77.01{\scriptsize$\pm$0.88} & 79.67{\scriptsize$\pm$0.87} & 78.34{\scriptsize$\pm$0.79} & 90.37{\scriptsize$\pm$5.97} & 90.02{\scriptsize$\pm$1.77} & 81.86{\scriptsize$\pm$2.35} & 74.67{\scriptsize$\pm$4.98} & 84.23{\scriptsize$\pm$1.89} & \underline{95.06{\scriptsize$\pm$0.30}} \\
        ReAct & 85.93{\scriptsize$\pm$0.83} & 88.29{\scriptsize$\pm$0.44} & 87.11{\scriptsize$\pm$0.61} & 92.81{\scriptsize$\pm$3.03} & 89.12{\scriptsize$\pm$3.19} & 89.38{\scriptsize$\pm$1.49} & 90.35{\scriptsize$\pm$0.78} & 90.42{\scriptsize$\pm$1.41} & \underline{95.06{\scriptsize$\pm$0.30}} \\
        ASH & 74.11{\scriptsize$\pm$1.55} & 76.44{\scriptsize$\pm$0.61} & 75.27{\scriptsize$\pm$1.04} & 83.16{\scriptsize$\pm$4.66} & 73.46{\scriptsize$\pm$6.41} & 77.45{\scriptsize$\pm$2.39} & 79.89{\scriptsize$\pm$3.69} & 78.49{\scriptsize$\pm$2.58} & \underline{95.06{\scriptsize$\pm$0.30}} \\
        SCALE & 81.27{\scriptsize$\pm$0.74} & 83.84{\scriptsize$\pm$0.02} & 82.55{\scriptsize$\pm$0.36} & 90.58{\scriptsize$\pm$2.96} & 84.63{\scriptsize$\pm$3.05} & 83.94{\scriptsize$\pm$0.34} & 86.41{\scriptsize$\pm$1.92} & 86.39{\scriptsize$\pm$1.86} & \underline{95.06{\scriptsize$\pm$0.30}} \\
        \midrule
        \multicolumn{10}{c}{\textit{Training-time regularization methods}} \\
        % MOS & 70.57{\scriptsize$\pm$3.04} & 72.34{\scriptsize$\pm$3.16} & 71.45{\scriptsize$\pm$3.09} & 74.81{\scriptsize$\pm$10.05} & 73.66{\scriptsize$\pm$9.14} & 70.35{\scriptsize$\pm$3.11} & 86.81{\scriptsize$\pm$1.85} & 76.41{\scriptsize$\pm$5.93} & 94.83{\scriptsize$\pm$0.37} \\
        % VOS & 86.57{\scriptsize$\pm$0.57} & 88.84{\scriptsize$\pm$0.48} & 87.70{\scriptsize$\pm$0.48} & 91.56{\scriptsize$\pm$2.37} & 92.18{\scriptsize$\pm$1.65} & 89.68{\scriptsize$\pm$1.32} & 89.90{\scriptsize$\pm$0.66} & 90.83{\scriptsize$\pm$0.92} & 94.31{\scriptsize$\pm$0.64} \\
        LogitNorm & \underline{90.95{\scriptsize$\pm$0.22}} & \underline{93.70{\scriptsize$\pm$0.06}} & \underline{92.33{\scriptsize$\pm$0.08}} & \underline{99.14{\scriptsize$\pm$0.45}} & \underline{98.25{\scriptsize$\pm$0.41}} & 94.77{\scriptsize$\pm$0.43} & \textbf{94.79{\scriptsize$\pm$0.16}} & \underline{96.74{\scriptsize$\pm$0.06}} & 94.30{\scriptsize$\pm$0.25} \\
        CIDER & 89.47{\scriptsize$\pm$0.19} & 91.94{\scriptsize$\pm$0.19} & 90.71{\scriptsize$\pm$0.16} & 93.30{\scriptsize$\pm$1.08} & \underline{98.06{\scriptsize$\pm$0.07}} & 93.71{\scriptsize$\pm$0.39} & \underline{93.77{\scriptsize$\pm$0.68}} & \underline{94.71{\scriptsize$\pm$0.36}} & -- \\
        UMAP & 89.67{\scriptsize$\pm$0.18} & 92.33{\scriptsize$\pm$0.09} & 91.00{\scriptsize$\pm$0.07} & \underline{97.70{\scriptsize$\pm$0.53}} & 94.30{\scriptsize$\pm$1.69} & 91.23{\scriptsize$\pm$0.32} & \underline{93.58{\scriptsize$\pm$0.48}} & 94.20{\scriptsize$\pm$0.36} & \underline{95.06{\scriptsize$\pm$0.30}} \\
        SNN & 89.32{\scriptsize$\pm$0.09} & 91.18{\scriptsize$\pm$0.09} & 90.25{\scriptsize$\pm$0.09} & 93.53{\scriptsize$\pm$1.04} & 91.98{\scriptsize$\pm$1.76} & 92.94{\scriptsize$\pm$0.34} & 91.49{\scriptsize$\pm$0.23} & 92.49{\scriptsize$\pm$0.78} & \textbf{95.11{\scriptsize$\pm$0.13}} \\
        T2FNorm & \textbf{91.56{\scriptsize$\pm$0.10}} & \textbf{94.02{\scriptsize$\pm$0.16}} & \textbf{92.79{\scriptsize$\pm$0.13}} & \textbf{99.28{\scriptsize$\pm$0.27}} & \textbf{98.81{\scriptsize$\pm$0.22}} & \textbf{95.44{\scriptsize$\pm$0.77}} & 94.40{\scriptsize$\pm$0.32} & \textbf{96.98{\scriptsize$\pm$0.23}} & 94.69{\scriptsize$\pm$0.07} \\
        \midrule
        SPCP & \underline{90.42{\scriptsize$\pm$0.17}} & \underline{92.72{\scriptsize$\pm$0.07}} & \underline{91.57{\scriptsize$\pm$0.11}} & 94.76{\scriptsize$\pm$0.69} & 95.35{\scriptsize$\pm$1.33} & \underline{95.00{\scriptsize$\pm$0.03}} & 93.09{\scriptsize$\pm$0.12} & 94.55{\scriptsize$\pm$0.42} & \underline{94.91{\scriptsize$\pm$0.27}} \\
        \bottomrule[1.5pt]
        \end{tabular}
    }
    \end{center}
\end{table*}
% \vspace{2cm}
\begin{table*}[t]
    \caption{Fine-grained results (FPR95$\downarrow$) on the CIFAR-10 benchmark.}
    \begin{center}
    \label{apd:r2}
    \resizebox{0.985\linewidth}{!}{
        \begin{tabular}{@{\hspace{0.0cm}}l|@{\hspace{0.15cm}}c@{\hspace{0.15cm}}c@{\hspace{0.15cm}}c@{\hspace{0.15cm}}|c@{\hspace{0.15cm}}c@{\hspace{0.15cm}}c@{\hspace{0.15cm}}c@{\hspace{0.15cm}}c@{\hspace{0.15cm}}|c@{\hspace{0.15cm}}c@{\hspace{0.0cm}}}
        \toprule[1.5pt]
         \multirow{2}{*}{\textbf{Method}} & \multicolumn{3}{c|}{\textbf{Near-OOD}} & \multicolumn{5}{c|}{\textbf{Far-OOD}} & \multirow{2}{*}{\textbf{ID ACC}} \\
        \cline{2-9}
        & \multicolumn{1}{c}{\textbf{CIFAR-100}} & \multicolumn{1}{c}{\textbf{TIN}} & \multicolumn{1}{c|}{\textbf{Average}} & \multicolumn{1}{c}{\textbf{MNIST}} & \multicolumn{1}{c}{\textbf{SVHN}} & \multicolumn{1}{c}{\textbf{Textures}} & \multicolumn{1}{c}{\textbf{Places365}} & \multicolumn{1}{c|}{\textbf{Average}} & \\
        \midrule
        \multicolumn{10}{c}{\textit{Post-hoc methods (vanilla training with cross-entropy)}} \\
        MSP & 53.08{\scriptsize$\pm$4.86} & 43.27{\scriptsize$\pm$3.00} & 48.17{\scriptsize$\pm$3.92} & 23.64{\scriptsize$\pm$5.81} & 25.82{\scriptsize$\pm$1.64} & 34.96{\scriptsize$\pm$4.64} & 42.47{\scriptsize$\pm$3.81} & 31.72{\scriptsize$\pm$1.84} & \underline{95.06{\scriptsize$\pm$0.30}} \\
        Energy & 66.60{\scriptsize$\pm$4.46} & 56.08{\scriptsize$\pm$4.83} & 61.34{\scriptsize$\pm$4.63} & 24.99{\scriptsize$\pm$12.93} & 35.12{\scriptsize$\pm$6.11} & 51.82{\scriptsize$\pm$6.11} & 54.85{\scriptsize$\pm$6.52} & 41.69{\scriptsize$\pm$5.32} & \underline{95.06{\scriptsize$\pm$0.30}} \\
        % VIM & 49.19{\scriptsize$\pm$3.15} & 40.49{\scriptsize$\pm$1.55} & 44.84{\scriptsize$\pm$2.31} & 18.36{\scriptsize$\pm$1.42} & 19.29{\scriptsize$\pm$0.41} & \underline{21.14{\scriptsize$\pm$1.83}} & 41.43{\scriptsize$\pm$2.17} & 25.05{\scriptsize$\pm$0.52} & \underline{95.06{\scriptsize$\pm$0.30}} \\
        DICE & 73.71{\scriptsize$\pm$7.67} & 66.37{\scriptsize$\pm$7.68} & 70.04{\scriptsize$\pm$7.64} & 30.83{\scriptsize$\pm$10.54} & 36.61{\scriptsize$\pm$4.74} & 62.42{\scriptsize$\pm$4.79} & 77.19{\scriptsize$\pm$12.60} & 51.76{\scriptsize$\pm$4.42} & \underline{95.06{\scriptsize$\pm$0.30}} \\
        ReAct & 67.40{\scriptsize$\pm$7.34} & 59.71{\scriptsize$\pm$7.31} & 63.56{\scriptsize$\pm$7.33} & 33.77{\scriptsize$\pm$18.00} & 50.23{\scriptsize$\pm$15.98} & 51.42{\scriptsize$\pm$11.42} & 44.20{\scriptsize$\pm$3.35} & 44.90{\scriptsize$\pm$8.37} & \underline{95.06{\scriptsize$\pm$0.30}} \\
        ASH & 87.31{\scriptsize$\pm$2.06} & 86.25{\scriptsize$\pm$1.58} & 86.78{\scriptsize$\pm$1.82} & 70.00{\scriptsize$\pm$10.56} & 83.64{\scriptsize$\pm$6.48} & 84.59{\scriptsize$\pm$1.74} & 77.89{\scriptsize$\pm$7.28} & 79.03{\scriptsize$\pm$4.22} & \underline{95.06{\scriptsize$\pm$0.30}} \\
        SCALE & 81.79{\scriptsize$\pm$3.98} & 79.12{\scriptsize$\pm$4.06} & 80.45{\scriptsize$\pm$4.02} & 48.69{\scriptsize$\pm$15.39} & 70.55{\scriptsize$\pm$8.68} & 80.38{\scriptsize$\pm$3.91} & 70.51{\scriptsize$\pm$5.98} & 67.53{\scriptsize$\pm$7.50} & \underline{95.06{\scriptsize$\pm$0.30}} \\
        \midrule
        \multicolumn{10}{c}{\textit{Training-time regularization methods}} \\
        % MOS & 79.38{\scriptsize$\pm$5.06} & 78.05{\scriptsize$\pm$6.69} & 78.72{\scriptsize$\pm$5.86} & 65.95{\scriptsize$\pm$17.54} & 57.79{\scriptsize$\pm$5.79} & 76.78{\scriptsize$\pm$3.86} & 51.09{\scriptsize$\pm$1.33} & 62.90{\scriptsize$\pm$6.62} & 94.83{\scriptsize$\pm$0.37} \\
        % VOS & 61.57{\scriptsize$\pm$3.24} & 52.49{\scriptsize$\pm$0.73} & 57.03{\scriptsize$\pm$1.92} & 35.92{\scriptsize$\pm$11.38} & 31.50{\scriptsize$\pm$8.38} & 46.53{\scriptsize$\pm$5.24} & 47.78{\scriptsize$\pm$3.62} & 40.43{\scriptsize$\pm$4.53} & 94.31{\scriptsize$\pm$0.64} \\
        LogitNorm & \underline{34.37{\scriptsize$\pm$1.30}} & \underline{24.30{\scriptsize$\pm$0.54}} & \underline{29.34{\scriptsize$\pm$0.81}} & \underline{3.93{\scriptsize$\pm$1.99}} & \underline{8.33{\scriptsize$\pm$1.78}} & \underline{21.94{\scriptsize$\pm$0.85}} & \textbf{21.04{\scriptsize$\pm$0.71}} & \underline{13.81{\scriptsize$\pm$0.20}} & 94.30{\scriptsize$\pm$0.25} \\
        CIDER & 35.60{\scriptsize$\pm$0.78} & 28.61{\scriptsize$\pm$1.10} & 32.11{\scriptsize$\pm$0.94} & 24.76{\scriptsize$\pm$2.82} & \underline{8.04{\scriptsize$\pm$0.43}} & 25.05{\scriptsize$\pm$3.29} & \underline{25.03{\scriptsize$\pm$1.36}} & 20.72{\scriptsize$\pm$0.85} & -- \\
        UMAP & 37.49{\scriptsize$\pm$0.42} & 28.54{\scriptsize$\pm$0.44} & 33.01{\scriptsize$\pm$0.06} & \underline{8.59{\scriptsize$\pm$1.81}} & 20.11{\scriptsize$\pm$6.72} & 31.60{\scriptsize$\pm$1.08} & 26.49{\scriptsize$\pm$1.88} & 21.70{\scriptsize$\pm$1.57} & \underline{95.06{\scriptsize$\pm$0.30}} \\
        SNN & 41.00{\scriptsize$\pm$0.76} & 33.43{\scriptsize$\pm$0.82} & 37.21{\scriptsize$\pm$0.70} & 22.92{\scriptsize$\pm$3.86} & 25.33{\scriptsize$\pm$3.64} & 24.34{\scriptsize$\pm$1.77} & 31.60{\scriptsize$\pm$1.04} & 26.05{\scriptsize$\pm$2.34} & \textbf{95.11{\scriptsize$\pm$0.13}} \\
        T2FNorm & \textbf{30.60{\scriptsize$\pm$0.45}} & \textbf{22.33{\scriptsize$\pm$0.37}} & \textbf{26.47{\scriptsize$\pm$0.35}} & \textbf{3.50{\scriptsize$\pm$1.33}} & \textbf{5.72{\scriptsize$\pm$0.66}} & \textbf{19.49{\scriptsize$\pm$2.58}} & \underline{22.27{\scriptsize$\pm$1.28}} & \textbf{12.75{\scriptsize$\pm$0.73}} & 94.69{\scriptsize$\pm$0.07} \\
        \midrule
        SPCP & \underline{35.74{\scriptsize$\pm$0.51}} & \underline{27.59{\scriptsize$\pm$0.48}} & \underline{31.67{\scriptsize$\pm$0.25}} & 19.09{\scriptsize$\pm$1.14} & 16.52{\scriptsize$\pm$2.74} & \underline{20.09{\scriptsize$\pm$0.57}} & 26.47{\scriptsize$\pm$0.54} & \underline{20.54{\scriptsize$\pm$0.67}} & \underline{94.91{\scriptsize$\pm$0.27}} \\
        \bottomrule[1.5pt]
        \end{tabular}
    }
    \end{center}
\end{table*}
% \vspace*{\fill}
% \newpage
% \vspace*{\fill}
\begin{table*}[t]
    \caption{Fine-grained results (AUROC$\uparrow$) on the CIFAR-100 benchmark.}
    \begin{center}
    \label{apd:r3}
    \resizebox{0.985\linewidth}{!}{
        \begin{tabular}{@{\hspace{0.0cm}}l|@{\hspace{0.15cm}}c@{\hspace{0.15cm}}c@{\hspace{0.15cm}}c@{\hspace{0.15cm}}|c@{\hspace{0.15cm}}c@{\hspace{0.15cm}}c@{\hspace{0.15cm}}c@{\hspace{0.15cm}}c@{\hspace{0.15cm}}|c@{\hspace{0.15cm}}c@{\hspace{0.0cm}}}
        \toprule[1.5pt]
         \multirow{2}{*}{\textbf{Method}} & \multicolumn{3}{c|}{\textbf{Near-OOD}} & \multicolumn{5}{c|}{\textbf{Far-OOD}} & \multirow{2}{*}{\textbf{ID ACC}} \\
        \cline{2-9}
        & \multicolumn{1}{c}{\textbf{CIFAR-10}} & \multicolumn{1}{c}{\textbf{TIN}} & \multicolumn{1}{c|}{\textbf{Average}} & \multicolumn{1}{c}{\textbf{MNIST}} & \multicolumn{1}{c}{\textbf{SVHN}} & \multicolumn{1}{c}{\textbf{Textures}} & \multicolumn{1}{c}{\textbf{Places365}} & \multicolumn{1}{c|}{\textbf{Average}} & \\
        \midrule
        \multicolumn{10}{c}{\textit{Post-hoc methods (vanilla training with cross-entropy)}} \\
        MSP & 78.47{\scriptsize$\pm$0.07} & 82.07{\scriptsize$\pm$0.17} & 80.27{\scriptsize$\pm$0.11} & 76.08{\scriptsize$\pm$1.86} & 78.42{\scriptsize$\pm$0.89} & 77.32{\scriptsize$\pm$0.71} & 79.22{\scriptsize$\pm$0.29} & 77.76{\scriptsize$\pm$0.44} & \underline{77.25{\scriptsize$\pm$0.10}} \\
        Energy & \underline{79.05{\scriptsize$\pm$0.11}}
        & 82.76{\scriptsize$\pm$0.08} & \underline{80.91{\scriptsize$\pm$0.08}} & 79.18{\scriptsize$\pm$1.37} & 82.03{\scriptsize$\pm$1.74} & 78.35{\scriptsize$\pm$0.83} & 79.52{\scriptsize$\pm$0.23} & 79.77{\scriptsize$\pm$0.61} & \underline{77.25{\scriptsize$\pm$0.10}} \\
        % VIM & 72.21{\scriptsize$\pm$0.41} & 77.76{\scriptsize$\pm$0.16} & 74.98{\scriptsize$\pm$0.13} & 81.89{\scriptsize$\pm$1.02} & 83.14{\scriptsize$\pm$3.71} & \textbf{85.91{\scriptsize$\pm$0.78}} & 75.85{\scriptsize$\pm$0.37} & 81.70{\scriptsize$\pm$0.62} & \underline{77.25{\scriptsize$\pm$0.10}} \\
        DICE & 78.04{\scriptsize$\pm$0.32} & 80.72{\scriptsize$\pm$0.30} & 79.38{\scriptsize$\pm$0.23} & 79.86{\scriptsize$\pm$1.89} & 84.22{\scriptsize$\pm$2.00} & 77.63{\scriptsize$\pm$0.34} & 78.33{\scriptsize$\pm$0.66} & 80.01{\scriptsize$\pm$0.18} & \underline{77.25{\scriptsize$\pm$0.10}} \\
        ReAct & 78.65{\scriptsize$\pm$0.05} & 82.88{\scriptsize$\pm$0.08} & 80.77{\scriptsize$\pm$0.05} & 78.37{\scriptsize$\pm$1.59} & 83.01{\scriptsize$\pm$0.97} & 80.15{\scriptsize$\pm$0.46} & 80.03{\scriptsize$\pm$0.11} & 80.39{\scriptsize$\pm$0.49} & \underline{77.25{\scriptsize$\pm$0.10}} \\
        ASH & 76.48{\scriptsize$\pm$0.30} & 79.92{\scriptsize$\pm$0.20} & 78.20{\scriptsize$\pm$0.15} & 77.23{\scriptsize$\pm$0.46} & 85.60{\scriptsize$\pm$1.40} & 80.72{\scriptsize$\pm$0.70} & 78.76{\scriptsize$\pm$0.16} & 80.58{\scriptsize$\pm$0.66} & \underline{77.25{\scriptsize$\pm$0.10}} \\
        SCALE & \textbf{79.26{\scriptsize$\pm$0.10}} & 82.71{\scriptsize$\pm$0.14} & \underline{80.99{\scriptsize$\pm$0.12}} & 80.27{\scriptsize$\pm$1.16} & 84.45{\scriptsize$\pm$1.09} & 80.50{\scriptsize$\pm$0.70} & \underline{80.47{\scriptsize$\pm$0.23}} & 81.42{\scriptsize$\pm$0.43} & \underline{77.25{\scriptsize$\pm$0.10}} \\
        \midrule
        \multicolumn{10}{c}{\textit{Training-time regularization methods}} \\
        % MOS & 78.54{\scriptsize$\pm$0.13} & 82.26{\scriptsize$\pm$0.25} & 80.40{\scriptsize$\pm$0.18} & 80.68{\scriptsize$\pm$1.65} & 81.59{\scriptsize$\pm$3.81} & 79.92{\scriptsize$\pm$0.57} & 78.50{\scriptsize$\pm$0.34} & 80.17{\scriptsize$\pm$1.21} & 76.98{\scriptsize$\pm$0.20} \\
        % VOS & \underline{79.14{\scriptsize$\pm$0.41}} & 82.73{\scriptsize$\pm$0.20} & \underline{80.93{\scriptsize$\pm$0.29}} & 82.29{\scriptsize$\pm$1.51} & 84.23{\scriptsize$\pm$1.33} & 78.41{\scriptsize$\pm$0.78} & 80.34{\scriptsize$\pm$0.03} & 81.32{\scriptsize$\pm$0.09} & 77.20{\scriptsize$\pm$0.10} \\
        LogitNorm & 74.57{\scriptsize$\pm$0.39} & 82.37{\scriptsize$\pm$0.24} & 78.47{\scriptsize$\pm$0.31} & \textbf{90.69{\scriptsize$\pm$1.38}} & 82.80{\scriptsize$\pm$4.57} & 72.37{\scriptsize$\pm$0.67} & 80.25{\scriptsize$\pm$0.61} & 81.53{\scriptsize$\pm$1.26} & 76.34{\scriptsize$\pm$0.17} \\
        CIDER & 67.55{\scriptsize$\pm$0.60} & 78.65{\scriptsize$\pm$0.35} & 73.10{\scriptsize$\pm$0.39} & 68.14{\scriptsize$\pm$3.98} & \textbf{97.17{\scriptsize$\pm$0.34}} & \underline{82.21{\scriptsize$\pm$1.93}} & 74.43{\scriptsize$\pm$0.64} & 80.49{\scriptsize$\pm$0.68} & -- \\
        UMAP & 76.96{\scriptsize$\pm$0.46} & 82.01{\scriptsize$\pm$0.15} & 79.49{\scriptsize$\pm$0.23} & \underline{84.90{\scriptsize$\pm$4.24}} & 85.25{\scriptsize$\pm$3.16} & 78.16{\scriptsize$\pm$1.36} & 78.17{\scriptsize$\pm$0.42} & 81.62{\scriptsize$\pm$1.37} & \underline{77.25{\scriptsize$\pm$0.10}} \\
        SNN & 76.99{\scriptsize$\pm$0.25} & \textbf{83.68{\scriptsize$\pm$0.20}} & 80.33{\scriptsize$\pm$0.22} & 82.90{\scriptsize$\pm$1.21} & 83.06{\scriptsize$\pm$2.44} & \textbf{82.98{\scriptsize$\pm$0.37}} & 79.74{\scriptsize$\pm$0.27} & \underline{82.17{\scriptsize$\pm$0.69}} & \underline{77.56{\scriptsize$\pm$0.27}} \\
        T2FNorm & 76.09{\scriptsize$\pm$0.81} & \underline{83.59{\scriptsize$\pm$0.02}} & 79.84{\scriptsize$\pm$0.40} & \underline{86.22{\scriptsize$\pm$2.29}} & \underline{86.04{\scriptsize$\pm$1.04}} & 77.32{\scriptsize$\pm$1.63} & \textbf{81.35{\scriptsize$\pm$0.33}} & \textbf{82.73{\scriptsize$\pm$1.01}} & 76.43{\scriptsize$\pm$0.13} \\
        \midrule
        SPCP & \underline{79.09{\scriptsize$\pm$0.22}} & \underline{83.40{\scriptsize$\pm$0.28}} & \textbf{81.25{\scriptsize$\pm$0.19}} & 80.98{\scriptsize$\pm$2.70} & \underline{86.02{\scriptsize$\pm$3.60}} & \underline{81.32{\scriptsize$\pm$0.28}} & \underline{80.37{\scriptsize$\pm$0.38}} & \underline{82.17{\scriptsize$\pm$1.32}} & \textbf{77.70{\scriptsize$\pm$0.20}} \\
        \bottomrule[1.5pt]
        \end{tabular}
    }
    \end{center}
\end{table*}
% \vspace{2cm}
\begin{table*}[t]
    \caption{Fine-grained results (FPR95$\downarrow$) on the CIFAR-100 benchmark.}
    \begin{center}
    \label{apd:r4}
    \resizebox{0.985\linewidth}{!}{
        \begin{tabular}{@{\hspace{0.0cm}}l|@{\hspace{0.15cm}}c@{\hspace{0.15cm}}c@{\hspace{0.15cm}}c@{\hspace{0.15cm}}|c@{\hspace{0.15cm}}c@{\hspace{0.15cm}}c@{\hspace{0.15cm}}c@{\hspace{0.15cm}}c@{\hspace{0.15cm}}|c@{\hspace{0.15cm}}c@{\hspace{0.0cm}}}
        \toprule
         \multirow{2}{*}{\textbf{Method}} & \multicolumn{3}{c|}{\textbf{Near-OOD}} & \multicolumn{5}{c|}{\textbf{Far-OOD}} & \multirow{2}{*}{\textbf{ID ACC}} \\
        \cline{2-9}
        & \multicolumn{1}{c}{\textbf{CIFAR-10}} & \multicolumn{1}{c}{\textbf{TIN}} & \multicolumn{1}{c|}{\textbf{Average}} & \multicolumn{1}{c}{\textbf{MNIST}} & \multicolumn{1}{c}{\textbf{SVHN}} & \multicolumn{1}{c}{\textbf{Textures}} & \multicolumn{1}{c}{\textbf{Places365}} & \multicolumn{1}{c|}{\textbf{Average}} & \\
        \midrule
        \multicolumn{10}{c}{\textit{Post-hoc methods (vanilla training with cross-entropy)}} \\
        MSP & \textbf{58.91{\scriptsize$\pm$0.93}} & 50.70{\scriptsize$\pm$0.34} & \textbf{54.80{\scriptsize$\pm$0.33}} & 57.23{\scriptsize$\pm$4.68} & 59.07{\scriptsize$\pm$2.53} & 61.88{\scriptsize$\pm$1.28} & 56.62{\scriptsize$\pm$0.87} & 58.70{\scriptsize$\pm$1.06} & \underline{77.25{\scriptsize$\pm$0.10}} \\
        Energy & \underline{59.21{\scriptsize$\pm$0.75}} & 52.03{\scriptsize$\pm$0.50} & \underline{55.62{\scriptsize$\pm$0.61}} & 52.62{\scriptsize$\pm$3.83} & 53.62{\scriptsize$\pm$3.14} & 62.35{\scriptsize$\pm$2.06} & 57.75{\scriptsize$\pm$0.86} & 56.59{\scriptsize$\pm$1.38} & \underline{77.25{\scriptsize$\pm$0.10}} \\
        % VIM & 70.59{\scriptsize$\pm$0.43} & 54.66{\scriptsize$\pm$0.42} & 62.63{\scriptsize$\pm$0.27} & 48.32{\scriptsize$\pm$1.07} & 46.22{\scriptsize$\pm$5.46} & \textbf{46.86{\scriptsize$\pm$2.29}} & 61.57{\scriptsize$\pm$0.77} & \underline{50.74{\scriptsize$\pm$1.00}} & \underline{77.25{\scriptsize$\pm$0.10}} \\
        DICE & 60.98{\scriptsize$\pm$1.10} & 54.93{\scriptsize$\pm$0.53} & 57.95{\scriptsize$\pm$0.53} & 51.79{\scriptsize$\pm$3.67} & 49.58{\scriptsize$\pm$3.32} & 64.23{\scriptsize$\pm$1.65} & 59.39{\scriptsize$\pm$1.25} & 56.25{\scriptsize$\pm$0.60} & \underline{77.25{\scriptsize$\pm$0.10}} \\
        ReAct & 61.30{\scriptsize$\pm$0.43} & 51.47{\scriptsize$\pm$0.47} & 56.39{\scriptsize$\pm$0.34} & 56.04{\scriptsize$\pm$5.66} & 50.41{\scriptsize$\pm$2.02} & 55.04{\scriptsize$\pm$0.82} & 55.30{\scriptsize$\pm$0.41} & 54.20{\scriptsize$\pm$1.56} & \underline{77.25{\scriptsize$\pm$0.10}} \\
        ASH & 68.06{\scriptsize$\pm$0.44} & 63.35{\scriptsize$\pm$0.90} & 65.71{\scriptsize$\pm$0.24} & 66.58{\scriptsize$\pm$3.88} & 46.00{\scriptsize$\pm$2.67} & 61.27{\scriptsize$\pm$2.74} & 62.95{\scriptsize$\pm$0.99} & 59.20{\scriptsize$\pm$2.46} & \underline{77.25{\scriptsize$\pm$0.10}} \\
        SCALE & \underline{59.11{\scriptsize$\pm$0.83}} & 52.26{\scriptsize$\pm$0.72} & 55.68{\scriptsize$\pm$0.69} & 51.64{\scriptsize$\pm$3.57} & 49.27{\scriptsize$\pm$2.60} & 58.46{\scriptsize$\pm$1.96} & 56.97{\scriptsize$\pm$0.68} & 54.09{\scriptsize$\pm$1.07} & \underline{77.25{\scriptsize$\pm$0.10}} \\
        \midrule
        \multicolumn{10}{c}{\textit{Training-time regularization methods}} \\
        % MOS & 60.60{\scriptsize$\pm$1.47} & 51.49{\scriptsize$\pm$0.69} & 56.05{\scriptsize$\pm$1.01} & 52.70{\scriptsize$\pm$3.81} & 56.33{\scriptsize$\pm$8.46} & 61.24{\scriptsize$\pm$2.06} & 58.86{\scriptsize$\pm$0.41} & 57.28{\scriptsize$\pm$3.29} & 76.98{\scriptsize$\pm$0.20} \\
        % VOS & 59.23{\scriptsize$\pm$0.59} & 51.89{\scriptsize$\pm$1.01} & \underline{55.56{\scriptsize$\pm$0.77}} & 48.56{\scriptsize$\pm$2.00} & 47.23{\scriptsize$\pm$3.04} & 62.55{\scriptsize$\pm$1.04} & 56.44{\scriptsize$\pm$0.36} & 53.70{\scriptsize$\pm$0.38} & 77.20{\scriptsize$\pm$0.10} \\
        LogitNorm & 73.88{\scriptsize$\pm$1.21} & 51.89{\scriptsize$\pm$0.10} & 62.89{\scriptsize$\pm$0.57} & \textbf{34.12{\scriptsize$\pm$8.32}} & 47.52{\scriptsize$\pm$8.02} & 77.38{\scriptsize$\pm$2.99} & \underline{55.44{\scriptsize$\pm$1.45}} & 53.61{\scriptsize$\pm$3.45} & 76.34{\scriptsize$\pm$0.17} \\
        CIDER & 82.71{\scriptsize$\pm$1.25} & 61.33{\scriptsize$\pm$0.64} & 72.02{\scriptsize$\pm$0.31} & 75.32{\scriptsize$\pm$4.21} & \textbf{17.82{\scriptsize$\pm$2.80}} & \underline{54.43{\scriptsize$\pm$2.56}} & 69.30{\scriptsize$\pm$1.81} & 54.22{\scriptsize$\pm$1.24} & -- \\
        UMAP & 64.73{\scriptsize$\pm$0.97} & 54.70{\scriptsize$\pm$0.33} & 59.71{\scriptsize$\pm$0.65} & \underline{42.80{\scriptsize$\pm$7.56}} & \underline{40.03{\scriptsize$\pm$6.52}} & 65.02{\scriptsize$\pm$4.09} & 60.61{\scriptsize$\pm$1.61} & \underline{52.11{\scriptsize$\pm$2.36}} & \underline{77.25{\scriptsize$\pm$0.10}} \\
        SNN & 71.56{\scriptsize$\pm$2.17} & \textbf{49.08{\scriptsize$\pm$0.73}} & 60.32{\scriptsize$\pm$1.44} & 46.31{\scriptsize$\pm$1.23} & 54.11{\scriptsize$\pm$6.62} & \textbf{54.14{\scriptsize$\pm$1.30}} & 59.51{\scriptsize$\pm$0.91} & 53.52{\scriptsize$\pm$1.77} & \underline{77.56{\scriptsize$\pm$0.27}} \\
        T2FNorm & 67.07{\scriptsize$\pm$1.90} & \underline{49.88{\scriptsize$\pm$0.85}} & 58.47{\scriptsize$\pm$1.35} & \underline{39.39{\scriptsize$\pm$5.38}} & 44.29{\scriptsize$\pm$3.14} & 66.82{\scriptsize$\pm$4.61} & \textbf{54.50{\scriptsize$\pm$0.52}} & \underline{51.25{\scriptsize$\pm$2.52}} & 76.43{\scriptsize$\pm$0.13} \\
        \midrule
        SPCP & 60.10{\scriptsize$\pm$0.95} & \underline{50.33{\scriptsize$\pm$0.85}} & \underline{55.21{\scriptsize$\pm$0.85}} & 49.33{\scriptsize$\pm$3.55} & \underline{43.00{\scriptsize$\pm$6.73}} & \underline{54.87{\scriptsize$\pm$1.21}} & \underline{55.16{\scriptsize$\pm$0.61}} & \textbf{50.59{\scriptsize$\pm$1.83}} & \textbf{77.20{\scriptsize$\pm$0.20}} \\
        \bottomrule
        \end{tabular}
    }
    \end{center}
\end{table*}
% \vspace*{\fill}
% \newpage
% \vspace*{\fill}
\begin{table*}[t]
    \caption{Fine-grained results (AUROC$\uparrow$) on the ImageNet-200 benchmark.}
    \begin{center}
    \label{apd:r5}
    \resizebox{0.985\linewidth}{!}{
        \begin{tabular}{@{\hspace{0.0cm}}l|@{\hspace{0.15cm}}c@{\hspace{0.15cm}}c@{\hspace{0.15cm}}c@{\hspace{0.15cm}}|c@{\hspace{0.15cm}}c@{\hspace{0.15cm}}c@{\hspace{0.15cm}}c@{\hspace{0.15cm}}|c@{\hspace{0.15cm}}c@{\hspace{0.0cm}}}
        \toprule[1.5pt]
         \multirow{2}{*}{\textbf{Method}} & \multicolumn{3}{c|}{\textbf{Near-OOD}} & \multicolumn{4}{c|}{\textbf{Far-OOD}} & \multirow{2}{*}{\textbf{ID ACC}} \\
        \cline{2-8}
        & \multicolumn{1}{c}{\textbf{SSB-hard}} & \multicolumn{1}{c}{\textbf{NINCO}} & \multicolumn{1}{c|}{\textbf{Average}} & \multicolumn{1}{c}{\textbf{iNaturalist}} & \multicolumn{1}{c}{\textbf{Textures}} & \multicolumn{1}{c}{\textbf{OpenImage-O}} & \multicolumn{1}{c|}{\textbf{Average}} & \\
        \midrule
        \multicolumn{9}{c}{\textit{Post-hoc methods (vanilla training with cross-entropy)}} \\
        MSP & \underline{80.38{\scriptsize$\pm$0.03}} & 86.29{\scriptsize$\pm$0.11} & \underline{83.34{\scriptsize$\pm$0.06}} & 92.80{\scriptsize$\pm$0.25} & 88.36{\scriptsize$\pm$0.13} & 89.24{\scriptsize$\pm$0.02} & 90.13{\scriptsize$\pm$0.09} & 86.37{\scriptsize$\pm$0.08} \\
        Energy & \underline{79.83{\scriptsize$\pm$0.02}} & 85.17{\scriptsize$\pm$0.11} & 82.50{\scriptsize$\pm$0.05} & 92.55{\scriptsize$\pm$0.50} & 90.79{\scriptsize$\pm$0.16} & 89.23{\scriptsize$\pm$0.26} & 90.86{\scriptsize$\pm$0.21} & 86.37{\scriptsize$\pm$0.08} \\
        % VIM & 74.04{\scriptsize$\pm$0.31} & 83.32{\scriptsize$\pm$0.19} & 78.68{\scriptsize$\pm$0.24} & 90.96{\scriptsize$\pm$0.36} & \underline{94.61{\scriptsize$\pm$0.12}} & 88.20{\scriptsize$\pm$0.18} & 91.26{\scriptsize$\pm$0.19} & 86.37{\scriptsize$\pm$0.08} \\
        DICE & 79.06{\scriptsize$\pm$0.05} & 84.49{\scriptsize$\pm$0.24} & 81.78{\scriptsize$\pm$0.14} & 91.81{\scriptsize$\pm$0.79} & 91.53{\scriptsize$\pm$0.21} & 89.06{\scriptsize$\pm$0.34} & 90.80{\scriptsize$\pm$0.31} & 86.37{\scriptsize$\pm$0.08} \\
        ReAct & 78.97{\scriptsize$\pm$1.33} & 84.76{\scriptsize$\pm$0.64} & 81.87{\scriptsize$\pm$0.98} & 93.65{\scriptsize$\pm$0.88} & 92.86{\scriptsize$\pm$0.47} & 90.40{\scriptsize$\pm$0.35} & 92.31{\scriptsize$\pm$0.56} & 86.37{\scriptsize$\pm$0.08} \\
        ASH & 79.52{\scriptsize$\pm$0.37} & 85.24{\scriptsize$\pm$0.08} & 82.38{\scriptsize$\pm$0.19} & 95.10{\scriptsize$\pm$0.47} & \textbf{94.77{\scriptsize$\pm$0.19}} & 91.82{\scriptsize$\pm$0.25} & \underline{93.90{\scriptsize$\pm$0.27}} & 86.37{\scriptsize$\pm$0.08} \\
        SCALE & \textbf{82.08{\scriptsize$\pm$0.27}} & \textbf{87.59{\scriptsize$\pm$0.30}} & \textbf{84.84{\scriptsize$\pm$0.28}} & 95.79{\scriptsize$\pm$0.46} & 94.10{\scriptsize$\pm$0.07} & \underline{92.04{\scriptsize$\pm$0.25}} & \underline{93.98{\scriptsize$\pm$0.25}} & 86.37{\scriptsize$\pm$0.08} \\
        \midrule
        \multicolumn{9}{c}{\textit{Training-time regularization methods}} \\
        % MOS & 66.54{\scriptsize$\pm$0.49} & 73.14{\scriptsize$\pm$0.47} & 69.84{\scriptsize$\pm$0.46} & 79.69{\scriptsize$\pm$1.38} & 81.38{\scriptsize$\pm$0.75} & 80.29{\scriptsize$\pm$0.68} & 80.46{\scriptsize$\pm$0.92} & 85.60{\scriptsize$\pm$0.20} \\
        % VOS & 79.68{\scriptsize$\pm$0.19} & 85.35{\scriptsize$\pm$0.10} & 82.51{\scriptsize$\pm$0.11} & 92.77{\scriptsize$\pm$0.54} & 90.95{\scriptsize$\pm$0.20} & 89.28{\scriptsize$\pm$0.15} & 91.00{\scriptsize$\pm$0.28} & 86.23{\scriptsize$\pm$0.19} \\
        LogitNorm & 78.42{\scriptsize$\pm$0.23} & 86.90{\scriptsize$\pm$0.07} & 82.66{\scriptsize$\pm$0.15} & \underline{96.26{\scriptsize$\pm$0.20}} & 91.85{\scriptsize$\pm$0.21} & 91.01{\scriptsize$\pm$0.27} & 93.04{\scriptsize$\pm$0.21} & 86.04{\scriptsize$\pm$0.15} \\
        CIDER & 76.04{\scriptsize$\pm$2.37} & 85.13{\scriptsize$\pm$1.13} & 80.58{\scriptsize$\pm$1.75} & 90.69{\scriptsize$\pm$2.13} & 92.38{\scriptsize$\pm$1.35} & 88.92{\scriptsize$\pm$1.58} & 90.66{\scriptsize$\pm$1.68} & -- \\
        UMAP & 77.94{\scriptsize$\pm$0.42} & 84.22{\scriptsize$\pm$0.66} & 81.08{\scriptsize$\pm$0.39} & 93.25{\scriptsize$\pm$0.69} & 93.19{\scriptsize$\pm$0.16} & 88.42{\scriptsize$\pm$0.35} & 91.62{\scriptsize$\pm$0.29} & 86.37{\scriptsize$\pm$0.08} \\
        SNN & 77.47{\scriptsize$\pm$0.24} & 85.18{\scriptsize$\pm$0.31} & 81.33{\scriptsize$\pm$0.19} & 92.82{\scriptsize$\pm$0.36} & \underline{94.55{\scriptsize$\pm$0.10}} & 89.45{\scriptsize$\pm$0.19} & 92.28{\scriptsize$\pm$0.21} & \underline{86.56{\scriptsize$\pm$0.03}} \\
        T2FNorm & 79.00{\scriptsize$\pm$0.16} & \underline{86.99{\scriptsize$\pm$0.02}} & 83.00{\scriptsize$\pm$0.07} & \textbf{96.87{\scriptsize$\pm$0.13}} & 91.95{\scriptsize$\pm$0.20} & \underline{91.81{\scriptsize$\pm$0.19}} & 93.55{\scriptsize$\pm$0.17} & \textbf{86.87{\scriptsize$\pm$0.19}} \\
        \midrule
        SPCP & 79.80{\scriptsize$\pm$0.05} & 85.24{\scriptsize$\pm$0.06} & 82.52{\scriptsize$\pm$0.03} & 92.91{\scriptsize$\pm$0.08} & 91.73{\scriptsize$\pm$0.21} & 89.75{\scriptsize$\pm$0.16} & 91.46{\scriptsize$\pm$0.14} & \underline{86.59{\scriptsize$\pm$0.10}} \\
        LogitNorm+SPCP & 79.02{\scriptsize$\pm$0.04} & \underline{87.37{\scriptsize$\pm$0.10}} & \underline{83.20{\scriptsize$\pm$0.07}} & \underline{96.23{\scriptsize$\pm$0.27}} & \underline{94.04{\scriptsize$\pm$0.31}} & \textbf{92.06{\scriptsize$\pm$0.28}} & \textbf{94.11{\scriptsize$\pm$0.28}} & 86.37{\scriptsize$\pm$0.09} \\
        \bottomrule[1.5pt]
        \end{tabular}
    }
    \end{center}
\end{table*}
% \vspace{2cm}
\begin{table*}[t]
    \caption{Fine-grained results (FPR95$\downarrow$) on the ImageNet-200 benchmark.}
    \begin{center}
    \label{apd:r6}
    \resizebox{0.985\linewidth}{!}{
        \begin{tabular}{@{\hspace{0.0cm}}l|@{\hspace{0.15cm}}c@{\hspace{0.15cm}}c@{\hspace{0.15cm}}c@{\hspace{0.15cm}}|c@{\hspace{0.15cm}}c@{\hspace{0.15cm}}c@{\hspace{0.15cm}}c@{\hspace{0.15cm}}|c@{\hspace{0.15cm}}c@{\hspace{0.0cm}}}
        \toprule[1.5pt]
         \multirow{2}{*}{\textbf{Method}} & \multicolumn{3}{c|}{\textbf{Near-OOD}} & \multicolumn{4}{c|}{\textbf{Far-OOD}} & \multirow{2}{*}{\textbf{ID ACC}} \\
        \cline{2-8}
        & \multicolumn{1}{c}{\textbf{SSB-hard}} & \multicolumn{1}{c}{\textbf{NINCO}} & \multicolumn{1}{c|}{\textbf{Average}} & \multicolumn{1}{c}{\textbf{iNaturalist}} & \multicolumn{1}{c}{\textbf{Textures}} & \multicolumn{1}{c}{\textbf{OpenImage-O}} & \multicolumn{1}{c|}{\textbf{Average}} & \\
        \midrule
        \multicolumn{9}{c}{\textit{Post-hoc methods (vanilla training with cross-entropy)}} \\
        MSP & \underline{66.00{\scriptsize$\pm$0.10}} & \underline{43.65{\scriptsize$\pm$0.75}} & \textbf{54.82{\scriptsize$\pm$0.35}} & 26.48{\scriptsize$\pm$0.73} & 44.58{\scriptsize$\pm$0.68} & 35.23{\scriptsize$\pm$0.18} & 35.43{\scriptsize$\pm$0.38} & 86.37{\scriptsize$\pm$0.08} \\
        Energy & 69.77{\scriptsize$\pm$0.32} & 50.70{\scriptsize$\pm$0.89} & 60.24{\scriptsize$\pm$0.57} & 26.41{\scriptsize$\pm$2.29} & 41.43{\scriptsize$\pm$1.85} & 36.74{\scriptsize$\pm$1.14} & 34.86{\scriptsize$\pm$1.30} & 86.37{\scriptsize$\pm$0.08} \\
        % VIM & 71.28{\scriptsize$\pm$0.49} & 47.10{\scriptsize$\pm$1.10} & 59.19{\scriptsize$\pm$0.71} & 27.34{\scriptsize$\pm$0.38} & \textbf{20.39{\scriptsize$\pm$0.17}} & 33.86{\scriptsize$\pm$0.63} & 27.20{\scriptsize$\pm$0.30} & 86.37{\scriptsize$\pm$0.08} \\
        DICE & 70.84{\scriptsize$\pm$0.30} & 52.91{\scriptsize$\pm$1.20} & 61.88{\scriptsize$\pm$0.67} & 29.66{\scriptsize$\pm$2.62} & 40.96{\scriptsize$\pm$1.87} & 38.91{\scriptsize$\pm$1.16} & 36.51{\scriptsize$\pm$1.18} & 86.37{\scriptsize$\pm$0.08} \\
        ReAct & 71.51{\scriptsize$\pm$1.92} & 53.47{\scriptsize$\pm$2.46} & 62.49{\scriptsize$\pm$2.19} & 22.97{\scriptsize$\pm$2.25} & 29.67{\scriptsize$\pm$1.35} & 32.86{\scriptsize$\pm$0.74} & 28.50{\scriptsize$\pm$0.95} & 86.37{\scriptsize$\pm$0.08} \\
        ASH & 72.14{\scriptsize$\pm$0.97} & 57.63{\scriptsize$\pm$0.98} & 64.89{\scriptsize$\pm$0.90} & 22.49{\scriptsize$\pm$2.24} & \underline{25.65{\scriptsize$\pm$0.80}} & 33.72{\scriptsize$\pm$0.97} & 27.29{\scriptsize$\pm$1.12} & 86.37{\scriptsize$\pm$0.08} \\
        SCALE & \underline{67.39{\scriptsize$\pm$0.63}} & 47.20{\scriptsize$\pm$1.26} & 57.29{\scriptsize$\pm$0.90} & 18.40{\scriptsize$\pm$1.86} & 29.77{\scriptsize$\pm$0.50} & 31.21{\scriptsize$\pm$1.06} & 26.46{\scriptsize$\pm$0.81} & 86.37{\scriptsize$\pm$0.08} \\
        \midrule
        \multicolumn{9}{c}{\textit{Training-time regularization methods}} \\
        % MOS & 74.35{\scriptsize$\pm$0.32} & 68.85{\scriptsize$\pm$0.68} & 71.60{\scriptsize$\pm$0.48} & 49.55{\scriptsize$\pm$0.73} & 51.27{\scriptsize$\pm$1.02} & 53.86{\scriptsize$\pm$0.30} & 51.56{\scriptsize$\pm$0.42} & 85.60{\scriptsize$\pm$0.20} \\
        % VOS & 69.93{\scriptsize$\pm$0.47} & 49.85{\scriptsize$\pm$0.71} & 59.89{\scriptsize$\pm$0.47} & 25.53{\scriptsize$\pm$1.36} & 39.74{\scriptsize$\pm$1.17} & 36.77{\scriptsize$\pm$0.94} & 34.01{\scriptsize$\pm$0.97} & 86.23{\scriptsize$\pm$0.19} \\
        LogitNorm & 67.46{\scriptsize$\pm$0.21} & 45.46{\scriptsize$\pm$0.69} & 56.46{\scriptsize$\pm$0.37} & \underline{15.70{\scriptsize$\pm$0.61}} & 32.13{\scriptsize$\pm$0.67} & \underline{30.49{\scriptsize$\pm$0.62}} & \underline{26.11{\scriptsize$\pm$0.52}} & 86.04{\scriptsize$\pm$0.15} \\
        CIDER & 75.50{\scriptsize$\pm$0.68} & 44.69{\scriptsize$\pm$0.88} & 60.10{\scriptsize$\pm$0.73} & 26.54{\scriptsize$\pm$2.27} & 31.51{\scriptsize$\pm$3.68} & 32.47{\scriptsize$\pm$2.40} & 30.17{\scriptsize$\pm$2.75} & -- \\
        UMAP & 70.33{\scriptsize$\pm$0.65} & 51.28{\scriptsize$\pm$1.55} & 60.81{\scriptsize$\pm$0.84} & 26.23{\scriptsize$\pm$1.76} & 31.71{\scriptsize$\pm$0.75} & 39.46{\scriptsize$\pm$1.02} & 32.47{\scriptsize$\pm$0.67} & 86.37{\scriptsize$\pm$0.08} \\
        SNN & 73.11{\scriptsize$\pm$0.93} & 46.60{\scriptsize$\pm$0.28} & 59.85{\scriptsize$\pm$0.46} & 25.66{\scriptsize$\pm$0.76} & \underline{24.81{\scriptsize$\pm$0.66}} & 33.64{\scriptsize$\pm$0.60} & 28.04{\scriptsize$\pm$0.64} & \underline{86.56{\scriptsize$\pm$0.03}} \\
        T2FNorm & \textbf{65.94{\scriptsize$\pm$0.27}} & \underline{44.09{\scriptsize$\pm$0.79}} & \underline{55.01{\scriptsize$\pm$0.36}} & \textbf{13.47{\scriptsize$\pm$0.76}} & 33.46{\scriptsize$\pm$0.34} & \underline{29.17{\scriptsize$\pm$0.69}} & \underline{25.37{\scriptsize$\pm$0.55}} & \textbf{86.87{\scriptsize$\pm$0.19}} \\
        \midrule
        SPCP & 70.23{\scriptsize$\pm$0.21} & 49.32{\scriptsize$\pm$0.40} & 59.77{\scriptsize$\pm$0.25} & 23.86{\scriptsize$\pm$0.94} & 33.83{\scriptsize$\pm$0.69} & 33.59{\scriptsize$\pm$0.34} & 30.43{\scriptsize$\pm$0.45} & \underline{86.59{\scriptsize$\pm$0.10}} \\
        LogitNorm+SPCP & 68.58{\scriptsize$\pm$0.51} & \textbf{42.09{\scriptsize$\pm$0.39}} & \underline{55.33{\scriptsize$\pm$0.45}} & \underline{14.94{\scriptsize$\pm$1.07}} & \textbf{24.50{\scriptsize$\pm$0.78}} & \textbf{26.40{\scriptsize$\pm$0.59}} & \textbf{21.95{\scriptsize$\pm$0.73}} & 86.37{\scriptsize$\pm$0.09} \\
        \bottomrule[1.5pt]
        \end{tabular}
    }
    \end{center}
\end{table*}

\newpage

\end{document}